\RequirePackage{fix-cm}

\documentclass[twocolumn]{svjour3}

\smartqed

\usepackage{amsmath}
\usepackage{hyperref}
\usepackage{cite}
\usepackage{doi}
\usepackage{multirow}
\usepackage{algorithmic}
\usepackage{arydshln}
\usepackage{afterpage}
\usepackage[caption=false,font=normalsize,labelfont=sf,textfont=sf]{subfig}
\usepackage{graphicx,color}
\usepackage{epstopdf}
\usepackage{bm}
\newcommand{\DEpfo}{DE$_{\text{pfo}}$}
\newcommand{\DElscr}{DE$_{\text{lscr}}$}
\newcommand{\DEcr}{DE$_{\text{cr}}$}
\newcommand{\DEls}{DE$_{\text{ls}}$}
\newcommand{\CALL}[2]{\texttt{{#1}(#2)}}
\newcommand{\IIF}[2]{\STATE\algorithmicif\ {#1}\ \algorithmicthen\ {#2} \algorithmicend\ \algorithmicif}
\newcommand{\IIFELSE}[3]{\STATE\algorithmicif\ {#1}\ \algorithmicthen\ {#2} \\ \algorithmicelse {~#3} \algorithmicend\ \algorithmicif}
\newcommand{\IDOWHILE}[2]{\STATE\algorithmicdo\ {#1}\ \algorithmicwhile\ {#2} \algorithmicend~\algorithmicdo}

\newcommand{\PROCEDURE}[1]{
	\renewcommand{\algorithmicwhile}{\textbf{procedure}}
	\renewcommand{\algorithmicdo}{ }
	\WHILE{#1}
	\renewcommand{\algorithmicwhile}{\textbf{while}}
	\renewcommand{\algorithmicdo}{\textbf{do}}	
}
\newcommand{\ENDPROCEDURE}{\ENDWHILE}
\definecolor{Gray}{rgb}{0.9,0.9,0.9}
\newcommand{\markg}[1]{\setlength{\fboxsep}{2pt}{\colorbox{Gray}{#1}}}

\renewcommand{\vec}[1]{\bm{#1}}

\begin{document}
\author{Borko Bo\v{s}kovi\'{c} \and Janez Brest}

\institute{B. Bo\v{s}kovi\'{c} \and J. Brest \at
              Faculty of Electrical Engineering and Computer Science,\\ University of Maribor, SI-2000 Maribor, Slovenia \\
              \email{borko.boskovic@um.si, janez.brest@um.si}            
}

\title{Protein Folding Optimization using Differential Evolution Extended with Local Search and Component Reinitialization}

\titlerunning{Protein Folding Optimization using DE with Local Search and Component Reinitialization}        

\dedication{\begin{tabular}{p{17cm}}
 \hline
 \multicolumn{1}{|p{17cm}|}{
 \footnotesize This version, created on \today, is an electronic reprint of the original article published in the Information Sciences Journal, 2018, Vol. 454-455,
 pp. 178--199, doi:
 \href{https://doi.org/10.1016/j.ins.2018.04.072}{10.1016/j.ins.2018.04.072}. This reprint differs from the original in pagination and typographic detail.} \\
 \hline
\end{tabular}}

\date{}

\maketitle


\begin{abstract}
This paper presents a novel Differential Evolution algorithm for protein folding optimization that is applied to a three-dimensional
AB off-lattice model. The proposed algorithm includes two new mechanisms. A local search is used to improve convergence speed and to reduce the runtime
complexity of the energy calculation. For this purpose, a local movement is introduced within the local search. The designed evolutionary algorithm has
fast convergence speed and, therefore, when it is trapped into the local optimum or a relatively good solution is located, it is hard to
locate a better similar solution. The similar solution is different from the good solution in only a few components. A component reinitialization method
is designed to mitigate this problem. Both the new mechanisms and the proposed algorithm were analyzed on well-known amino acid sequences
that are used frequently in the literature. Experimental results show that the employed new mechanisms improve the efficiency of our algorithm and 
that the proposed algorithm is superior to other state-of-the-art algorithms. It obtained a hit ratio of 100\% for sequences up to 18
monomers, within a budget of $10^{11}$ solution evaluations. New best-known solutions were obtained for most of the sequences. The existence
of the symmetric best-known solutions is also demonstrated in the paper.

\end{abstract}

\section{Introduction}
\label{sec:intro}
The protein structure prediction represents the problem of how to predict the native structure of a protein from its amino acid sequence. This problem is one of
the more important challenges of this century~\cite{Kennedy05} and, because of its nature, it attracts scientists from different fields, such as Physics, Chemistry,
Biology, Mathematics, and Computer Science. Within the protein structure prediction, the Protein Folding Optimization (PFO) represents a computational problem for
simulating the protein folding process and finding a native structure. Most proteins must fold into a unique three-dimensional structure, known as a native structure,
to perform their biological function~\cite{Balchin16}. A protein's function is determined by its structure. The inability of a protein to form its native structure prevents
a protein from fulfilling its function correctly, and this may be the basis of various human diseases~\cite{Alfonso15}.

The PFO belongs to the class of NP-hard problems~\cite{Fraenkel93} and, with current algorithms and computational resources, it is possible to predict the native
structures of relatively small proteins. The reason for that is the huge and multimodal search space. For example, a polypeptide that has only 18 amino acids,
will have 31 angles within a simplified AB model (see Section~\ref{sec:model}). Using uniform discretization with only 10 values for each angle, there would
be $10^{31}$ possible configurations. To evaluate and select the correctly folded conformation among all these conformations in the time elapsed since the
Big Bang, we need the huge computational speed of $10^{31}/(4.32\cdot10^{17})=2.31\cdot10^{13}$ conformation evaluations per second. This is much faster than
the speed obtained within our experiment, where we can evaluate only $5.73\cdot10^5$ conformations per second. From these numbers, we can see that the search
space is huge, even in the simplified model, which makes this problem very hard. However, in reality, the proteins fold into their native conformation on a time
scale of seconds, and this contradiction is known as Levinthal's paradox~\cite{buxbaum2007}. An optimization algorithm can give good results of a PFO problem only
if it can locate good solutions and evaluate solutions efficiently. Here, the approximation techniques, such as heuristic and metaheuristic, with
efficient data structures, become the only viable alternatives as the problem size increases.

Some simplified protein models exist, such as HP models within different lattices~\cite{BoskovicHP16} and the AB off-lattice model~\cite{Stillinger93}.
Simplified protein models were designed for development, testing, and comparison of different approaches. The AB off-lattice
model was used in the paper for demonstrating the efficiency of the proposed algorithm. This model takes into account the hydrophobic interactions which
represent the main driving forces of a protein structure formation and, as such, still imitates its main features realistically~\cite{Huang06}. Although this
model is incomplete, it allows the development, testing, and comparison of various search algorithms, and offers a global perspective of protein structures.
It can be helpful in confirming or questioning important theories~\cite{Bazzoli04}. 

Our algorithm is based on the Differential Evolution (DE) algorithm that was proposed by Storn and Price~\cite{ga97aRStorn}. It is a powerful
stochastic population-based  algorithm. Three simple operators, mutation, crossover, and selection,
were used inside the DE algorithm to transform real-coded individuals with the purpose to locate optimal or sub-optimal solutions. Because
of its simplicity and efficiency, it was used in various numerical optimization problems, such as an animated trees reconstruction~\cite{Zamuda14a},
an intrusion detection~\cite{Abdulla17}, and an image thresholding~\cite{Mlakar16}.
An advanced DE variant, such as L-SHADE~\cite{Tanabe14} was also the winner of the recent CEC (IEEE Congress on Evolutionary Computation)
competitions. For more details about DE, we refer the reader to~\cite{Piotrowski17} and to survey~\cite{Das16}.

It has been shown that the PFO has a highly rugged landscape structure containing many local optima and needle-like
funnels~\cite{Jana17a}, and, therefore, the algorithms that follow more attractors simultaneously are ineffective. In our recent
work~\cite{BoskovicAB16}, to overcome this weakness, we proposed a Differential Evolution (DE) algorithm that uses the
$\mathit{DE/best/1/bin}$ strategy. With this strategy, our algorithm follows only one attractor. The temporal locality 
mechanism~\cite{Wong12} and self-adaptive mechanism~\cite{Brest06} of the main control parameters were used additionally
to speed up the convergence speed. When the algorithm was trapped in a local optimum, then random reinitialization was used.
This algorithm belongs to the ab-initio PFO methods, which optimize structures from scratch, and do not require any information
about related sequences. It showed a very fast convergence speed, and it was capable of obtaining significantly better results than
other state-of-the-art algorithms.

Taking into account the finding of the previous paragraph, we propose two new mechanisms, that, additionally, improve the
efficiency of our algorithm. A new local search mechanism was designed in order to improve convergence speed and to reduce the runtime complexity
of the algorithm. A similar idea was already used within the HP model~\cite{BoskovicHP16}, where it is applied to the cubic lattice. Using a
simple local search mechanism, where only one solution's component is changed, can produce a structure whereby a lot of monomers are moved.
This means their positions must be recalculated and efficient energy calculation is not possible. In contrast to simple local search, our
mechanism improves the quality of conformations using the local movements within the three-dimensional AB off-lattice model. We define
a local movement as a transformation of conformation, whereby only two consecutive monomers are moved locally in such a way that the 
remaining monomers remain in their positions. The described local movement allows efficient evaluation of neighborhood
solutions and faster convergence speed.

With the fast convergence speed the algorithm can locate good solutions quickly, but it has a problem locating good similar
solutions. For example, if an algorithm locates a good solution that is different from the global best solution in only one or few components,
then the random restart, that was used in our previous work, is not an efficient solution. For that purpose,
a component reinitialization was designed and incorporated within our algorithm. This mechanism is employed when the local best solution is
detected. Instead of the random restart, it produces similar solutions that are different from the local best solution in only a few components.

We called the proposed algorithm \DElscr\ and it was tested on two sets of amino acid sequences that were used frequently in the literature.
The first set
included 18 real peptide sequences, and the second set included 4 well-known artificial Fibonacci sequences with different lengths. Experimental results
show that the proposed mechanisms improve the efficiency of the algorithm, and the algorithm is superior to other state-of-the-art algorithms. Its
superiority is especially evident for longer sequences. With the proposed algorithm, that is stochastic, we cannot prove the optimality of
the obtained conformations. However, we can infer about them according to the observed hit ratio. The experimental results show that our algorithm obtained
a hit ratio of 100\% for sequences that contain up to 18 monomers. For all longer sequences, we can only report the best-known conformations
that are almost surely not optimal. Based on these observations, the main contributions of this paper are: 
\begin{itemize}
  \item [1.] The proposed new DE algorithm for the PFO on a three-dimensional AB off-lattice model.
  \item [2.] The local search mechanism that improves convergence speed and reduces runtime complexity of solution evaluations within the neighborhood.
  \item [3.] The component reinitialization, which increases the likelihood of finding a good similar solution.
  \item [4.] With the observed hit ratios, we show how difficult the PFO is, even in a simplified model, and that, with the current algorithm,
	    we can confirm solutions with a hit ratio of 100\% only for sequences that have up to 18 monomers.
  \item [5.] An approach for determining the algorithm's asymptotic average-case performances.
  \item [6.] The existence of two best-known (potentially global best) structures that are symmetrical for all sequences with up to 25 monomers.
  \item [7.] The new best-known conformations for most of the sequences.
\end{itemize}

The remainder of this paper is organized as follows. A related work for the PFO on a three-dimensional AB off-lattice model is described in Section~\ref{sec:related}.
The three-dimensional AB off-lattice model is described in Section~\ref{sec:model}. A description of the introduced algorithm, with the emphasis on new mechanisms 
is given in Section~\ref{sec:algorithm}. The experimental setup and numerical results are presented in Section~\ref{sec:experiments}. Section~\ref{sec:conclusions} 
concludes this paper.

\section{Related work}
\label{sec:related}

Over the years, different algorithms have been applied successfully to the PFO on a three-dimensional AB off-lattice model.
In~\cite{Hsu03b}, the low energy configurations are optimized using the Pruned-Enriched-Rosenbluth Method (PERM). 
This method was also applied to the lattice model quite successfully~\cite{Thachuk2007}. Its improved variants
are still state-of-the-art for the lattice model. Although PERM showed potential, it was not successful for more realistic models
such as the AB off-lattice model.
The conformational space annealing was studied using Fibonacci sequences in~\cite{Kim05} and compared with nPERMis (new PERM with
importance sampling)~\cite{Hsu03a}. Next, an algorithm that outperforms PERM was proposed in~\cite{Chen06}.
In this work, the problem is converted from a nonlinear constraint-satisfied problem to an unconstrained optimization problem which
can be solved by the well-known gradient method. The statistical temperature molecular dynamics based algorithm ``statistical
temperature annealing'' was applied to an AB model in~\cite{Kim07}. This algorithm shows the ability to find better
conformations in comparison to previous algorithms. The efficiency of an improved tabu search algorithm was analyzed 
in~\cite{Zhang08}. According to the characteristics of PFO, the following improved strategies were incorporated into
the tabu search: 
(1) A heuristic method of generating an initial solution. Within these initial solutions, hydrophobic monomers are located in the core,
whereas hydrophilic monomers are located outside of the core of the conformation, (2) A method for neighborhood generation
that is based on the mutation method from genetic algorithms, (3) Selection of a candidate set that specifies the subset of the neighborhood of the current
solution. The purpose of a candidate set was to provide solutions that can replace the current solution, and (4) A mechanism for avoiding
stagnation within local optima. The following hybrid algorithms were also developed for the AB model: A hybrid algorithm that combines
the genetic algorithm and tabu search algorithm~\cite{Wang11}, particle swarm optimization and levy flight~\cite{Chen11},
the particle swarm optimization, genetic algorithm, and tabu search algorithm~\cite{Zhou14a}, and improved genetic algorithm and particle
swarm optimization algorithm with multiple populations~\cite{Zhou14b}. An improved harmony search algorithm, that is combined
with dimensional mean based perturbation strategy~\cite{Jana17b} and an artificial bee colony algorithm~\cite{Li15} were
also applied to PFO on the AB off-lattice model. A Balance Evolution Artificial Bee Colony (BE-ABC) algorithm outperforms all predecessors
significantly. This algorithm is featured by the adaptive adjustment of search intensity to cater for the varying needs during the entire
optimization process.

The authors in~\cite{Jana17a} determined the structural features of the PFO using Fitness Landscape Analysis (FLA) techniques based on the generated landscape path. 
From the results of FLA, it has been shown that the PFO has a highly rugged landscape structure containing many local optima and needle-like funnels, with no global
structure that characterizes the PFO complexity. The obtained results also show that the artificial bee colony algorithm outperforms all other
algorithms significantly in all instances for the three-dimen-sional AB off-lattice model.

\begin{figure*}[t!]
\centering
\subfloat[]{\makebox[.49\textwidth]{\includegraphics[scale=0.15]{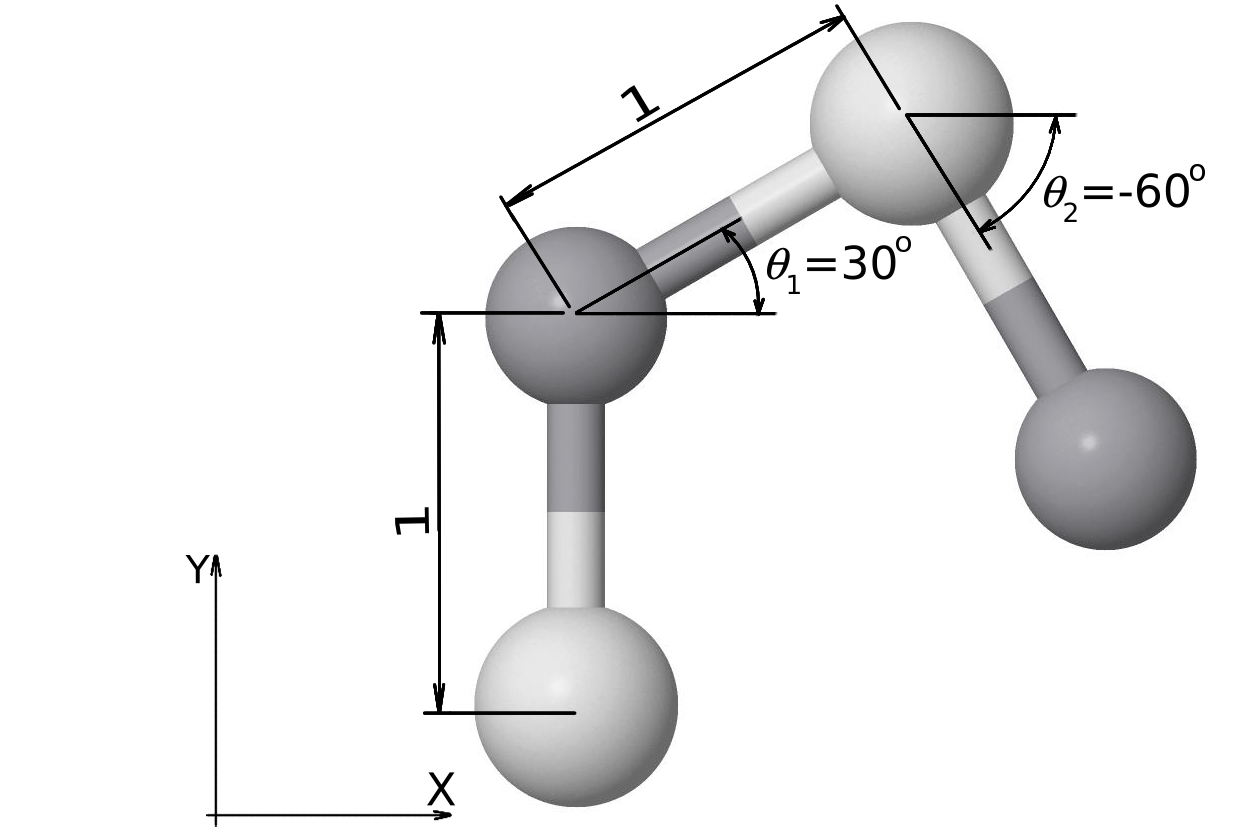}}}
\subfloat[]{\makebox[.49\textwidth]{\includegraphics[scale=0.15]{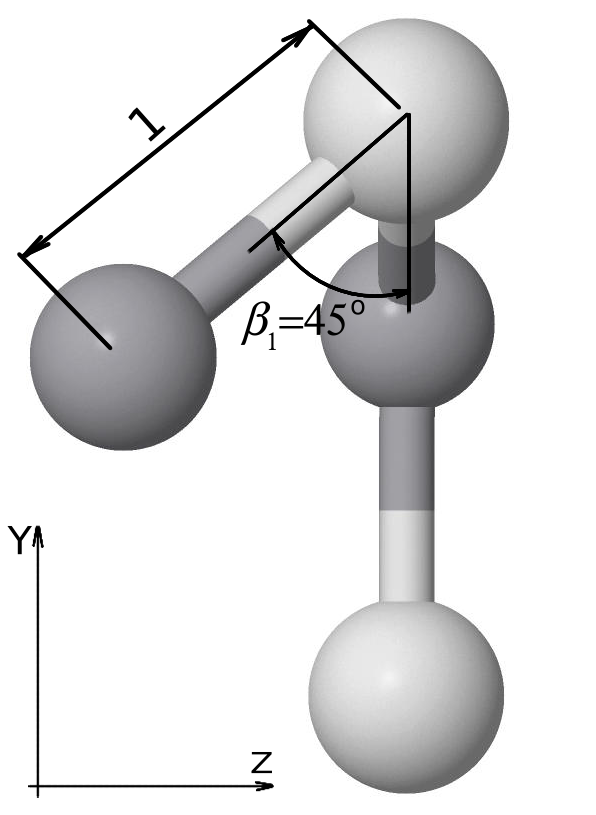}}}
\caption{A schematic diagram of the sequence ABAB. (a) Projection of a structure with $\theta_1=30$, $\theta_2=-60$ and $\beta_1=0$ onto the XY-plane. 
(b) Projection of a structure with $\theta_1=30$, $\theta_2=-60$ and $\beta_1=45$ onto the ZY-plane.}
\label{fig:model}
\end{figure*}

In our recent work~\cite{BoskovicAB16}, we proposed a Differential Evolution algorithm that is adapted to PFO on a three-dimensional AB off-lattice model.
In contrast to previous population-based algorithms for PFO, this algorithm was designed to follow only one attractor. Within this algorithm, we
incorporated a self-adaptive mechanism, a mutation strategy for the fast convergence speed and a temporal locality. The obtained results of this algorithm show
that it is superior to the algorithms from the literature, including the artificial bee colony algorithm, and significantly lower free energy values were
obtained for longer AB sequences.

\section{Three-dimensional AB off-lattice model}
\label{sec:model}

The basic building blocks of proteins are amino acids. The linear chain of amino acids is a polypeptide, and a protein contains at least one long polypeptide.
Each polypeptide can be represented with a unique amino acid sequence. The polypeptide must fold into a specific three-dimensional native structure before it
can perform its biological function(s)~\cite{petsko2004}. Thus, all information necessary for folding must be contained in the amino acid sequence, and this
is known as the Anfinsen-hypothesis~\cite{buxbaum2007}.

From the amino acid sequence, it is possible to generate different conformations, which is also dependent on the used model. In general, two types of 
simplified models exist: Off-lattice and lattice. The lattice model maps each position of amino acid to a point on a discrete lattice. In contrast to the lattice model,
the off-lattice model allows any position and, as such, is more accurate. The simplified three-dimensional AB off-lattice model was proposed in~\cite{Stillinger93}. 
Instead of 20 standard amino acids, this model uses only two different types of amino acids: $A$ -- hydrophobic and $B$ -- hydrophilic. Thus, an
amino acid sequence is represented as a string $\vec{s}=\{s_1, s_2, ..., s_L\}$, $s_i \in \{A, B\}$, where $A$ represents a hydrophobic, $B$ a hydrophilic 
amino acid and $L$ the length of the sequence. The three-dimensional structure of an AB sequence is defined by bond angles 
$\vec{\theta}=\{\theta_1,\theta_2, ..., \theta_{L-2}\}$, torsional angles $\vec{\beta}=\{\beta_1, \beta_2, ...,$ $ \beta_{L-3}\}$ and the unit-length chemical bond between
two consecutive amino acids (see Fig.~\ref{fig:model}).

Different energy calculations can be used within different models. Within an AB model, the free energy value is calculated using a simple trigonometric form of backbone
bend potentials $E_1(\vec{\theta})$ and a species-dependent Lennard-Jones 12,6 form of non-bonded interactions $E_2(\vec{s},\vec{\theta},\vec{\beta})$
as shown in the following equation~\cite{Stillinger93}:

\begin{footnotesize}
\begin{eqnarray}
  \label{eq:energy}
  E(\vec{s},\vec{\theta},\vec{\beta}) &=& E_1(\vec{\theta})+E_2(\vec{s},\vec{\theta},\vec{\beta}) \nonumber \\
  E_{\text{\tiny 1}}(\vec{\theta}) &=& \frac{1}{4}\displaystyle\sum_{i=1}^{L-2}[1-cos(\theta_i)]\\
  E_{\text{\tiny 2}}(\vec{s},\vec{\theta},\vec{\beta}) &=& 4\displaystyle\sum_{i=1}^{L-2}\displaystyle\sum_{\text{{\it j}={\it i}+}2}^{L}[d(\vec{p}_i,\vec{p}_j)^{\text{--}12} \text{--} c(s_i,s_j)\cdot d(\vec{p}_i,\vec{p}_j)^{\text{--}6}] \nonumber
\end{eqnarray}
\end{footnotesize}

\noindent
where $\vec{p}_i = \{x_i,y_i,z_i\}$ represents the position of the $i$-th amino acid within the three-dimensional space.
These positions are determined as shown in Fig.~\ref{fig:model} and by the following equation:

\begin{eqnarray}
 \label{eq:positions}
 \vec{p}_i=\begin{cases}
	\{0,0,0\}&\text{if}~i=1,\\
	\{0,1,0\}&\text{if}~i=2,\\
	\{cos(\theta_1),1+sin(\theta_1),0\}&\text{if}~i=3,\\
	\{x_{i-1}+cos(\theta_{i-2})\cdot cos(\beta_{i-3}),\\ 
	  ~~y_{i-1}+sin(\theta_{i-2})\cdot cos(\beta_{i-3}),&\text{if 4}\le i\le L.\\
	 ~~z_{i-1}+sin(\beta_{i-3})\}\\
	\end{cases}
\end{eqnarray}

\noindent
In Eq.~(\ref{eq:energy}) $d(\vec{p}_i,\vec{p}_j)$ denotes the Euclidean distance between positions $\vec{p}_i$ and $\vec{p}_j$, while $c(s_i,s_j)$ 
determines the attractive, weak attractive or weak repulsive non-bonded interaction for the pair $s_i$ and $s_j$, as shown in the following equation:

\begin{eqnarray}
 c(s_i,s_j) = \begin{cases}
                1    & \text{if }s_i = A \text{ and } s_j = A,\\ 
                0.5 & \text{if }s_i = B \text{ and } s_j = B,\\
                -0.5  & \text{if }s_i \neq s_j.\\
               \end{cases}
               \nonumber
\end{eqnarray}

\noindent
The objective of PFO within the context of an AB off-lattice model is to simulate the folding process and to find the angles' vector or conformation
that minimizes the free-energy value:

\begin{equation}
\{\vec{\theta}^*,\vec{\beta}^*\} = \operatorname*{arg\,min} E(\vec{s},\vec{\theta},\vec{\beta}). \nonumber
\end{equation}

\section{Method}
\label{sec:algorithm}
In this paper, we extend our Differential Evolution algorithm~\cite{BoskovicAB16} with two new mechanisms. The first mechanism is a local search that
improves convergence speed and reduces runtime complexity for solution evaluation within a specific neighborhood. The second mechanism is component
reinitialization, which allows the algorithm to locate good similar conformations according to the local best solution. 

\begin{figure}[!t]
\scriptsize
\begin{algorithmic}[1]
\algsetup{linenosize=\scriptsize}
\PROCEDURE{\CALL{\DElscr}{$\vec{s},\mathit{Np}$}}	
\STATE {Initialize a population $P$ \label{alg:init} \\ 
	$\{ \vec{x}_{i},~~F_{i}=0.5,~~\mathit{Cr}_{i}=0.9,~~e_i=$\CALL{$E$}{$\vec{s},\vec{x_i}$}$\} \in P$ \\
	$ x_{i,j} = -\pi + 2 \cdot \pi \cdot \mathit{rand}_{[0,1]}$ \\
	$ i = 1, 2, ..., \mathit{Np};~~j = 1, 2, ..., D;~~D = 2 \cdot $ \CALL{$length$}{$\vec{s}$}$ - 5$}\\
	\colorbox{Gray}{$ \{\vec{x}_{b},e_{b}\} = \{\vec{x}_{b}^{l},e_{b}^{l}\} = \{\vec{x}_{b}^{p},e_{b}^{p}\} = $ \CALL{BEST}{$P$}}
\WHILE {stopping criteria is not met} \label{alg:gen}
	\FOR{$i=1$ \TO $\mathit{Np}$}
		\IIFELSE{$\mathit{rand}_{[0,1]} < 0.1 $} {$F = 0.1 + 0.9 \cdot \mathit{rand}_{[0,1]} $}{$F = F_{i}$} \label{alg:jde:a}
		\IIFELSE{$\mathit{rand}_{[0,1]} < 0.1 $}{$\mathit{Cr} = \mathit{rand}_{[0,1]}$}{$\mathit{Cr} = \mathit{Cr}_{i}$} \label{alg:jde:b}
		\IDOWHILE{$r_{_1}\text{=}\mathit{rand}_{\{1,\mathit{Np}\}}$}{$r_{_1}\text{=}i$} \label{alg:str:a}
		\IDOWHILE{$r_{_2}\text{=}\mathit{rand}_{\{1,\mathit{Np}\}}$}{$r_{_2}\text{=}i~\text{\bf or}~r_{_2}\text{=}r_{_1}$}
		\STATE{$j_{\mathit{rand}} = \mathit{rand}_{\{1,\mathit{D}\}}$}
		\FOR{$j=1$ \TO $D$}
 			\IF{$\mathit{rand}_{[0,1]} < \mathit{Cr}~\text{\bf or }~j = j_{\mathit{rand}}$}
 				\STATE{$ u_{j} = x_{b,j}+F\cdot(x_{r_1,j}-x_{r_2,j} ) $}
 				\IIF{$u_{j}\le\text{-}\pi$}{$u_{j} \text{\,=\,} 2\cdot\pi+u_{j}$}
 				\IIF{$u_{j}>\pi$}{$u_{j} \text{\,=\,} 2\cdot(\text{-}\pi)+u_{j}$}
 			\ELSE
				\STATE{$u_{j} = x_{i,j}$}
 			\ENDIF
		\ENDFOR \label{alg:str:b}
		\STATE{$e_u=$\CALL{$E$}{$\vec{s},\vec{u}$} // Energy calculation} \label{alg:eval}
		\IF{$e_u \le e_i$} \label{alg:impr:a}
			\STATE{// Temporal locality}
			\FOR{$j=1$ \TO $D$}
				\STATE{$u^*_{j} = x_{b,j} + 0.5 \cdot (u_{j} - x_{i,j}) $}
 				\IIF{$u^*_{j}\le\text{-}\pi$}{$u^*_{j} \text{\,=\,} 2\cdot\pi+u^*_{j}$}
 				\IIF{$u^*_{j}>\pi$}{$u^*_{j} \text{\,=\,} 2\cdot(\text{-}\pi)+u^*_{j}$}
			\ENDFOR
			\STATE{$e^*_u=$\CALL{$E$}{$\vec{s},\vec{u}^*$}}
			\IF{$e^*_{u} \le e_u$}
				\STATE{$\{\vec{x_i},F_i,\mathit{Cr_i},e_i\}=\{u^*,F,\mathit{Cr},e^*_{u}\}$}
			\ELSE
				\STATE{$\{\vec{x_i},F_i,\mathit{Cr_i},e_i\}=\{u,F,\mathit{Cr},e_{u}\}$}
			\ENDIF
			\STATE {\markg{// Local Search}}
			\FOR{\markg{$n=2$ \TO $L-1$}}
				\STATE \markg{$\theta_{n\text{-}1} = \mathit{rand}_{[0,1]}\cdot ( x^{p}_{b,n\text{-}1} - x_{i,n\text{-}1} )$}
				\STATE \markg{$\beta_{n\text{-}2} = \mathit{rand}_{[0,1]}\cdot ( x^{p}_{b,n+(L\text{-}4)} - x_{i,n+(L\text{-}4)})$}
				\STATE \markg{\{${\vec{v}, e_v}\} = $ \CALL{LOCAL\_MOVEMENT}{$\vec{x}^{p}_{b},n,\theta_{n\text{-}1},\beta_{n\text{-}2}$}} \label{alg:lm}
				\IIF{\markg{$e_{v}\le e_{b}$}}{\markg{$ \{\vec{x}^{p}_{b},e^{p}_{i}\}=\{\vec{v},e_{v}\}$}}
			\ENDFOR
		\ENDIF \label{alg:impr:b}
	\ENDFOR
	\STATE {\markg{$ \{\vec{x}_{b}^{p},e_{b}^{p}\} = $ \CALL{BEST}{$P$}}}
	\IIF{\markg{$e_{b}^{p} \le e_{b}$}}{\markg{$ \{\vec{x}_{b},e_{b}\} = \{\vec{x}_{b}^{p},e_{b}^{p}\}$ }}
	\STATE \markg{\CALL{REINITIALIZATION}{$\{\vec{x}_{b}^{p},e_{b}^{p}\}$,$\{\vec{x}_{b}^{l},e_{b}^{l}\}$,$P$}} \label{alg:reinit}
\ENDWHILE
\RETURN \markg{$\{\vec{x}_{b},e_{b}\}$}
\ENDPROCEDURE
\end{algorithmic}
\caption{The proposed \DElscr\ algorithm.}
\label{fig:algorithm}
\end{figure}

\subsection{Proposed algorithm}
Hereinafter, we will describe briefly the \DElscr\ algorithm that is shown in Fig.~\ref{fig:algorithm}. 
The lines that represent improvements according to the previous version are
highlighted with a gray background. The optimization process begins with initialization (line~\ref{alg:init}). Each iteration
of the $\mathit{while}$ loop (line~\ref{alg:gen}) represents one generation of the evolutionary process. Mutation, crossover,
and selection are performed for each population's individual \{$\vec{x}_1,\vec{x}_2, ..., \vec{x}_{\mathit{Np}}$\}
within one generation. The $\textit{DE/best/}1$ mutation strategy and binary crossover (lines \ref{alg:str:a} -- \ref{alg:str:b}) are
used for creating a trial individual~$\vec{u}$. The values of mutation $F$, and crossover $Cr$ control parameters
are set with the self-adaptive jDE mechanism (lines~\ref{alg:jde:a} and~\ref{alg:jde:b})~\cite{Brest06}. The trial individual is evaluated in
line~\ref{alg:eval} using Eq.~(\ref{eq:energy}). From this line, we can see that the individuals are
D-dimensional vectors that contain real coded bond $\vec{\theta}$ and torsional $\vec{\beta}$ angles:

\begin{eqnarray}
 \vec{x}_i &=& \{x_{i,1},x_{i,2},..., x_{i,D}\} = \nonumber \\ 
  &=& \{\theta_1,\theta_2,..., \theta_{L-2}, \beta_1,\beta_2,..., \beta_{L-3}\} \nonumber \\
 x_{i,j} &\in& [-\pi,\pi]; ~~~ D = 2 \cdot L-5 \nonumber \\
 i &=& 1,2,..., \mathit{Np}; ~~~j = 1,2,...,D. \nonumber 
\end{eqnarray}

\begin{figure*}[t!]
\centering
\includegraphics[width=\textwidth]{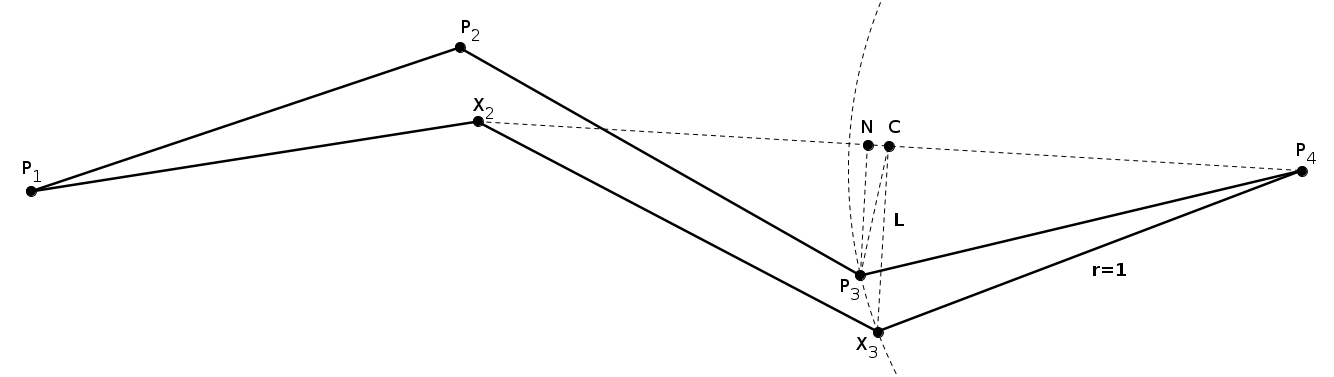}
\caption{A schematic diagram of a local movement. The polygon $\vec{P_1}$, $\vec{P_2}$, $\vec{P_3}$, $\vec{P_4}$ is transformed to polygon $\vec{P_1}$, 
$\vec{X_2}$, $\vec{X_3}$, $\vec{P_4}$ in such a way that only the two consecutive points $\vec{P_2}$ and $\vec{P_3}$ are moved to points $\vec{X_2}$ and
$\vec{X_3}$, while the remaining points $\vec{P_1}$ and $\vec{P_4}$ stay unchanged in their positions.}
\label{fig:lovalMove}
\end{figure*}

The selection mechanism, temporal locality and local search are shown in lines~\ref{alg:impr:a} -- \ref{alg:impr:b}. If the trial vector
is better than the corresponding vector from the population ($e_u < e_i$), then the second trial vector $\vec{u}^*$ is generated
using temporal locality~\cite{Wong12}, and the better vector replaces population vector $\vec{x}_i$. The next mechanism is a local
search. Within this mechanism, a local movement is used for improving the best population individual $\vec{x}_{b}^{p}$ according
to each pair $\{\theta_{n-1},\beta_{n-2}\}$. The values of $\{\theta_{n-1},\beta_{n-2}\}$ represent the angles that specify the 
position of the ($n+1$)-th monomer according to the position of the $n$-th monomer. After each generation, either random or component
reinitialization will be performed if the reinitialization criteria are satisfied (line~\ref{alg:reinit}). At the end of the evolutionary process,
the algorithm returns the best obtained solution $\vec{x}_b$ and its energy value $e_b$, as shown in line 46. The local search and
reinitialization are described in more detail in the following subsections. For a detailed description of the rest mechanisms that
were described briefly within this paragraph, we refer readers to~\cite{BoskovicAB16}.

\subsection{Local search}
The temporal locality and local search are performed if the trial vector is better than the corresponding population vector.
The local search calculates
the values of pair $\{\theta_{n-1},\beta_{n-2}\}$ for each monomer, with the exception of the first two monomers. The positions
of these two monomers are fixed
within the AB-model, see Eq.~(\ref{eq:positions}). For the rest of the monomers, the angle values are defined with a randomly scaled
difference between the 
best population individual $\vec{x}_b^p$ and current individual $\vec{x}_i$, as shown in the following equation:
\begin{eqnarray}
 \theta_{n-1} &=& \mathit{rand}_{[0,1]}\cdot (x^{p}_{b,n-1} - x_{i,n-1}) \nonumber \\ 
 \beta_{n-2} &=& \mathit{rand}_{[0,1]}\cdot (x^{p}_{b,n+(L-4)} - x_{i,n+(L-4)}) \nonumber \\
 i &=& 1, ..., \mathit{Np}; ~~~ n = 2, ..., L-1; \nonumber
\end{eqnarray}
A meticulous reader may notice that the $\theta$ has index $n-1$, while $\beta$ has index $n-2$. This means that, for $n=2$, the value of $\beta_0$
is calculated using the values of $\theta$ components. The reason for that is in the position of the third monomer, which is dependent only on
the $\theta_1$, see Eq.~(\ref{eq:positions}). Therefore, within the local movement for $n=2$, the value of $\beta_0$ is ignored, and local movement
takes into account only the value of $\theta_1$.

The local movement is a transformation of conformation whereby only two consecutive monomers are moved locally in such a way that the remaining 
mono-mers remain in their positions. There is only one exception, for the last monomer only one monomer is moved, while all the
remaining monomers remain in their positions. Fig.~\ref{fig:lovalMove} shows an example of two monomers' local movement. The polygon that is
defined with points $\vec{P_1}$, $\vec{P_2}$, $\vec{P_3}$,
$\vec{P_4}$ represents the section of original conformation where points represent the monomer positions. The local movement moves the point $\vec{P_2}$
to the point $\vec{X_2}$ according to the pair $\{\theta_{n-1},\beta_{n-2}\}$ while the position of point $\vec{X_3}$ is calculated using Eqs. 
(\ref{eq:c}) -- (\ref{eq:x3}). In these calculations, the point $\vec{X_3}$ is the nearest to the point $\vec{P_3}$ in such a way
that the new polygon $\vec{P_1}$, $\vec{X_2}$, $\vec{X_3}$, $\vec{P_4}$ must end at the point $\vec{P_4}$.

The calculation begins with the determination of point $\vec C$, whose position is in the middle of points $\vec{P_4}$ and $\vec{X_2}$:
\begin{equation}
 \label{eq:c}
 \vec{C} = \vec{X_2} + \frac{\vec{P_4} - \vec{X_2}}{2}.
\end{equation}
The length $L$ between points $\vec{C}$ and $\vec{X_3}$ is calculated by using the triangle $\vec{P_4}$, $\vec{C}$, $\vec{X_3}$ and Pythagoras's theorem:
\begin{equation}
 \label{eq:L}
 L = \sqrt{1 - \frac{\left\lVert \vec{P_4} - \vec{X_2} \right\rVert ^2}{4}}.
\end{equation}
The vector projection
of $(\vec{P_3} - \vec{C})$ onto line $\vec{X_2}$, $\vec{P_4}$ is calculated with the following equation:
\begin{equation}
 \label{eq:cn}
 \vec{C_N} = (\vec{P_3} - \vec{C}) \cdot \frac{\vec{P_4} - \vec{X_2}}{\left\lVert \vec{P_4} - \vec{X_2} \right\rVert}.
\end{equation}
At the end, point $\vec{X_3}$ is calculated by scaling 
of vector $(\vec{P_3} - \vec{C}) - \vec{C_N}$ as follows:
\begin{equation}
 \label{eq:x3}
 \vec{X_3} = \vec{C} + \frac{(\vec{P_3} - \vec{C}) - \vec{C_N}}{\left\lVert (\vec{P_3} - \vec{C}) - \vec{C_N} \right\rVert} \cdot L.
\end{equation}
The created polygon contains unchanged points $\vec{P_1}$ and $\vec{P_4}$, which means only monomers $\vec{X_2}$ and $\vec{X_3}$ are moved locally, while the
remaining monomers stay in their unchanged positions. This feature allows us to design the fast conformation evaluation within the local movement.
Two additional data structures, $\vec{E_1}$ and $\vec{E_2}$, were used for this purpose. The values of elements within these data structures are determined
according to Eq.~(\ref{eq:energy}), as follows:
\begin{eqnarray}
 \mathit{E}_{1_{i}} &=& 1-cos(\theta_i) \nonumber \\
 \mathit{E}_{2_{i,j}} &=& d(\vec{p}_i,\vec{p}_j)^{-12} - c(s_i,s_j)\cdot d(\vec{p}_i,\vec{p}_j)^{-6} \nonumber \\
 i &\in& \{1,...,L-2\}; ~~~ j \in \{i+2,...,L\}. \nonumber
\end{eqnarray}
Using these data structures for the best population individual and its energy value, we can calculate the energy of the conformation created by local
movement efficiently, as is shown in Eqs. (\ref{eq:e1}) -- (\ref{eq:new_e}), where $n$ represents the variable that was sent to the
local movement procedure, as shown in Fig. \ref{fig:algorithm} (see line~\ref{alg:lm}).
\begin{eqnarray}
 \label{eq:e1}
 \Delta e_1 &=& \mathit{E}_{1_{n-1}} - (1-cos(\theta_{n-1})) + \nonumber \\ 
     & & \mathit{E}_{1_n} - (1-cos(\theta_{n})) +  \\
     & & \mathit{E}_{1_{n+1}} - (1-cos(\theta_{n+1})) \nonumber
\end{eqnarray}
\begin{eqnarray}
 \label{eq:e2}
 \Delta e_2 &=& \sum_{i=n+1}^{n+3}\sum_{j=n+2}^{L} \Big[ \mathit{E}_{2_{i,j}} - \nonumber \\ 
     \hspace{1cm} & & \big( d(\vec{p}_i,\vec{p}_j)^{-12} - c(s_i,s_j)\cdot d(\vec{p}_i,\vec{p}_j)^{-6} \big) \Big] + \nonumber \\
     & & \sum_{i=n+1}^{n+3} \sum_{j=1}^{i} \Big[ \mathit{E}_{2_{j,i}} - \\
     \hspace{1cm} & & \big( d(\vec{p}_i,\vec{p}_j)^{-12} - c(s_i,s_j)\cdot d(\vec{p}_i,\vec{p}_j)^{-6} \big) \Big] \nonumber 
\end{eqnarray}
\begin{equation}
 \label{eq:new_e}
 e_{v} = e^{p}_{b} - (\frac{\Delta e_1}{4} + 4 \cdot \Delta e_2)
\end{equation}
From Eqs.~(\ref{eq:e1}) -- (\ref{eq:new_e}) we can observe that the time complexity of energy calculation is reduced from $\frac{L^2}{2}$ to $2L$.
In this way, the designed local movement allows faster evaluation of neighborhood solutions, and its usage wit\-hin local search
improves convergence speed.

\subsection{Reinitialization}
After each generation, reinitialization will be performed if the reinitialization criteria are satisfied. In our previous work,
random reinitialization was performed if the best population individual stayed unchanged within the evolution process for more than
$10^5$ evaluations. This number includes the number of evaluations for all individuals until the best population individual stayed unchanged.
Its value was determined in a way to prevent premature restarts, and to ensure some likelihood that the algorithm cannot improve the best
population individual by using the current population. In this work, we design a new reinitialization mechanism which has some advantages
over our previous work. For that purpose, the algorithm includes three types of the best individuals and three new control parameters,
as shown in Fig. 4. {\bf The individuals $\vec{x}^p_b, \vec{x}^l_b, \vec{x}_b$ represent the best population, local best and
global best individuals. The best population individual is the best individual in the current population, the local best
individual is the best individual among all similar individuals, and the global best individual is the best individual obtained within the
evolutionary process.} From this description of the best individuals, we can see that the main advantages of the proposed
reinitialization are to allow the following:

\begin{figure}[!t]
\scriptsize
\begin{algorithmic}[1]
\algsetup{linenosize=\scriptsize}
\PROCEDURE{\CALL{REINITIALIZATION}{$\{\vec{x}_{b}^{p},e_{b}^{p}\}$,$\{\vec{x}_{b}^{l},e_{b}^{l}\}$,$P$}}
	\IF{$\vec{x}_{b}^{p}$ is unchanged for at least $P_b\cdot D$ evaluations} \label{alg:cond:reinit}
		\IIF{$e_{b}^{p} \le e_{b}^{l}$}{$ \{\vec{x}_{b}^{l},e_{b}^{l}\} = \{\vec{x}_{b}^{p},e_{b}^{p}\}$ }
 		\IF{$\vec{x}_{b}^{l}$ is unchanged for $L_b \cdot D$ reinitializations} \label{alg:cond:componet}
			\STATE {// Random reinitialization}
			\STATE $\vec{x}_{i} = $ \CALL{RANDOM}{} ;~~~ $i = 1, 2, ..., \mathit{Np}$
			\STATE {$ \{\vec{x}_{b}^{l},e_{b}^{l}\} = \{\vec{x}_{b}^{p},e_{b}^{p}\} = $ \CALL{BEST}{$P$}}
		\ELSE
 			\STATE {// Component reinitialization}
			\STATE $\vec{x}_{i} = $ \CALL{RANDOM}{$\vec{x}_b^l,C$};~~~ $i = 1, 2, ..., \mathit{Np}$ \label{alg:componet}
			\STATE {$ \{\vec{x}_{b}^{p},e_{b}^{p}\} = $ \CALL{BEST}{$P$}}
		\ENDIF
	\ENDIF
\ENDPROCEDURE
\end{algorithmic}
\caption{The reinitialization mechanism.}
\label{alg:reinitialization}
\end{figure}
\begin{itemize}
    \item[$\bullet$] Locate the best individual by using the current population,
    \item[$\bullet$] Locate the best similar individual by using component reinitialization, and
    \item[$\bullet$] Locate the global best individual by using random reinitialization.
\end{itemize}

\begin{table*}[t!]
\centering
 \caption{Details of amino acid sequences used in experiments.}
 \label{tab:sequences}
 \scalebox{0.90}{
 \begin{tabular}{rrrp{15cm}}
 Label & Length & $D$ & Sequence \\
 \hline
 1BXP & 13 &  21 & ABBBBBBABBBAB \\ 
 1CB3 & 13 &  21 & BABBBAABBAAAB \\
 1BXL & 16 &  27 & ABAABBAAAAABBABB \\
 1EDP & 17 &  29 & ABABBAABBBAABBABA \\
 2ZNF & 18 &  31 & ABABBAABBABAABBABA \\
 1EDN & 21 &  37 & ABABBAABBBAABBABABAAB \\
 2H3S & 25 &  45 & AABBAABBBBBABBBABAABBBBBB \\
 1ARE & 29 &  53 & BBBAABAABBABABBBAABBBBBBBBBBB \\
 2KGU & 34 &  63 & ABAABBAABABBABAABAABABABABABAAABBB\\
 1TZ4 & 37 &  69 & BABBABBAABBAAABBAABBAABABBBABAABBBBBB \\
 1TZ5 & 37 &  69 & AAABAABAABBABABBAABBBBAABBBABAABBABBB \\
 1AGT & 38 &  71 & AAAABABABABABAABAABBAAABBABAABBBABABAB \\
 1CRN & 46 &  87 & BBAAABAAABBBBBAABAAABABAAAABBBAAAAAAAABAAABBAB \\
 2KAP & 60 & 115 & BBAABBABABABABBABABBBBABAABABAABBBBBBABBBAABAAABBABBABBAAAAB \\
 1HVV & 75 & 145 & BAABBABBBBBBAABABBBABBABBABABAAAAABBBABAABBABBBABBAABBABBAABBBB \\
      &    &     & BAABBBBBABBB \\
 1GK4 & 84 & 163 & ABABAABABBBBABBBABBABBBBAABAABBBBBAABABBBABBABBBAABBABBBBBAABAB \\
      &    &     & AAABABAABBBBAABABBBBA \\
 1PCH & 88 & 171 & ABBBAAABBBAAABABAABAAABBABBBBBABAAABBBBABABBAABAAAAAABBABBABABA \\
      &    &     & BABBABBAABAABBBAABBAAABA \\
 2EWH & 98 & 191 & AABABAAAAAAABBBAAAAAABAABAABBAABABAAABBBAAAABABAAABABBAAABAAABA \\ 
      &    &     & AABAABBAABAAAAABAAABABBBABBAAABAABA \\
 \hdashline
 F13  & 13 &  21 & ABBABBABABBAB \\
 F21  & 21 &  37 & BABABBABABBABBABABBAB \\
 F34  & 34 &  63 & ABBABBABABBABBABABBABABBABBABABBAB \\
 F55  & 55 & 105 & BABABBABABBABBABABBABABBABBABABBABBABABBABABBABBABABBAB \\
 F89  & 89 & 173 & ABBABBABABBABBABABBABABBABBABABBABBABABBABABBABBABABBABABBABBAB \\
      &    &     & ABBABBABABBABABBABBABABBAB \\
 \end{tabular}
 }
\end{table*}

\noindent
The following new control parameters are introduced to the reinitialization mechanism: $P_b, L_b$, and $C$. In our
previous work, the reinitialization was defined with a constant number of evaluations ($10^5$). In this work, restarts are dependent
on the values of parameters $P_b$, $L_b$ and the dimension of the problem. The $P_b$ defines how long the best population individual
can stay unchanged within the evolutionary process (line~2). For example, with $P_b=100$ and dimension $21$, the reinitialization is 
performed if the best population individual would stay unchanged in the evolutionary process for at
least $P_b \cdot D = 100 \cdot 21 = 2100$ evaluations. When this condition is satisfied, the algorithm performs random or component 
reinitialization according to the parameter $L_b$. If the local best individual is not changed for $L_b \cdot D$ reinitializations, 
then random reinitialization is performed, and component reinitialization otherwise (line~4). The last parameter $C$ determines the 
number of components that are different between the local best individual and individuals generated by component reinitialization 
(line~10). Within the component reinitialization, the $C$ components of each population individual $\vec{x}_i$ are 
selected randomly, and their values are replaced with random values on the interval $[-\pi,\pi]$, while all the remaining components 
get the values from the local best individual.	

\section{Experiments}
\label{sec:experiments}
The \DElscr\ algorithm was compiled with a GNU C++ compiler 4.6.3 and executed using an Intel Core i7
computer with 2.93 GHz CPU and 8 GB RAM under Linux Mint 13 Maya and a grid environment (Slovenian Initiative for
National Grid\footnote{Available at \url{http://www.sling.si/sling/}}). In order to evaluate the efficiency of the 
proposed algorithm, we used a set of amino acid sequences as shown in Table~\ref{tab:sequences}. This set includes 5
Fibonacci sequences and 18 real peptide sequences from the Protein Data Bank database\footnote{Available at
\url{https://www.rcsb.org/pdb/home/home.do}}. The $\text{K-D}$ method is used to transform real peptide sequences to the AB sequences.
In this method, the amino acids I, V, P, L, C, M, A, and G are transformed to hydrophobic ones (A) and amino acids D, E, H, F,
K, N, Q, R, S, T, W, and Y to hydrophilic ones (B). The selected sequences have different lengths, which enabled
us to analyze the algorithm according to different problem dimensions and, because they were used frequently in literature, 
they enabled us to compare the proposed algorithm with different algorithms. In order to analyze the efficiency of the introduced 
mechanisms and algorithm, we measured the following statistics:
\begin{itemize}
 \item[$\bullet$] The mean obtained energy value for all runs:
    \begin{equation}
    E_{\mathit{mean}} = \frac{\sum_{i=1}^{N}E_i}{N} \nonumber
    \end{equation}
    where $E_i$ denotes the obtained energy value for the $i$-th run and $N$ is the number of runs.
 \item[$\bullet$] The best obtained energy value among all runs:
    \begin{equation}
    E_{\mathit{best}} = max\{E_1, E_2, ..., E_N\}. \nonumber
    \end{equation} 
    Note that all energy values within our experiments are multiplied by $-1$, which means that all energy values are positive and higher values are better.
 \item[$\bullet$] The standard deviation of energy values for all runs:  
    \begin{equation}
    E_{\mathit{std}} = \sqrt{\frac{\sum_{i=1}^{N}{(E_i-E_{\mathit{mean}})}^2}{N-1}} \nonumber
    \end{equation}
 \item[$\bullet$] The hit ratio or percentage of runs during which the best solution has equal or better energy value according to the target value ($target$):
    \begin{equation}
    \mathit{hit_r} = \frac{N_h}{N} \nonumber
    \end{equation}
    where $N_h$ denotes the number of runs where the best obtained solution has good enough energy value $e_b$ according to the target value ($e_b \ge \mathit{target}$).
 \item[$\bullet$] The mean number of solution evaluation for all $N_h$ runs:
    \begin{equation}
     \mathit{NSE}_{\mathit{mean}} = \frac{\sum_{i=1}^{N_h}\mathit{NSE}_{i}}{N_h} \nonumber
    \end{equation}
    where $\mathit{NSE}_{i}$ represents the number of solution evaluations for the $i$-th run.
 \item[$\bullet$] The standard deviation of solution evaluations for all $N_h$ runs:
    \begin{equation}
    \mathit{NSE}_{\mathit{std}} = \sqrt{\frac{\sum_{i=1}^{N_h}{(\mathit{NSE}_i-\mathit{NSE}_{\mathit{mean}})}^2}{N_h-1}} \nonumber     
    \end{equation}
 \item[$\bullet$] The mean runtime for all runs:
    \begin{equation}
     \mathit{t}_{mean} = \frac{\sum_{i=1}^{N}\mathit{t_i}}{N} \nonumber
    \end{equation}
    where $\mathit{t_i}$ represents the runtime of the $i$-th run.
 \item[$\bullet$] The mean speed for all runs:
    \begin{equation}
     \mathit{v}_{mean} = \frac{\sum_{i=1}^{N}\mathit{v_i}}{N} \nonumber
    \end{equation}
    where $\mathit{v_i}$ represents the speed (the number of solution evaluations per second) of the $i$-th run.
\end{itemize}
The listed statistics were measured within the context of the following stopping conditions:
\begin{itemize}
 \item[$\bullet$] The maximum number of solution evaluations\\$\mathit{NSE_{\mathit{lmt}}}$: $\mathit{NSE}_i \ge \mathit{NSE}_{lmt}$.
 \item[$\bullet$] The runtime limit $\mathit{t_{lmt}}$: $t_i \ge t_{lmt}$.
 \item[$\bullet$] The energy value of the best obtained solution\\$\mathit{target}$: $e_{b} \ge \mathit{target}$.
\end{itemize}
Our algorithm belongs to stochastic algorithms, therefore, all the reported results of the proposed algorithm within this work are
based on $N=100$ independent runs. The described statistics, the defined stopping criteria and the determined number of independent runs were used to analyze the influence of new parameters and mechanisms on the algorithm's efficiency. The algorithm was also compared with the state-of-the art algorithms.
The asymptotic average-case performances were determined for the 6 shortest sequences, and an analysis of the obtained conformations
was also performed and will be given in the continuation of the paper.

\begin{table*}[t!]
\centering
\footnotesize
\caption{The analysis of the new control parameters ($P_b, L_b, C$). $N=100$ independent runs were performed for each setting and
the stopping conditions were the maximum number of solution evaluations $\mathit{NSE_{\mathit{lmt}}}=10^{10}$ and target value.}
\label{tab:settings}
\subfloat[F13, $target = 6.9961$]{
\label{tab:f13}
\begin{tabular}{@{}r@{~~}r@{~~}r@{~}|r@{~~}r@{~~}r@{~~}r@{~~}r@{}}
   $P_b$  & $L_b$ & $C$ & $E_{\mathit{mean}}$ & $E_{\mathit{std}}$ & $\mathit{hit_{r}}$ & $\mathit{NSE}_{\mathit{mean}}$ & $\mathit{NSE}_{\mathit{std}}$ \\
  \hline
        50  &       2  &      5  &       6.9961  &      0.00  &     100  &      6.41E+07  &      5.59E+07  \\
       100  &       2  &      5  &       6.9961  &      0.00  &     100  &      6.88E+07  &      5.96E+07  \\
        25  &       2  &      5  &       6.9961  &      0.00  &     100  &      7.52E+07  &      7.01E+07  \\
        50  &       5  &      5  &       6.9961  &      0.00  &     100  &      9.07E+07  &      8.06E+07  \\
        50  &       2  &     10  &       6.9961  &      0.00  &     100  &      9.52E+07  &      9.67E+07  \\
        50  &       1  &      5  &       6.9961  &      0.00  &     100  &      9.59E+07  &      9.55E+07  \\
   {\bf 50} & {\bf 10} & {\bf 5} &  {\bf 6.9961} & {\bf 0.00} &{\bf 100} & {\bf 9.77E+07} & {\bf 9.48E+07} \\
        50  &       2  &      2  &       6.9847  &      0.03  &      90  &      3.15E+09  &      2.67E+09  \\
  \end{tabular}
}
\hfil
\subfloat[F21, $target = 16.5544$]{
\label{tab:f21}
\begin{tabular}{@{}r@{~~}r@{~~}r@{~}|r@{~~}r@{~~}r@{~~}r@{~~}r@{}}
   $P_b$  & $L_b$ & $C$ & $E_{\mathit{mean}}$ & $E_{\mathit{std}}$ &$\mathit{hit_r}$ & $\mathit{NSE}_{\mathit{mean}}$ &  $\mathit{NSE}_{\mathit{std}}$\\
  \hline
        25  &      20  &      5  &      16.4500  &      0.09  &      21  &      4.91E+09  &      2.75E+09  \\
        25  &      10  &      5  &      16.4492  &      0.08  &      19  &      4.89E+09  &      2.81E+09  \\
   {\bf 50} & {\bf 10} & {\bf 5} & {\bf 16.4432} & {\bf 0.08} & {\bf 17} & {\bf 3.92E+09} & {\bf 2.98E+09} \\
        50  &      20  &      5  &      16.4415  &      0.10  &      22  &      4.86E+09  &      3.17E+09  \\
        10  &      20  &      5  &      16.4307  &      0.09  &      15  &      3.93E+09  &      2.51E+09  \\
        25  &      50  &      5  &      16.4254  &      0.12  &      33  &      4.84E+09  &      2.85E+09  \\
        25  &      20  &     10  &      16.4037  &      0.08  &       2  &      7.94E+09  &      2.53E+09  \\
        25  &      20  &      2  &      15.4393  &      0.45  &       0  &             -  &             -  \\
  \end{tabular}
}

\subfloat[F34]{
\label{tab:f34}
\begin{tabular}{rrr|rr}
   $P_b$  & $L_b$ & $C$ & $E_{\mathit{mean}}$ & $E_{\mathit{std}}$ \\
  \hline
    {\bf 50} & {\bf 10} & {\bf  5} & {\bf 30.0670} & {\bf 0.45} \\
         50  &      20  &       5  &      30.0596  &      0.40  \\
         25  &      10  &       5  &      30.0519  &      0.47  \\
         50  &       5  &       5  &      29.9108  &      0.38  \\
        100  &      10  &       5  &      29.9034  &      0.47  \\
         50  &      10  &      10  &      29.3722  &      0.35  \\
         50  &      10  &       2  &      24.2650  &      1.94  \\
             & \\
  \end{tabular}
}
\hfil
\subfloat[F55]{
\label{tab:f55}
\begin{tabular}{rrr|rr}
   $P_b$  & $L_b$ & $C$ & $E_{\mathit{mean}}$ & $E_{\mathit{std}}$\\
  \hline
   {\bf 25} & {\bf 5} & {\bf 10} & {\bf 49.0262} & {\bf 0.78} \\
        25 &      10  &      10  &      49.0233  &      1.26 \\
        10 &       5  &      10  &      49.0148  &      1.05 \\
        50 &       5  &      10  &      48.9379  &      1.19 \\
        25 &       2  &      10  &      48.9192  &      1.03 \\
        25 &       5  &       5  &      47.8458  &      1.74 \\
        25 &       5  &      20  &      47.4250  &      0.88 \\
           & \\
  \end{tabular}
}
\hfil
\subfloat[F89]{
\label{tab:f89}
\begin{tabular}{rrr|rr}
   $P_b$  & $L_b$ & $C$ & $E_{\mathit{mean}}$ & $E_{\mathit{std}}$\\
  \hline
         50  &      2  &      10  &      76.8608 &      1.64  \\
         25  &      2  &      10  &      76.6879 &      1.89  \\
         50  &      5  &      10  &      76.5090 &      1.88  \\
    {\bf 25} & {\bf 5} & {\bf 10} & {\bf 76.4541}& {\bf 1.93} \\
         50  &      1  &      10  &      76.3478 &      1.40  \\
        100  &      2  &      10  &      76.3275 &      1.71  \\
         50  &      2  &       5  &      75.1975 &      2.62  \\
         50  &      2  &      20  &      75.0143 &      1.52  \\
  \end{tabular}
}
\end{table*}

\subsection{Parameter settings}
The influence of the new control parameters ($P_b, L_b, C$) on the algorithm's efficiency was analyzed by using Fibonacci sequences.
In this analysis, the stopping condition was the maximum number of solution evaluations $\mathit{NSE_{\mathit{lmt}}} = 10^{10}$. 
For each sequence, we started with the following setting: $P_b=50$, $L_b=10$,
and $C=5$. Using 6 settings, where only one value of each setting was changed to the nearest higher or lower value, we tried to get
better settings. The parameter values are used from the following sets:
\begin{eqnarray}
 P_b &\in& \{10, 25, 50, 100\} \nonumber \\
 L_b &\in& \{1, 2, 5, 10, 20, 50\}  \nonumber \\
 C &\in& \{2, 5, 10, 20\} \nonumber
\end{eqnarray}
For the started setting the following 6 settings were used:
\begin{eqnarray}
&\{& \bm{P_b=10}, L_b=10, C=5\},\{\bm{P_b=25}, L_b=10, C=5\}, \nonumber \\
&\{& P_b=50, \bm{L_b=5}, C=5\},\{P_b=50, \bm{L_b=20}, C=5\},  \nonumber \\
&\{& P_b=50, L_b=10, \bm{C=2}\},\{P_b=50, L_b=10, \bm{C=10}\}. \nonumber
\end{eqnarray}
We repeated this process until a new better setting was found. The results of the least iterations, together with recommended settings, are shown in
Table~\ref{tab:settings}. For clarity, the recommended settings and their results are shown in bold typeface.
The displayed results show that each sequence has its own optimal setting, but it is still possible to select settings that
can be used for any unknown sequence. We define these settings according to the dimension of the problem, as follows:
\begin{eqnarray}
  \{P_b,L_b,C\} = 
	\begin{cases}
		\{50, 10, 5\} & \text{if } n<45 \\
		\{25, 5, 10\} & \text{otherwise}.\\
	\end{cases} \nonumber
\end{eqnarray}
These settings are used in all the following experiments, because they can provide a good hit ratio for short sequences and good energy
values for longer sequences. The search space for longer sequences is huge, which means the algorithm almost surely cannot reach optimal solutions in
a reasonable runtime e.g. 4 days. Therefore, for these sequences, the algorithm has to perform more reinitializations, and the component reinitialization
has to change more components randomly.

The displayed results confirm, additionally, that the variable $\mathit{NSE}$ have near-exponential or near-geometric distribution
($\mathit{NSE}_{\mathit{mean}} \approx \mathit{NSE}_{\mathit{std}}$). Under such distributions, given the $N_h = 100$ runs in
all of our experiments, a reliable rule-of-thumb estimates a 95\% confidence interval: 
\begin{eqnarray}
\mathit{CI}_{95} &\approx& [(1 - \frac{1.96}{\sqrt{N_h}})\cdot\mathit{NSE}_{\mathit{mean}},(1+\frac{1.96}{\sqrt{N_h}})\cdot\mathit{NSE}_{\mathit{mean}}] \nonumber \\
&\approx& [0.8\cdot\mathit{NSE}_{\mathit{mean}},1.2\cdot\mathit{NSE}_{\mathit{mean}}]. \nonumber
\end{eqnarray}

\subsection{Local search}
\begin{table*}[t!]
\centering
\footnotesize
\caption{The influence of the local search to the algorithm's efficiency according to two algorithms: \DElscr\ - with local search
and \DEcr\ - without local search. Two different comparisons were made according to two different stopping conditions: 
$\mathit{NSE}_{\mathit{lmt}}=10^{7}$ and $t_\mathit{lmt}=t_\mathit{mean}($\DEcr$)$. The reported mean speed $v_{\mathit{mean}}$ 
represents the mean number of function evaluations per second, $c_v$ represents the speed up factor 
$v_{\mathit{mean}}$(\DElscr)$/v_{\mathit{mean}}$(\DEcr) and the mean runtime $t_{\mathit{mean}}$ is given in seconds.}
\label{tab:localSearch}
\scalebox{0.9}{
\begin{tabular}{@{~~}r@{~~} | r@{~~~}r@{~~}r@{~~}r@{~~} | r@{~~~}r@{~~}r@{~~}r@{~~} | r@{~~~}rr@{~~}r@{~~}r@{~~}}
  \multirow{3}{*}{Label} & \multicolumn{8}{c|}{$\mathit{NSE}_{\mathit{lmt}}=10^{7}$} & \multicolumn{5}{c}{$t_\mathit{lmt}=t_\mathit{mean}($\DEcr$)$} \\
        & \multicolumn{4}{c|}{\DElscr} & \multicolumn{4}{|c|}{\DEcr} & \multicolumn{5}{c}{\DElscr} \\
        & \multicolumn{1}{c}{$t_\mathit{mean}$} & \multicolumn{1}{c}{$v_\mathit{mean}$} & \multicolumn{1}{c}{$E_{\mathit{mean}}$} & \multicolumn{1}{c|}{$E_{\mathit{std}}$} &  
          \multicolumn{1}{c}{$t_\mathit{mean}$} & \multicolumn{1}{c}{$v_\mathit{mean}$} & \multicolumn{1}{c}{$E_{\mathit{mean}}$} & \multicolumn{1}{c|}{$E_{\mathit{std}}$} & 
          \multicolumn{1}{c}{$t_\mathit{mean}$} & \multicolumn{1}{c}{$v_\mathit{mean}$} & \multicolumn{1}{c}{$c_{v}$} & \multicolumn{1}{c}{$E_{\mathit{mean}}$} & \multicolumn{1}{c}{$E_{\mathit{std}}$} \\
  \hline
  1BXP & 12.54  & 7.98E+05 &   4.7280 &  0.24 &  18.35 & 5.45E+05 &      4.6772  &  0.26 &  18.35 & 7.98E+05 & 1.46  &   {\bf 4.7831} &  0.26 \\ 
  1CB3 & 12.18  & 8.22E+05 &   7.9643 &  1.02 &  18.00 & 5.56E+05 & {\bf 8.1511} &  0.65 &  18.00 & 8.22E+05 & 1.48  &        8.1340  &  0.92 \\ 
  1BXL & 15.02  & 6.67E+05 &  16.2149 &  0.51 &  24.81 & 4.03E+05 &     16.2338  &  0.66 &  24.81 & 6.72E+05 & 1.66  &  {\bf 16.3452} &  0.48 \\ 
  1EDP & 16.00  & 6.25E+05 &  13.7281 &  1.07 &  27.39 & 3.65E+05 &     13.3930  &  1.80 &  27.39 & 6.07E+05 & 1.66  &  {\bf 14.1388} &  0.54 \\ 
  2ZNF & 17.44  & 5.74E+05 &  16.1670 &  1.86 &  29.78 & 3.36E+05 &     14.4350  &  3.01 &  29.78 & 5.73E+05 & 1.70  &  {\bf 16.7171} &  1.24 \\ 
  1EDN & 20.27  & 4.94E+05 &  17.9565 &  2.51 &  38.05 & 2.63E+05 &     16.5951  &  3.10 &  38.05 & 4.97E+05 & 1.89  &  {\bf 18.9328} &  1.85 \\ 
  2H3S & 24.85  & 4.03E+05 &  15.1685 &  2.36 &  50.64 & 1.98E+05 &     15.1545  &  2.79 &  50.64 & 4.06E+05 & 2.06  &  {\bf 16.1873} &  2.34 \\ 
  1ARE & 29.37  & 3.41E+05 &  19.3024 &  1.92 &  63.85 & 1.57E+05 &     18.6434  &  2.48 &  63.85 & 3.44E+05 & 2.20  &  {\bf 20.2815} &  1.80 \\ 
  2KGU & 34.89  & 2.87E+05 &  43.6622 &  3.46 &  84.42 & 1.19E+05 &     40.6789  &  4.66 &  84.42 & 2.92E+05 & 2.46  &  {\bf 46.1607} &  2.28 \\ 
  1TZ4 & 37.47  & 2.68E+05 &  28.7054 &  4.77 &  95.07 & 1.05E+05 &     25.3528  &  5.30 &  95.07 & 2.74E+05 & 2.61  &  {\bf 31.7309} &  4.26 \\ 
  1TZ5 & 36.66  & 2.73E+05 &  34.3793 &  4.36 &  95.11 & 1.05E+05 &     32.9423  &  5.08 &  95.11 & 2.81E+05 & 2.67  &  {\bf 36.7141} &  4.66 \\ 
  1AGT & 38.81  & 2.58E+05 &  52.9353 &  4.99 & 100.39 & 9.97E+04 &     47.6395  &  5.86 & 100.39 & 2.66E+05 & 2.67  &  {\bf 56.5688} &  3.80 \\ 
  1CRN & 51.66  & 1.94E+05 &  78.8070 &  4.74 & 139.22 & 7.19E+04 &     78.8601  &  5.69 & 139.22 & 2.02E+05 & 2.82  &  {\bf 82.8484} &  2.14 \\ 
  2KAP & 76.79  & 1.31E+05 &  54.6804 &  6.60 & 220.23 & 4.54E+04 &     55.3668  &  6.08 & 220.23 & 1.37E+05 & 3.01  &  {\bf 59.9772} &  5.93 \\ 
  1HVV & 104.64 & 9.62E+04 &  57.5815 &  6.55 & 320.20 & 3.12E+04 &     57.3717  &  6.56 & 320.20 & 1.03E+05 & 3.29  &  {\bf 63.5027} &  6.29 \\ 
  1GK4 & 122.54 & 8.23E+04 &  72.3524 &  7.15 & 394.79 & 2.53E+04 &     72.8575  &  7.87 & 394.79 & 9.10E+04 & 3.59  &  {\bf 78.8594} &  6.17 \\ 
  1PCH & 130.19 & 7.76E+04 & 103.4913 & 13.35 & 438.63 & 2.28E+04 &    101.7607  & 11.42 & 438.63 & 8.71E+04 & 3.82  & {\bf 116.0248} & 11.99 \\ 
  2EWH & 162.03 & 6.24E+04 & 189.5316 & 14.33 & 546.97 & 1.83E+04 &    182.2880  & 16.48 & 546.97 & 7.14E+04 & 3.91  & {\bf 205.3507} & 10.62 \\ 
  \hdashline                                                                                                                
  F13 & 12.42   & 8.05E+05 &   6.0591 &  0.94 &  18.08 & 5.53E+05 &      6.0951  &  1.09 &  18.08 & 8.07E+05 & 1.46  &   {\bf 6.3373} &  0.83 \\ 
  F21 & 20.90   & 4.79E+05 &  12.0495 &  1.66 &  37.84 & 2.64E+05 &     11.3298  &  1.75 &  37.84 & 4.82E+05 & 1.82  &  {\bf 12.8967} &  1.70 \\ 
  F34 & 35.89   & 2.79E+05 &  20.0652 &  2.59 &  82.90 & 1.21E+05 &     18.3942  &  2.29 &  82.90 & 2.87E+05 & 2.38  &  {\bf 21.4521} &  2.62 \\ 
  F55 & 71.27   & 1.41E+05 &  35.2611 &  3.55 & 191.88 & 5.21E+04 &     35.3539  &  3.02 & 191.88 & 1.46E+05 & 2.80  &  {\bf 37.9751} &  2.44 \\ 
  F89 & 140.58  & 7.19E+04 &  55.4025 &  5.06 & 455.38 & 2.20E+04 &     55.0803  &  5.55 & 455.38 & 7.92E+04 & 3.61  &  {\bf 59.1793} &  4.40 \\ 
\end{tabular}
}
\end{table*}

\begin{table*}[t!]
\centering
\caption{The influence of the component reinitialization to the algorithm's efficiency according to two algorithms: \DElscr\ - with component
reinitialization and \DEls\ - without component reinitialization. Two stopping conditions were used: $\mathit{NSE}_{\mathit{lmt}}=10^{11}$
and $target$ values that were set to the best-known energy values (see Table \ref{tab:energy}). The shown $C_{\mathit{NSE}}$ represents the
reduction of $\mathit{NSE}_{\mathit{mean}}$: $\mathit{NSE}_{\mathit{mean}}$(\DElscr) $/$  $\mathit{NSE}_{\mathit{mean}}$(\DEls).}
\label{tab:componentReinitilaziation}
\begin{tabular}{rrrr|rrrr|rrr}
  \multirow{2}{*}{Label} & \multirow{2}{*}{Length} & \multirow{2}{*}{$D$} & \multirow{2}{*}{$target$} & \multicolumn{4}{c|}{\DElscr} & \multicolumn{3}{c}{\DEls} \\
  & & & & $\mathit{NSE}_{\mathit{mean}}$ & $C_{\mathit{NSE}}$& $\mathit{NSE}_{\mathit{std}}$ & $hit_r$ & $\mathit{NSE}_{\mathit{mean}}$ & $\mathit{NSE}_{\mathit{std}}$ & $hit_r$ \\ 
  \hline
  1BXP & 13 & 21 &  5.6104 & {\bf 1.56E+09} &  ~~ 0.38 & 1.68E+09 & {\bf 100} & 4.07E+09 & 4.20E+09 & {\bf 100} \\ 
  1CB3 & 13 & 21 &  8.4589 & {\bf 3.61E+07} &  ~~ 0.18 & 4.26E+07 & {\bf 100} & 1.99E+08 & 1.77E+08 & {\bf 100} \\ 
  1BXL & 16 & 27 & 17.3962 & {\bf 1.24E+10} & $< 0.37$ & 1.24E+10 & {\bf 100} & 3.36E+10 & 2.71E+10 &       32  \\ 
  1EDP & 17 & 29 & 15.0092 & {\bf 4.58E+09} & $< 0.22$ & 4.21E+09 & {\bf 100} & 2.14E+10 & 2.10E+10 &       96  \\ 
  2ZNF & 18 & 31 & 18.3402 & {\bf 2.10E+09} & $< 0.05$ & 1.92E+09 & {\bf 100} & 4.31E+10 & 2.75E+10 &       50  \\ 
   F13 & 13 & 21 &  6.9961 & {\bf 8.92E+07} &  ~~ 0.08 & 8.52E+07 & {\bf 100} & 1.14E+09 & 1.27E+09 & {\bf 100} \\ 
\end{tabular}
\end{table*}

The local search within our algorithm was designed to increase the speed of algorithm convergence and speed of neighborhood solution evaluations. In order to
demonstrate these advantages, the algorithm was analyzed with (\DElscr) and without (\DEcr) local search. Within this analysis, the algorithms
were compared using the following stopping conditions: $\mathit{NSE}_{\mathit{lmt}} = 10^7$ and $t_{\mathit{lmt}} = t_{\mathit{mean}}$(\DEcr),
as shown in Table \ref{tab:localSearch}. With the first stopping condition, we show that the local search improves mean energy value $E_{\mathit{mean}}$ on
16 out of 23 sequences and reduces the mean runtime $t_{\mathit{mean}}$ for all sequences. For the longest sequence 2EWH,
$E_{\mathit{mean}}$ was improved from 182.2880 to 189.5316, or by 7.2436, and for this improvement, the runtime was 
reduced from 546.97 to 162.03 seconds, or by factor 3.376.
Using the second stopping condition, both algorithms were limited with the same runtime and, in this case, local search improved $E_{\mathit{mean}}$ in 22 out
of 23 sequences. These values are marked in bold typeface within the table. The value 205.3507 was shown in bold typeface for the sequence 2EWH.
This means that, by using the local search and $t_{\mathit{lmt}} = 546.97$ seconds, $E_{\mathit{mean}}$ was improved by 23.0627.
Results also show that the speed up factor $c_{v}$ was 1.46 for the shortest sequences (F13 and 1BXP) and 3.91 for the
longest sequence 2EWH, while the $E_{\mathit{mean}}$ is worse by 0.0171 for sequence 1CB3 and better for all other sequences.

From the obtained results, we can conclude that the local search improves the convergence speed of the algorithm for most of the sequences,
while the speed of solution evaluation is increased for all sequences. Somebody would expect better speed up factors, but note that some conditions
must be satisfied for local search and, therefore, the speed up factor is dependent on the relationship between the number of solution evaluations inside and
outside the local search. However, using the local search, the algorithm is capable of obtaining better energy values for almost all sequences, and this
improvement of energy values increases for longer sequences.

\begin{figure*}[t!]
\centering
\subfloat[1BXP]{\makebox[.4\textwidth]{\includegraphics[scale=.4]{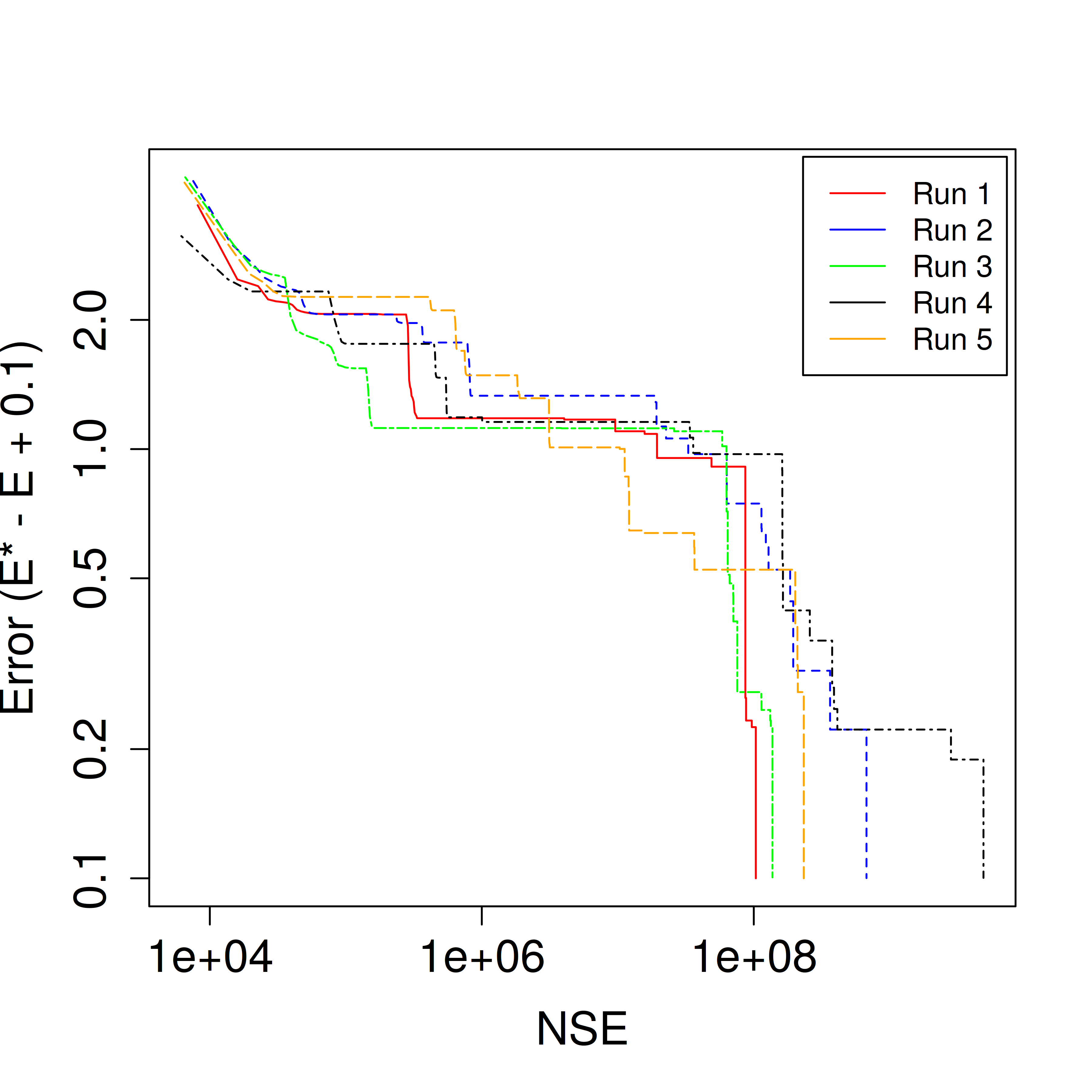}}}
\subfloat[1CB3]{\makebox[.4\textwidth]{\includegraphics[scale=.4]{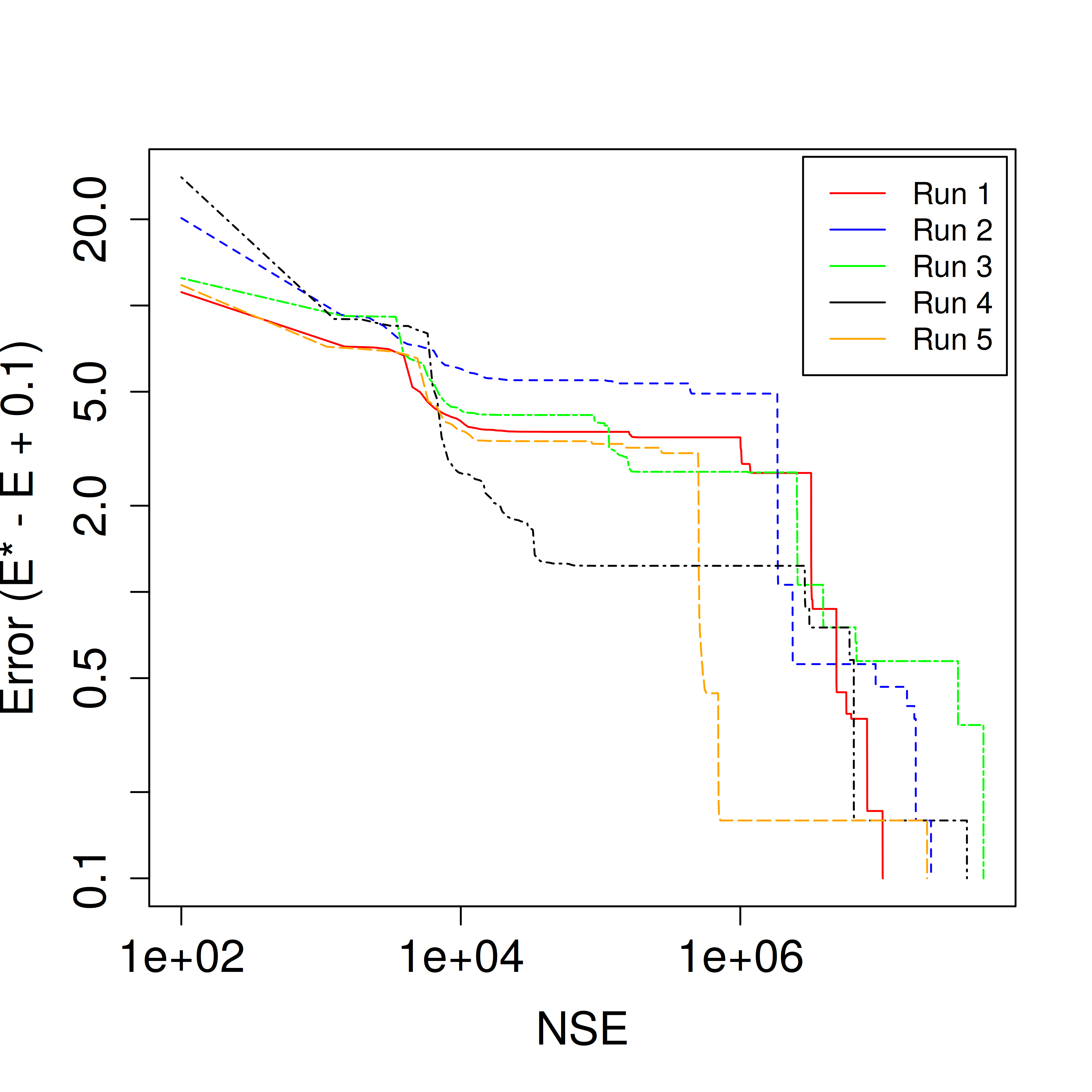}}}

\subfloat[1BXL]{\makebox[.4\textwidth]{\includegraphics[scale=.4]{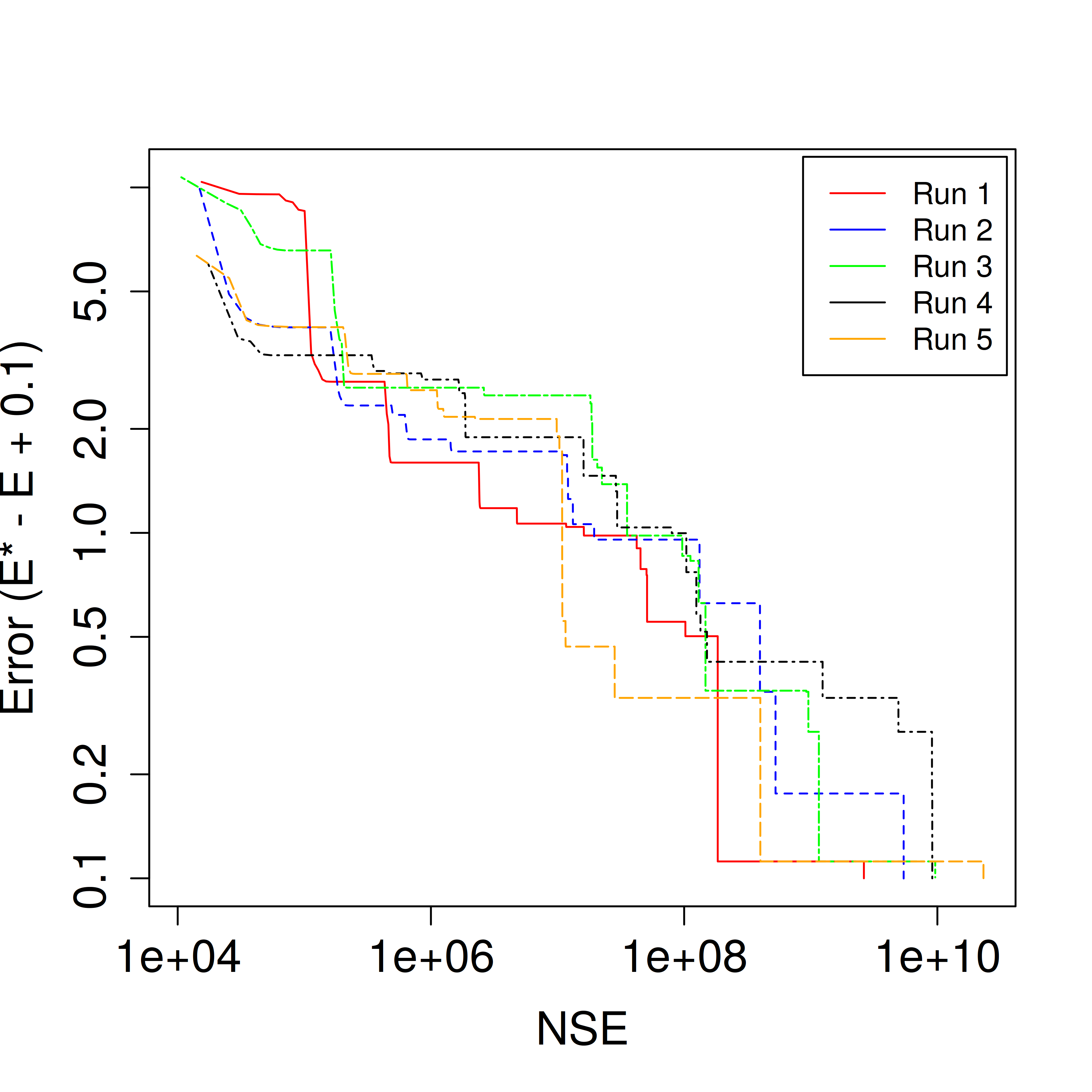}}}
\subfloat[1EDP]{\makebox[.4\textwidth]{\includegraphics[scale=.4]{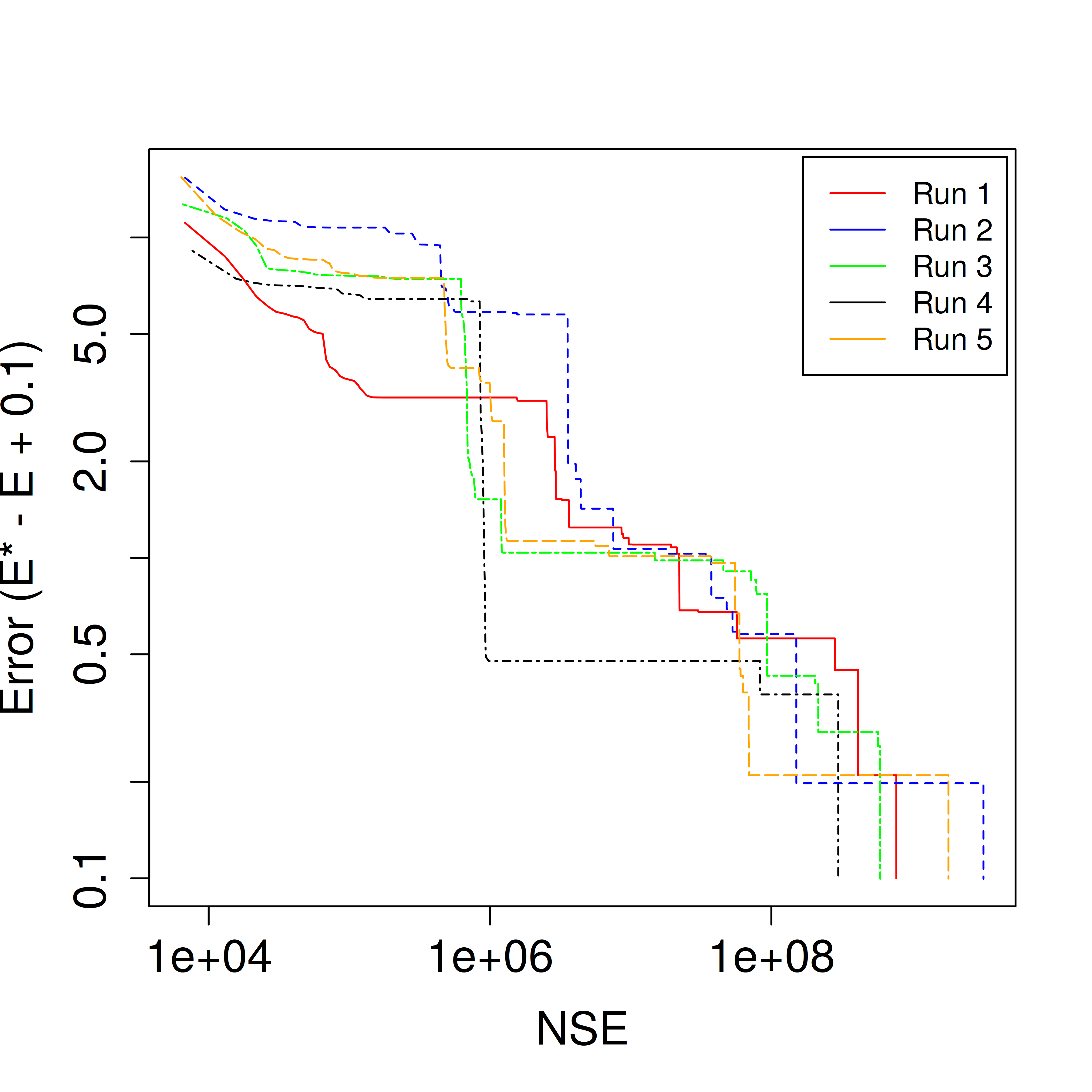}}}

\subfloat[2ZNF]{\makebox[.4\textwidth]{\includegraphics[scale=.4]{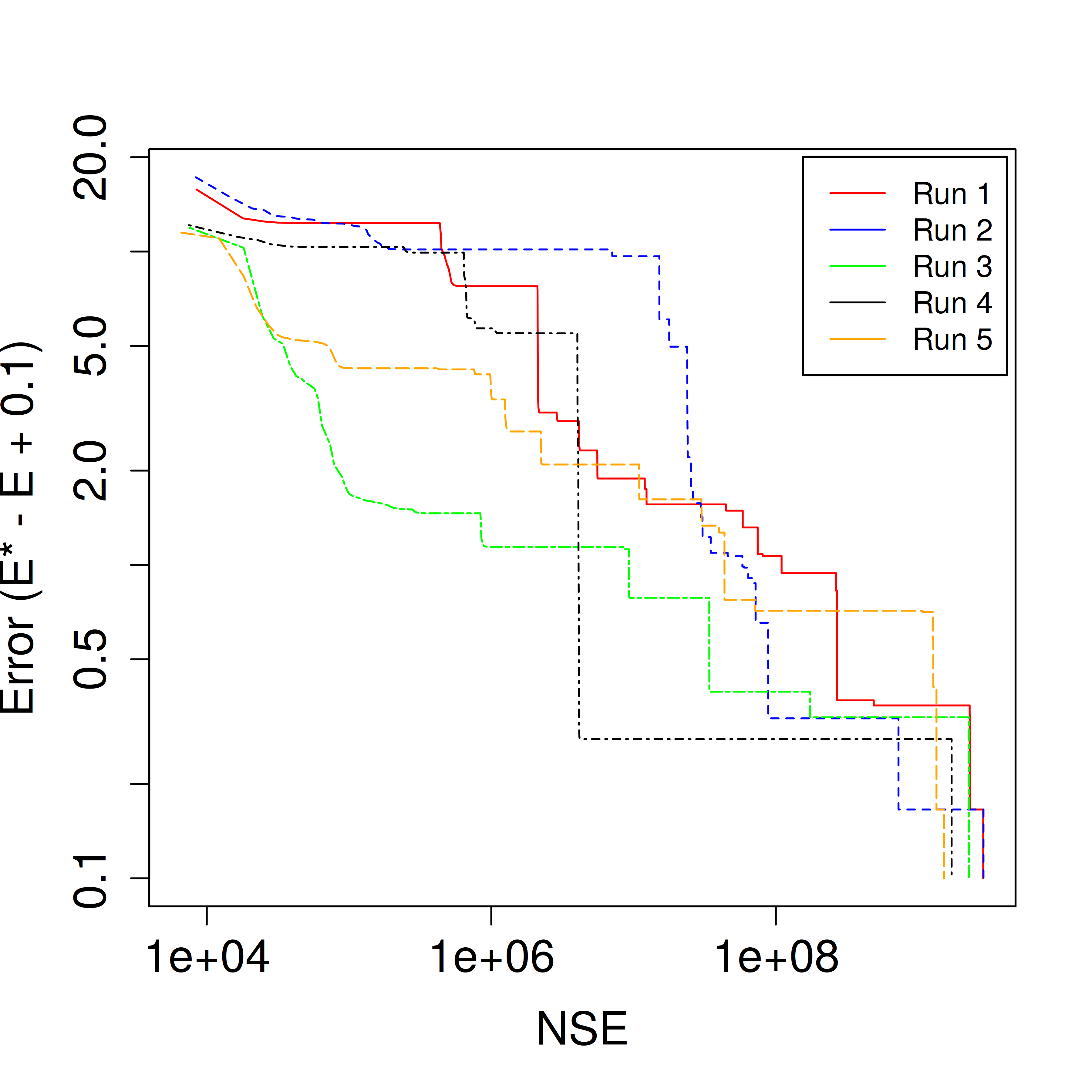}}}
\subfloat[F13]{\makebox[.4\textwidth]{\includegraphics[scale=.4]{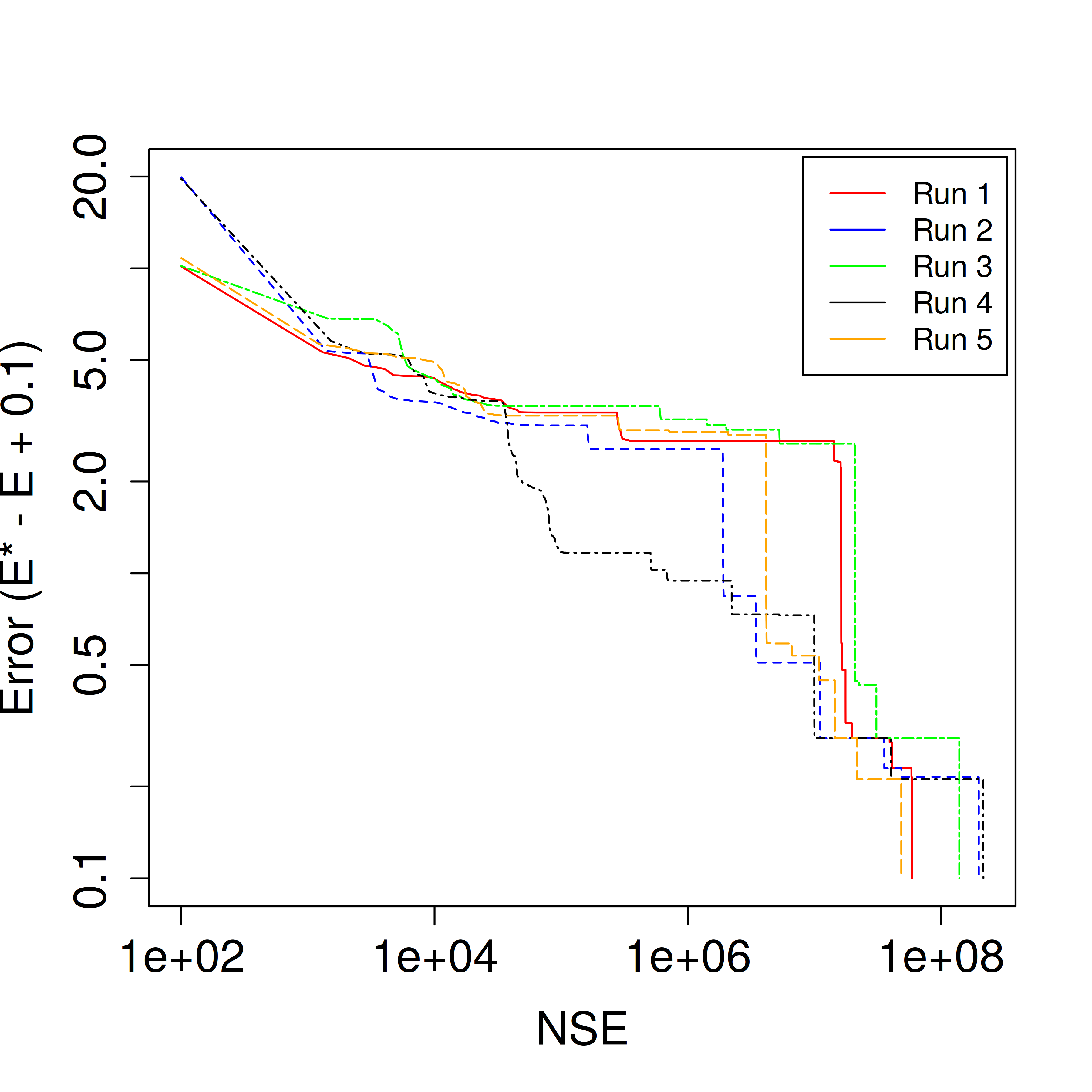}}}
\caption{The convergence graphs of the evolutionary process for 5 randomly chosen runs. The error represents the difference between the
best-known energy value ($E^{*}$) and the energy value ($E$) of the global best individual within the evolutionary process. A small value of 0.1 was added
to the error because of the logarithmic scale.}
\label{fig:convergence}
\end{figure*}

\subsection{Component reinitialization}
The main goal of the component reinitialization is to redirect the evolutionary process in such a way that a similar good solution can be located according to
the local best solution. To demonstrate the influence of this mechanism on the algorithm's efficiency, the algorithm was analyzed with (\DElscr) and without 
(\DEls) component reinitialization. Within this analysis, the algorithms were compared using the target values of the best-known energy values and 
${\mathit{NSE}}_{\mathit{lmt}}=10^{11}$, as shown in Table \ref{tab:componentReinitilaziation}. The best values of ${\mathit{NSE}}_{\mathit{mean}}$ and
$\mathit{hit}_r$ are marked in bold typeface. From the results, we can see that the algorithm that uses component reinitialization is capable of reaching
the best-known energy value in all runs, and for that it required significantly less solution evaluations (${\mathit{NSE}}$).
For example, ${\mathit{NSE}}_{\mathit{mean}}$ was reduced from $1.14$E$+09$ to $8.92$E$+07$ for sequence F13.
On the other hand, the algorithm without component reinitialization was not capable of reaching $\mathit{hit}_r=100$ within the budget of ${\mathit{NSE}}_{\mathit{lmt}}=10^{11}$ for sequences 1BXL, 1EDP and 2ZNF. From these observations, we can conclude that the proposed component
reinitialization allows the algorithm to locate good similar solutions and to
reach the best-known energy values. This is shown clearly in Table~\ref{tab:componentReinitilaziation} with $C_{\mathit{NSE}}$ which represents the
relationship between the obtained values of 
${\mathit{NSE}}_{\mathit{mean}}$ for both algorithms. The component reinitialization reduces the ${\mathit{NSE}}_{\mathit{mean}}$ from 0.38 
for sequence 1BXP to less than 0.05 for sequence 2ZNF.
Finally, it is obvious that the reinitialization mechanism is responsible for the \DElscr\ obtaining $\mathit{hit}_r=100$ for all the sequences shown in
Table \ref{tab:componentReinitilaziation}. The convergence curves of 5 randomly chosen runs per sequence are shown in Fig. \ref{fig:convergence}.
As we can see, the best-known energy value was reached in all the runs. It is shown with the energy error of 0.1. The energy error represents
the difference between the best-known energy value ($E^{*}$) and the energy value ($E$) of the global best individual. Note that, both axes are shown
on a logarithmic scale, therefore a small value of 0.1 was added to the energy error.

Results show additionally that the value of ${\mathit{NSE}}_{\mathit{mean}}$ is not only dependent on the sequence length or problem dimension, but
also on the sequence itself. For example, for \DElscr\ the value of ${\mathit{NSE}}_{\mathit{mean}}$ is 5.9 times smaller for sequence 2ZNF in comparison
with sequence 1BXL, although the dimension of the first sequence is greater than the dimension of the second sequence.

\begin{table*}
 \centering
 \caption{Asymptotic average-case performances for \DElscr. Values marked  with the * are obtained by using the grid environment. In these cases $t_{\mathit{mean}}$
 is calculated as follows: $t_{\mathit{mean}} = \frac{\mathit{NSE}_{\mathit{mean}}}{v_{\mathit{mean}}}$, where $v_{\mathit{mean}}$ represents the obtained mean speed 
 of three  independent runs on our test computer in a given period of time ($t_{\mathit{lmt}} =3600$ seconds). All other results are obtained on our test computer.}
 \label{tab:performances}
 \scalebox{0.9}{
 \begin{tabular}{rr|rrrrrrr|r}
  Label & & \multicolumn{1}{c}{1} & \multicolumn{1}{c}{2} & \multicolumn{1}{c}{3} & \multicolumn{1}{c}{4} & \multicolumn{1}{c}{5} & \multicolumn{1}{c}{6} & \multicolumn{1}{c|}{7} & \multicolumn{1}{c}{Asymptotic model} \\
  \hline
  \multirow{3}{*}{1BXP} & $\mathit{target}$              &   1.8013 &   2.0063 &   2.6838 &   3.1846 &   4.0191 ~ &   4.9321 ~ &   5.6104 ~ &                             \\
                        & $t_{\mathit{mean}}$            &     0.16 &    10.52 &   395.94 &  1349.36 &    12.39 ~ &   125.70 ~ &  1965.08 ~ & $0.0015 \cdot 2.8911^L$     \\
                        & $\mathit{NSE}_{\mathit{mean}}$ & 2.45E+05 & 1.38E+07 & 4.62E+08 & 1.42E+09 & 1.17E+07 ~ & 1.09E+08 ~ & 1.56E+09 ~ & $4589.5644 \cdot 2.5970^L$  \\
  \hline
  \multirow{3}{*}{1CB3} & $\mathit{target}$              &   1.9174 &   1.9786 &   2.3884 &   4.0429 &   6.0209 ~ &   8.4088 ~ &   8.4589 ~ &                             \\
                        & $t_{\mathit{mean}}$            &     0.06 &     1.01 &     1.67 &     4.22 &     4.70 ~ &    80.85 ~ &    44.47 ~ & $0.0001 \cdot 2.8662 ^ L$   \\
                        & $\mathit{NSE}_{\mathit{mean}}$ & 1.03E+05 & 1.41E+06 & 2.06E+06 & 4.82E+06 & 4.52E+06 ~ & 7.11E+07 ~ & 3.61E+07 ~ & $344.4917 \cdot 2.5502^L$   \\
  \hline
  \multirow{3}{*}{1BXL} & $\mathit{target}$              &  11.1862 &  13.8397 &  13.6386 &  14.0105 &  16.8991 ~ &  16.9404 ~ &  17.3962 ~ &                             \\
                        & $t_{\mathit{mean}}$            &   167.93 &  1472.79 &   611.86 &   508.47 & 25965.63 * & 113126.9 * & 39706.68 * & $0.0030 \cdot 2.7931 ^ L$   \\
                        & $\mathit{NSE}_{\mathit{mean}}$ & 8.96E+07 & 7.07E+08 & 2.68E+08 & 2.14E+08 & 9.93E+09 * & 4.05E+10 * & 1.24E+10 * & $6240.7027 \cdot 2.5976^L$  \\
  \hline
  \multirow{3}{*}{1EDP} & $\mathit{target}$              &   6.3823 &   8.9122 &   8.7042 &   9.1152 &  11.5309 ~ &  11.7522 ~ &  15.0092 ~ &                             \\
                        & $t_{\mathit{mean}}$            &   677.33 &   437.82 &  1280.42 &   776.01 &  3494.96 ~ &   922.22 ~ &  7272.60 ~ & $10.7641 \cdot 1.4097 ^ L$  \\
                        & $\mathit{NSE}_{\mathit{mean}}$ & 6.74E+08 & 3.88E+08 & 1.04E+09 & 5.97E+08 & 2.47E+09 ~ & 6.27E+08 ~ & 4.58E+09 ~ & 2.32E+07 $\cdot$ 1.3103$^L$ \\
  \hline
  \multirow{3}{*}{2ZNF} & $\mathit{target}$              &   9.3228 &  12.1166 &  11.9772 &  12.3307 &  14.6296 ~ &  14.6733 ~ &  18.3402 ~ &                             \\
                        & $t_{\mathit{mean}}$            &  4153.21 &   270.26 &   488.05 &   755.42 &   562.56 ~ & 41844.24 * &  7437.15 ~ & $16.6030 \cdot 1.3505^L$    \\
                        & $\mathit{NSE}_{\mathit{mean}}$ & 3.67E+09 & 2.18E+08 & 3.65E+08 & 5.50E+08 & 3.77E+08 ~ & 1.32E+10 * & 2.10E+09 ~ & 3.19E+07$ \cdot 1.2642^L$   \\
  \hline
  \multirow{3}{*}{F13}  & $\mathit{target}$              &   1.8225 &   2.0453 &   4.6082 &   4.6858 &   5.0428 ~ &   6.8092 ~ &   6.9961 ~ &                             \\
                        & $t_{\mathit{mean}}$            &     0.42 &     3.34 &     4.48 &     8.03 &    10.47 ~ &    70.98 ~ &   110.54 ~ & $0.0019 \cdot 2.3266 ^ L$   \\
                        & $\mathit{NSE}_{\mathit{mean}}$ & 6.62E+05 & 4.52E+06 & 5.20E+06 & 8.43E+06 & 9.74E+06 ~ & 6.19E+07 ~ & 8.92E+07 ~ & $6138.4492 \cdot 2.0846^L$  \\
 \end{tabular}
 }
\end{table*}

\begin{figure*}[t!]
\centering
\subfloat[1BXP]{\makebox[.4\textwidth]{\includegraphics[scale=.35]{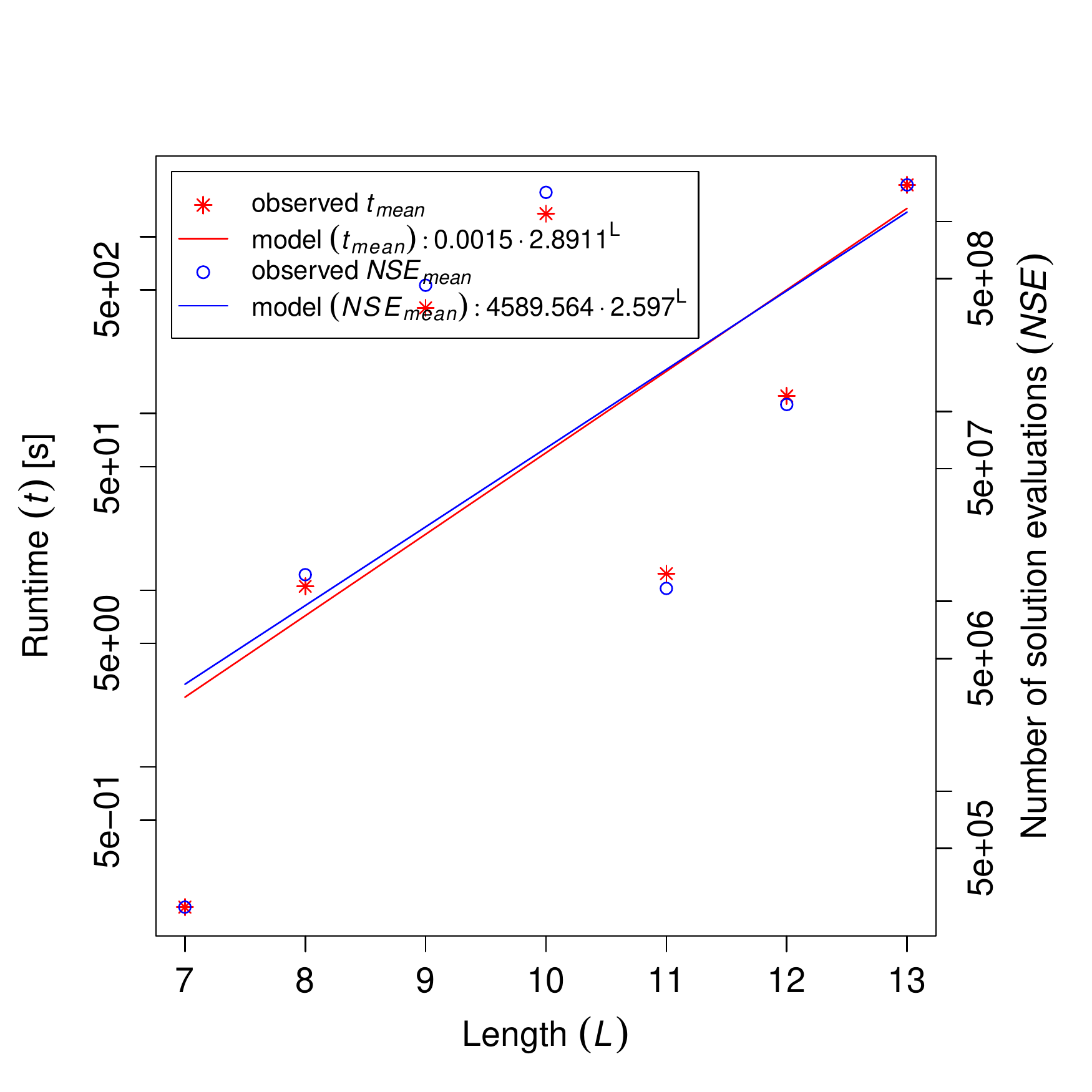}}}
\subfloat[1CB3]{\makebox[.4\textwidth]{\includegraphics[scale=.35]{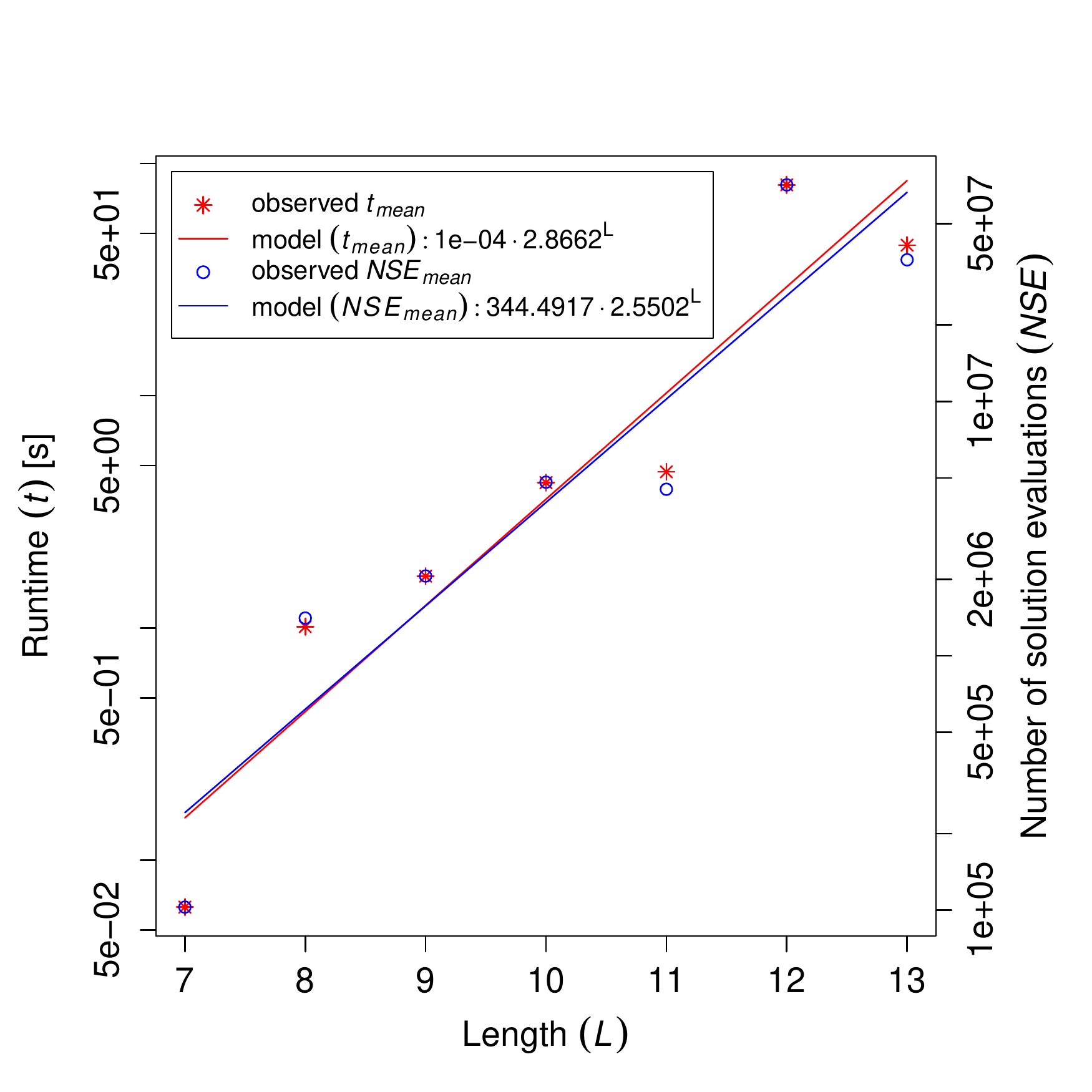}}}

\subfloat[1BXL]{\makebox[.4\textwidth]{\includegraphics[scale=.35]{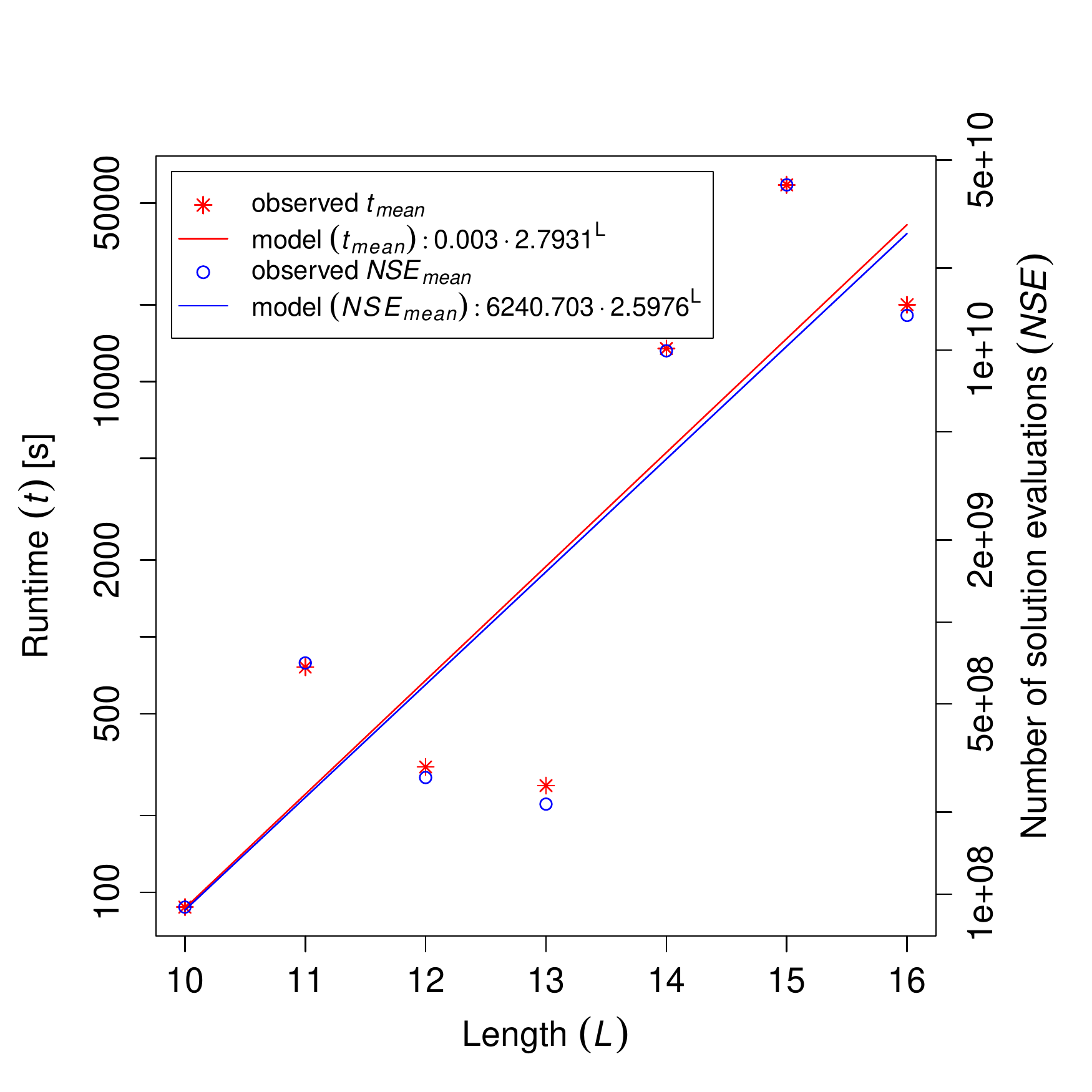}}}
\subfloat[1EDP]{\makebox[.4\textwidth]{\includegraphics[scale=.35]{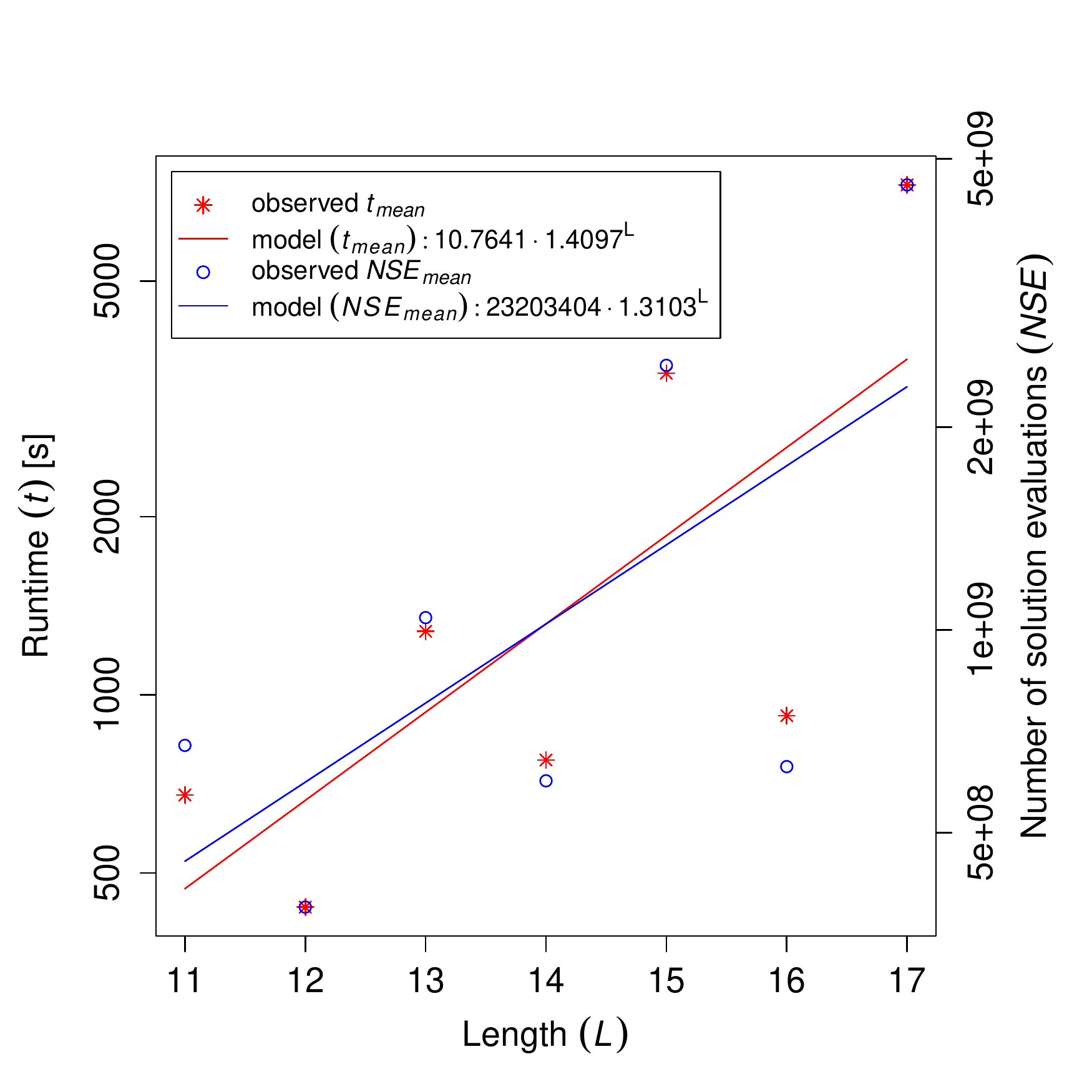}}}

\subfloat[2ZNF]{\makebox[.4\textwidth]{\includegraphics[scale=.35]{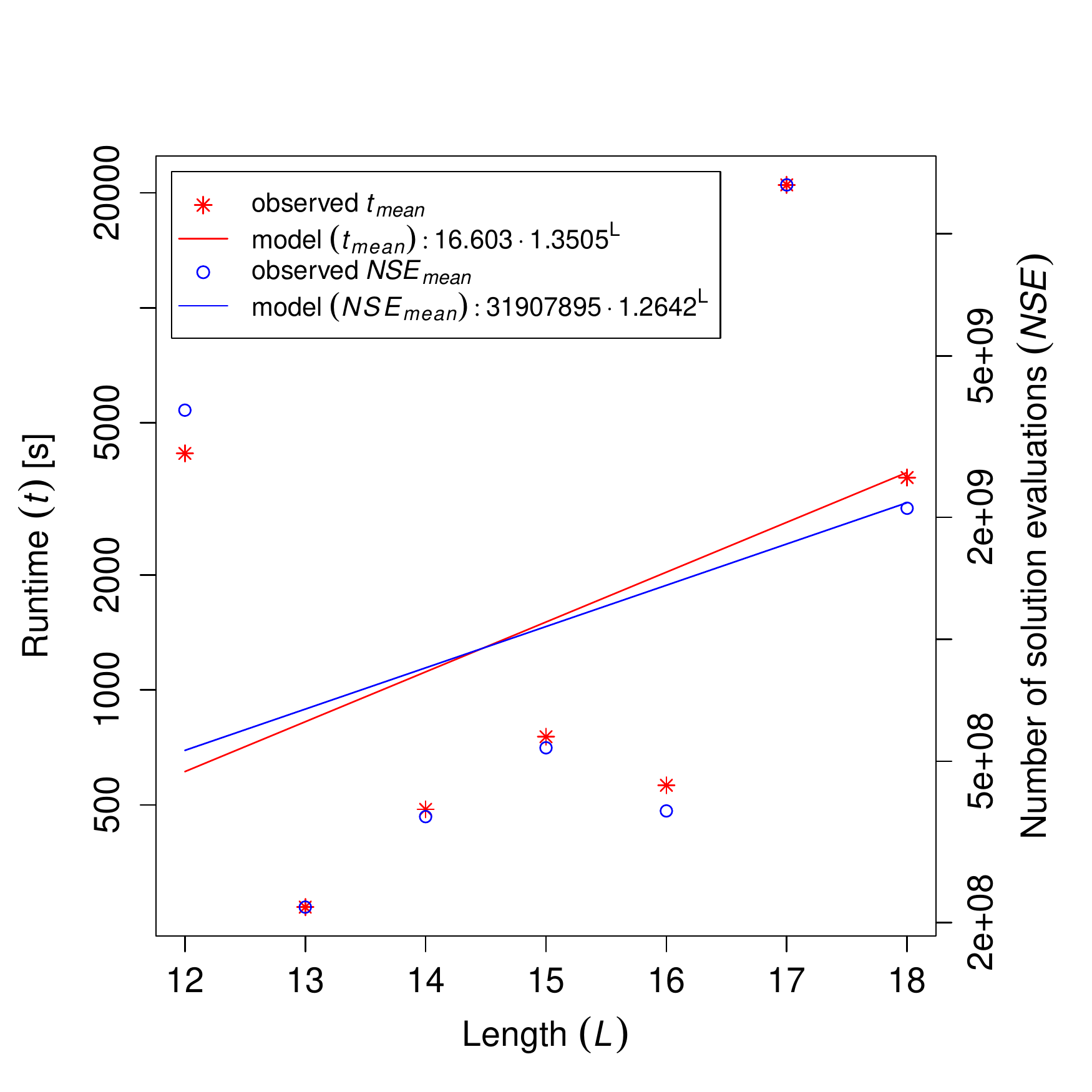}}}
\subfloat[F13]{\makebox[.4\textwidth]{\includegraphics[scale=.35]{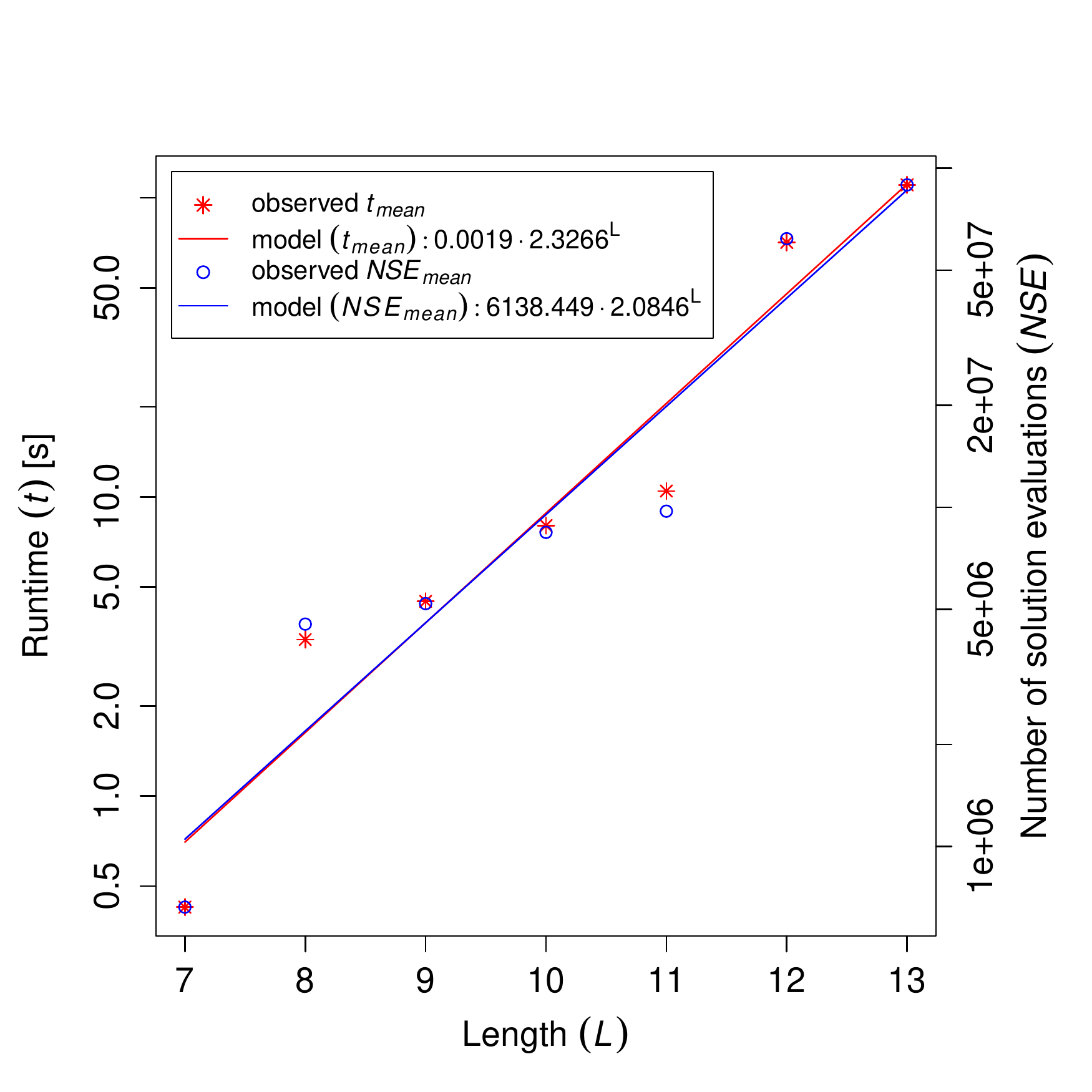}}}
\caption{Asymptotic average-case performances for \DElscr.}
\label{fig:performance}
\end{figure*}

\subsection{Asymptotic average-case performance}
In this section, we introduce an approach to determine asymptotic average-case performance of the algorithm for short sequences. The condition
for this is the ability of the algorithm to obtain the best-known solution with $\mathit{hit_r} = 100$. Until now, only our algorithm has been able to
fulfill this condition for the 6 shortest sequences. Six subsequences are generated for each of these sequences. The first subsequence has removed
the last monomer, the second subsequence has removed the last two monomers, etc. This means that the length of each next subsequence is decreased by 1.
For example, for sequence 1CB3 (BABBBAABBAAAB) the following six subsequences are generated:
\begin{itemize}
 \item [1.] BABBBAABBAAA ($L=12, \mathit{target}=8.4088$), 
 \item [2.] BABBBAABBAA ($L=11, \mathit{target}=6.0209$), 
 \item [3.] BABBBAABBA ($L=10, \mathit{target}=4.0429$),
 \item [4.] BABBBAABB ($L=9, \mathit{target}=2.3884$),
 \item [5.] BABBBAAB ($L=8, \mathit{target}=1.9786$),
 \item [6.] BABBBAA ($L=7, \mathit{target}=1.9174$).
\end{itemize}

We determined the best-known or target values for all subsequences. For this purpose, we performed one run for each
subsequence with $t_{\mathit {lmt}}=4$ days, and the best reached energy value is used as a target value. Using these target values as a stopping condition, 
it is possible to calculate asymptotic average-case performance. The original sequence is also included within this calculation. This means the asymptotic 
average-case performance is determined by using 7 sequences.

Table \ref{tab:performances} and Figure \ref{fig:performance} display the target values, obtained mean values and asymptotic average-case
performances for \DElscr. From the results, we observe that the best runtime asymptotic average-case performance was obtained for sequence 2ZNF 
($16.603 \cdot 1.3505^L$), while the worst for sequence 1BXP ($0.0015\cdot 2.8911^L$). Similarly, the best $\mathit{NSE}$ asymptotic average-case performance
was obtained for sequence 2ZNF ($3.19$E$+07\cdot 1.2642^L$), while the worst for sequence 1BXL ($6240.7027\cdot 2.5976^L$).
We can again observe that the value of $\mathit{NSE_{mean}}$ and $\mathit{t_{mean}}$ are not only dependent on the sequence length. Only one monomer can
influence these values significantly. For example, \DElscr\ requires less solution evaluations ($\mathit{NSE}$) and runtime~($\mathit{t}$) to reach the
target value for the subsequence of sequence 1EDP that has a length of 16 in comparison with a subsequence of length 13. From these results, we can
conclude that the structure of the sequence has a big influence on the difficulty of the problem.

\begin{table*}[t!]
\centering
\caption{Comparison of the \DElscr\ algorithm with state-of-the-art algorithms. Entries that are shown as `-' imply that no `best energy values'
have been reported in the literature.}
\label{tab:our:vs}
\scalebox{0.85}{
\begin{tabular}{rr|rrr|rrr|rrr|rrr}

  \multirow{3}{*}{Label} & \multirow{3}{*}{$M$} 
  & \multicolumn{9}{c|}{$\mathit{NSE_{lmt}} = M \cdot 10^4$} 
  & \multicolumn{3}{c}{$\mathit{NSE_{lmt}} = M \cdot 10^4 \cdot c_v$} \\
  
  & & \multicolumn{3}{c|}{\DElscr}
  & \multicolumn{3}{c|}{\DEpfo\ \cite{BoskovicAB16}} 
  & \multicolumn{3}{c|}{BE-ABC \cite{Li15,Li15f}} 
  & \multicolumn{3}{c}{\DElscr} \\
  
  & & \multicolumn{1}{c}{$E_{\mathit{best}}$} & \multicolumn{1}{c}{$E_{\mathit{mean}}$} & \multicolumn{1}{c|}{$E_{\mathit{std}}$}
    & \multicolumn{1}{c}{$E_{\mathit{best}}$} & \multicolumn{1}{c}{$E_{\mathit{mean}}$} & \multicolumn{1}{c|}{$E_{\mathit{std}}$}
    & \multicolumn{1}{c}{$E_{\mathit{best}}$} & \multicolumn{1}{c}{$E_{\mathit{mean}}$} & \multicolumn{1}{c|}{$E_{\mathit{std}}$}
    & \multicolumn{1}{c}{$E_{\mathit{best}}$} & \multicolumn{1}{c}{$E_{\mathit{mean}}$} & \multicolumn{1}{c}{$E_{\mathit{std}}$}  \\
 \hline
 1CB3 & 20 &   7.7450 &   4.5108 &  2.13 & {\bf  8.3690} &        5.5884  &  1.96 & --      & {\bf 5.9417} & 0.78 &        7.7450 &         4.5929  &  2.16 \\
 1BXL & 20 &  16.2618 &  12.5045 &  2.17 &      16.3443  &       12.6104  &  2.53 & --      &     11.6942  & 1.13 & {\bf  16.7137}&  {\bf  13.1940} &  2.22 \\
 1EDP & 20 &  13.1764 &   8.1986 &  2.78 & {\bf 13.5620} & {\bf   8.6666} &  2.56 & --      &      8.0500  & 0.93 &       13.1895 &         8.5313  &  2.81 \\
 2H3S & 20 &  17.1724 &  11.5310 &  2.45 &      16.5030  &       10.6767  &  2.75 & --      &     10.4618  & 1.13 & {\bf  17.4858}&  {\bf  11.9565} &  2.48 \\
 2KGU & 20 &  41.0221 &  33.6539 &  3.99 &      44.3369  &       35.3850  &  4.70 & --      &     22.7195  & 2.01 & {\bf  44.0110}&  {\bf  36.4642} &  4.39 \\
 1TZ4 & 20 &  34.5265 &  21.6863 &  3.62 &      30.9211  &       20.4361  &  5.28 & --      &     14.9436  & 2.22 & {\bf  35.3505}&  {\bf  24.9569} &  4.55 \\
 1TZ5 & 20 &  37.8896 &  25.9996 &  4.12 &      38.1868  &       27.3412  &  4.08 & --      &     17.4859  & 1.37 & {\bf  40.0161}&  {\bf  28.9335} &  3.60 \\
 1AGT & 20 &  49.9861 &  39.1897 &  5.21 &      50.6311  &       39.0268  &  5.34 & --      &     25.6024  & 2.34 & {\bf  54.0897}&  {\bf  43.4210} &  5.45 \\
 1CRN & 20 &  74.7849 &  62.2668 &  7.60 &      74.4068  &       60.2444  &  7.58 & --      &     42.3083  & 2.96 & {\bf  82.5999}&  {\bf  68.3890} &  7.28 \\
 1HVV & 20 &  45.0054 &  35.9335 &  4.92 &      44.7264  &       34.8059  &  5.29 & --      &     21.5386  & 3.53 & {\bf  57.1990}&  {\bf  46.3685} &  5.61 \\
 1GK4 & 20 &  49.9316 &  42.0261 &  4.77 &      52.0651  &       44.8591  &  4.72 & --      &     27.0410  & 3.24 & {\bf  69.5798}&  {\bf  56.6853} &  5.16 \\
 1PCH & 80 & 121.0579 &  87.5748 & 11.42 &     103.1776  &       79.4878  &  8.85 & --      &     51.6674  & 3.50 & {\bf 128.4882}&  {\bf  99.3441} & 14.52 \\ 
 2EWH & 80 & 193.8143 & 162.3482 & 16.60 &     171.6390  &      144.9060  & 12.84 & --      &     94.5785  & 5.70 & {\bf 210.7021}&  {\bf 181.5912} & 17.49 \\
 \hdashline                                                                                                                                                
 F13  &  4 &   4.9533 &   3.0907 &  0.78 & {\bf  5.7290} & {\bf   3.6040} &  0.66 &  3.3945 &      2.8196  & 0.38 &        4.9704 &         3.1977  &  0.79 \\
 F21  &  4 &  11.1304 &   6.5538 &  1.53 &      11.2211  & {\bf   7.9567} &  1.53 &  6.9065 &      5.2674  & 0.76 & {\bf  11.7522}&         7.6885  &  1.75 \\
 F34  & 12 &  19.9550 &  13.3057 &  2.47 &      19.3529  &       14.0749  &  2.09 & 10.4224 &      8.3239  & 0.92 & {\bf  21.0345}&  {\bf  15.4491} &  2.85 \\
 F55  & 20 &  29.5163 &  22.4019 &  3.58 &      31.9554  &       24.6243  &  3.57 & 18.8385 &     14.4556  & 1.56 & {\bf  33.1788}&  {\bf  26.8111} &  3.34 \\ 
\end{tabular}
}
\end{table*}

\subsection{Comparison with other algorithms}
In this section, our algorithm is compared with other algorithms according to two stopping conditions $\mathit{NSE_{lmt}}$ and ${\mathit{t_{lmt}}}$, and
according to the best obtained energy values. 

The obtained results for stopping conditions $\mathit{NSE_{lmt}}$, that were set according to the literature \cite{Li15,BoskovicAB16} and three algorithms
\DElscr, \DEpfo\ and BE-ABC, are shown in Table \ref{tab:our:vs}. The best obtained energy values are marked in bold typeface. It can be observed that 
\DElscr\ and \DEpfo\ are comparable, and both outperformed BE-ABC. Results that take into account speed up factor, that are shown in Table
\ref{tab:localSearch}, are shown in the last column of Table \ref{tab:our:vs}. In this case, both algorithms \DElscr\ and \DEpfo\ spend approximately
the same amount of runtime, and \DElscr\ outperformed \DEpfo\ on most sequences. For the sequence 2EWH, the obtained values of $\mathit{E_{mean}}$
were 94.5785, 144.906, 162.3482 and 181.5912 for BE-ABC, \DEpfo, \DElscr, and \DElscr\ that take into account speed up factor. From these values, it is evident
that the proposed algorithm is superior in comparison with BE-ABC and \DEpfo.

\begin{table*}[t!]
  \centering
  \caption{The obtained results for \DElscr\ and \DEpfo within a runtime limit of 4 days. Entries that are shown as `-' imply that no results have
been reported in the literature.}
  \label{tab:4days}
 \begin{tabular}{rrrrrrrrrrrr}
  \multirow{2}{*}{Label} & \multirow{2}{*}{L} & \multicolumn{4}{c}{\DElscr, number of independent runs N=100} 
  & & \multicolumn{4}{c}{\DEpfo~\cite{BoskovicAB16}, number of independent runs N=30} \\ \cline{3-6} \cline{8-11}
  & & \multicolumn{1}{c}{$\mathit{E_{best}}$} 
    & \multicolumn{1}{c}{$\mathit{E_{mean}}$} 
    & \multicolumn{1}{c}{$\mathit{E_{std}}$} 
    & \multicolumn{1}{c}{$\mathit{hit_r}$} 
  & & \multicolumn{1}{c}{$\mathit{E_{best}}$} 
    & \multicolumn{1}{c}{$\mathit{E_{mean}}$} 
    & \multicolumn{1}{c}{$\mathit{E_{std}}$} 
    & \multicolumn{1}{c}{$\mathit{hit_r}$} \\
  \hline
  1BXP & 13 & {\bf   5.6104} & {\bf  5.6104} & 0.0000 & {\bf 100.00} & &           --  &           --   &    -- &          --  \\
  1CB3 & 13 & {\bf   8.4589} & {\bf  8.4589} & 0.0000 & {\bf 100.00} & & {\bf  8.4589} & {\bf  8.4589} & 0.0000 & {\bf 100.00} \\
  1BXL & 16 & {\bf  17.3962} & {\bf 17.3962} & 0.0000 & {\bf 100.00} & & {\bf 17.3962} &      17.1916  & 0.0878 &        6.67  \\
  1EDP & 17 & {\bf  15.0092} & {\bf 15.0092} & 0.0000 & {\bf 100.00} & & {\bf 15.0092} &      14.9423  & 0.0471 &       13.33  \\
  2ZNF & 18 & {\bf  18.3402} & {\bf 18.3402} & 0.0000 & {\bf 100.00} & &           --  &           --   &    -- &          --  \\
  1EDN & 21 & {\bf  21.4703} & {\bf 21.3669} & 0.0431 & {\bf   7.00} & &           --  &           --   &    -- &          --  \\
  2H3S & 25 & {\bf  21.1519} & {\bf 20.9956} & 0.0995 & {\bf  19.00} & &      20.0979  &      19.6147  & 0.2699 &        0.00  \\
  1ARE & 29 & {\bf  25.2800} & {\bf 24.5444} & 0.1718 & {\bf   1.00} & &           --  &           --   &    -- &          --  \\
  2KGU & 34 & {\bf  52.7165} & {\bf 51.7233} & 0.3829 & {\bf   1.00} & &      50.2960  &      49.1661  & 0.6334 &        0.00  \\
  1TZ4 & 37 & {\bf  43.0229} & {\bf 41.8734} & 0.4285 & {\bf   1.00} & &      39.7340  &      37.8329  & 0.9983 &        0.00  \\
  1TZ5 & 37 & {\bf  49.3868} & {\bf 48.6399} & 0.3292 & {\bf   1.00} & &      47.1513  &      43.9959  & 1.4087 &        0.00  \\
  1AGT & 38 & {\bf  65.1990} & {\bf 64.1285} & 0.4173 & {\bf   1.00} & &      62.8951  &      60.4175  & 1.0439 &        0.00  \\
  1CRN & 46 & {\bf  92.9853} & {\bf 89.8223} & 0.6514 & {\bf   1.00} & &      89.2001  &      86.0390  & 1.4529 &        0.00  \\
  2KAP & 60 &  {\bf 85.5099} & {\bf 83.1503} & 1.0041 & {\bf   1.00} & &           --  &           --   &    -- &          --  \\
  1HVV & 75 & {\bf  95.4475} & {\bf 91.4531} & 1.9215 & {\bf   1.00} & &      82.1427  &      68.8332  & 4.0852 &        0.00  \\
  1GK4 & 84 & {\bf 106.4190} & {\bf 99.6704} & 3.0377 & {\bf   1.00} & &      90.9140  &      84.6836  & 3.3356 &        0.00  \\
  1PCH & 88 & {\bf 156.5250} & {\bf153.1003} & 2.7117 & {\bf   1.00} & &     131.7787  &     117.7603  & 6.2617 &        0.00  \\
  2EWH & 98 & {\bf 245.5190} & {\bf240.2247} & 2.1421 & {\bf   1.00} & &     225.0968  &     203.6813  & 7.1844 &        0.00  \\
  \hdashline                                                       
  F13  & 13 & {\bf   6.9961} & {\bf  6.9961} & 0.0000 & {\bf 100.00} & & {\bf  6.9961} & {\bf  6.9961} & 0.0000 & {\bf 100.00} \\
  F21  & 21 & {\bf  16.5544} & {\bf 16.5304} & 0.0329 & {\bf  65.00} & &      16.2250  &      15.8894  & 0.1849 &        0.00  \\
  F34  & 34 & {\bf  31.3455} & {\bf 30.4913} & 0.3458 & {\bf   1.00} & &      28.2509  &      25.6602  & 1.0523 &        0.00  \\
  F55  & 55 & {\bf  51.9030} & {\bf 49.5009} & 0.8817 & {\bf   1.00} & &      45.0942  &      41.8670  & 1.4693 &        0.00  \\
  F89  & 89 & {\bf  81.5297} & {\bf 76.4804} & 2.0603 & {\bf   1.00} & &           --  &           --  &     -- &          --  \\
 \end{tabular}
\end{table*}
\begin{table*}[t!]
 \centering
  \caption{Comparisons of the best energy values reported in the literature and the best energy values obtained by \DElscr. 
  The solution vectors obtained by \DElscr\ are shown in Tables~\ref{tab:solutions1} and~\ref{tab:solutions2}. 
  Entries that shown as `-' imply that no `best energy values' have been reported in the literature.}
  \label{tab:energy}
  \begin{tabular}{rrrrrrrrrrrr}
  \multirow{2}{*}{Label} & \multirow{2}{*}{\DElscr} & \DEpfo & ImHS & BE-ABC 
                         & I-PSO & PGATS & MPGPSO  & ABC & GATS & C-ABC \\
                         & &  \multicolumn{1}{c}{\cite{BoskovicAB16}}  &  \multicolumn{1}{c}{\cite{Jana17b}} & \multicolumn{1}{c}{\cite{Li15,Li15f}} 
                         & \multicolumn{1}{c}{\cite{Chen11}} &  \multicolumn{1}{c}{\cite{Zhou14a}} &  \multicolumn{1}{c}{\cite{Zhou14b}}
                         &  \multicolumn{1}{c}{\cite{Li14}} &  \multicolumn{1}{c}{\cite{Wang09,Wang11}} & \multicolumn{1}{c}{\cite{Wang13}} \\
  \hline
  1BXP    & {\bf   5.6104} &            -- &  4.498 &   2.8930 &      -- &      -- &      -- &      -- &      -- &           -- \\
  1CB3    & {\bf   8.4589} & {\bf  8.4589} &     -- &   8.4580 &      -- &      -- &      -- &      -- &  8.2515 &           -- \\
  1BXL    & {\bf  17.3962} & {\bf 17.3962} & 15.200 &  15.9261 &      -- &      -- &      -- &      -- & 15.8246 &           -- \\
  1EDP    & {\bf  15.0092} & {\bf 15.0092} &     -- &  13.9276 &      -- &      -- &      -- &      -- & 13.7769 &           -- \\
  2ZNF    & {\bf  18.3402} &            -- & 15.056 &   5.8150 &      -- &      -- &      -- &      -- &      -- &           -- \\
  1EDN    & {\bf  21.4703} &            -- & 17.721 &   7.6890 &      -- &      -- &      -- &      -- &      -- &           -- \\
  2H3S    & {\bf  21.1519} &       20.0979 & 15.340 &  18.3299 &      -- &      -- &      -- &      -- & 18.1640 &           -- \\
  1ARE    & {\bf  25.2800} &            -- & 17.416 &  10.2580 &      -- &      -- &      -- &      -- &      -- &           -- \\ 
  2KGU    & {\bf  52.7165} &       50.2960 & 40.696 &  28.1423 & 20.9633 & 32.2599 &      -- & 31.9480 &      -- &           -- \\
  1TZ4    & {\bf  43.0229} &       39.7340 &     -- &  39.4901 &      -- &      -- &      -- &      -- & 39.3444 &           -- \\
  1TZ5    & {\bf  49.3868} &       47.1513 &     -- &  45.3233 &      -- &      -- &      -- &      -- & 45.3019 &           -- \\
  1AGT    & {\bf  65.1990} &       62.8951 & 40.300 &  51.8019 &      -- &      -- &      -- &      -- & 46.0842 &           -- \\
  1CRN    & {\bf  92.9853} &       89.2001 & 61.426 &  54.7253 & 28.7591 & 49.6487 & 43.9339 & 52.3249 &      -- &           -- \\
  2KAP    & {\bf  85.5099} &            -- & 44.972 &  27.1400 & 15.9988 & 28.1052 & 18.9513 & 30.3643 & 25.1003 &           -- \\
  1HVV    & {\bf  95.4475} &       82.1427 &     -- &  47.4484 &      -- &      -- &      -- &      -- &      -- &           -- \\
  1GK4    & {\bf 106.4193} &       90.9140 &     -- &  49.4871 &      -- &      -- &      -- &      -- &      -- &           -- \\
  1PCH    & {\bf 156.5252} &      131.779  &     -- &  91.3508 & 46.4964 & 49.5729 & 38.2766 & 63.4272 &      -- &           -- \\
  2EWH    & {\bf 245.5193} &      225.097  &     -- & 146.8231 &      -- &      -- &      -- &      -- &      -- &           -- \\
  \hdashline                                                   
  F13     &        6.9961  &        6.9961 &     -- &   6.9961 &      -- &      -- &      -- &      -- &  6.9539 & {\bf 7.0025} \\
  F21     & {\bf  16.5544} &       16.2250 &     -- &  15.6258 &      -- &      -- &      -- &      -- & 14.7974 &     14.9570  \\
  F34     & {\bf  31.3459} &       28.2509 &     -- &  28.0516 &      -- &      -- &      -- &      -- & 27.9897 &     28.0055  \\
  F55     & {\bf  52.0558} &       45.0942 &     -- &  42.5814 &      -- &      -- &      -- &      -- & 42.4746 &     42.2769  \\
  F89     & {\bf  83.5761} &            -- &     -- &       -- &      -- &      -- &      -- &      -- &      -- &           -- \\
  \end{tabular}
\end{table*}

Within this comparison, the value of $\mathit{NSE_{lmt}}$ was relatively small. This means the reinitialization mechanism did not have a
significant impact on the obtained results. Therefore, \DEpfo\ and \DElscr\ were also compared according to the ${\mathit{t_{lmt}}}$
that was set to 4 days. A grid environment was used within this comparison and results are shown in Table \ref{tab:4days}. In this comparison,
\DElscr\ obtained better values of $\mathit{E_{mean}}$, $\mathit{E_{best}}$ and $\mathit{hit_r}$ in most sequences, and equal values for the
shortest sequences. \DElscr\ obtained $\mathit{hit_r}$ of 100, 100, 19 and 65 for shorter sequences 1BXL, 1EDP, 2H3S and F21. For the same sequences,
\DEpfo\ obtained $\mathit{hit_r}$ of $6.67, 13.33, 0$ and $0$, respectively. Significant improvement was obtained for longer sequences too.
For example, the best energy values were improved from $90.914, 131.7787$ and $225.0968$ to $106.419, 156.525$ and $245.519$ for sequences
1GK4, 1PCH and 2EWH, respectively. The energy values were improved by 15.505, 24.7463 and 20.4222. Note that \DElscr\ obtained
the new best-known solutions for all sequences with $L \ge 18$, the $\mathit{hit_r}=100$ for 6 shortest sequences, and $\mathit{hit_r}>1$ for 9
sequences by using $\mathit{t_{lmt}} = 4$ days.

The most important results in this paper are shown in Table~\ref{tab:energy}, which collects the best energy values from all experiments that were described
in previous sections, and the best-known energy values from the literature. It is evident that \DElscr\ confirmed the best-known energy values for the 3
shortest sequences, and reached the new best-known energy values for all other sequences, except for sequence F13. Solutions for the best energy values
reached by \DElscr\ are shown in Tables~\ref{tab:solutions1} and~\ref{tab:solutions2}.

In~\cite{Kim16} an efficient global optimization method is applied to the sequence F89. 
Within this work, 32,200 distinct conformations were obtained, and the best obtained energy was 73.1065. \DElscr\ improves this energy
by 10.4695, as shown in Table~\ref{tab:energy}.

\subsection{Analysis of the obtained structures}
\begin{figure*}[t!]
\centering
\subfloat[2ZNF]{\label{fig:rmse:2znf}\makebox[.4\textwidth]{\includegraphics[scale=0.4]{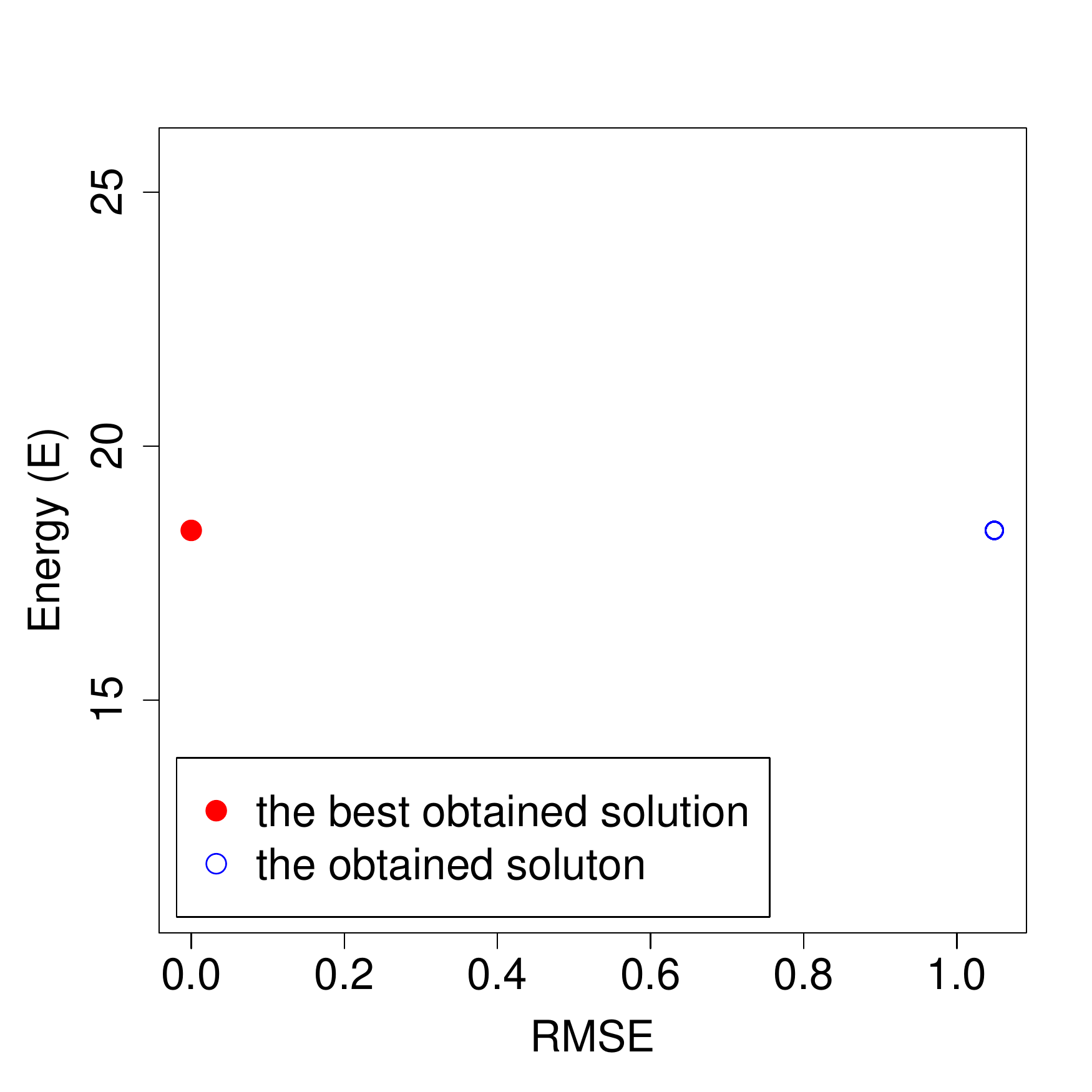}}}
\hfil
\subfloat[1EDN]{\label{fig:rmse:1edn}\makebox[.4\textwidth]{\includegraphics[scale=0.4]{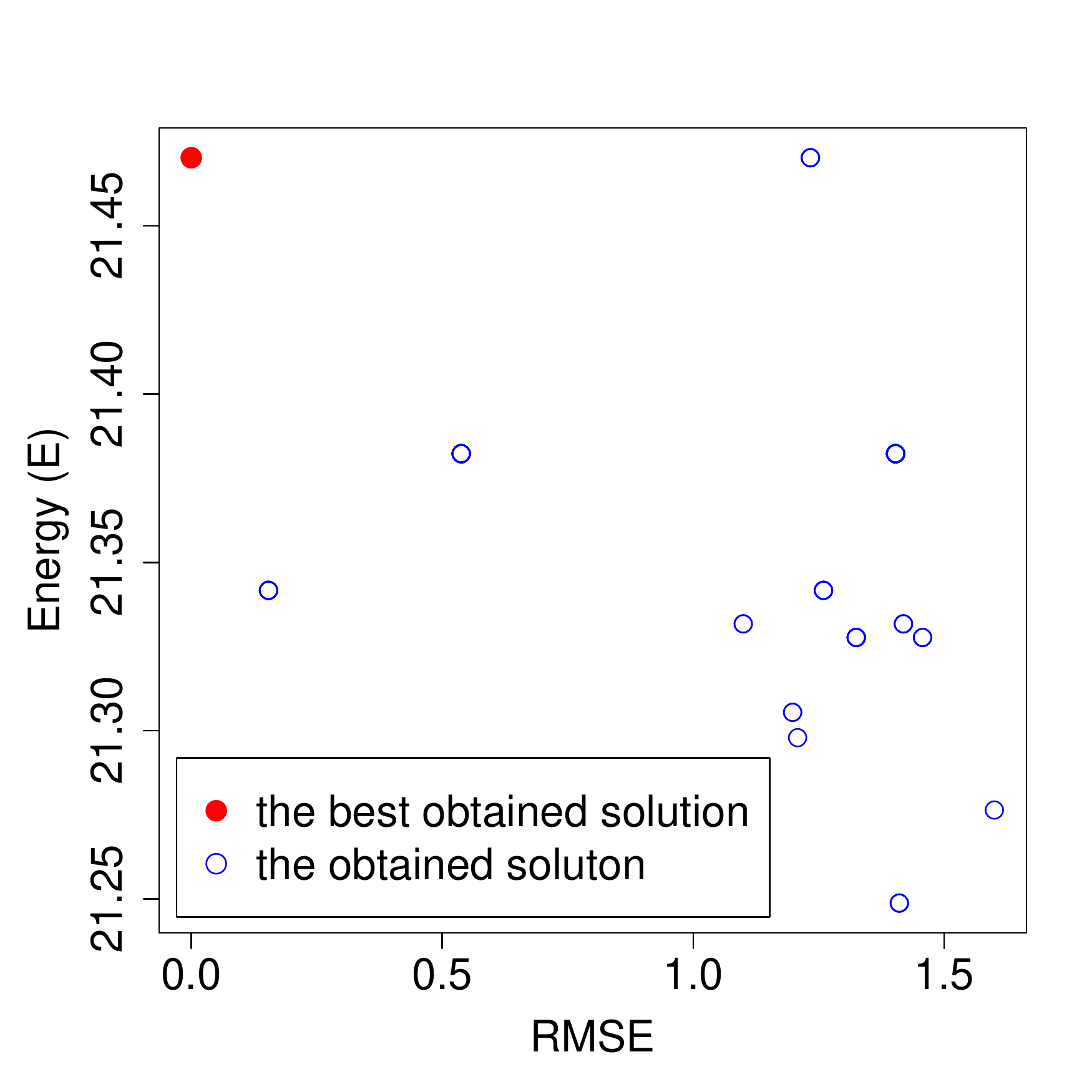}}}

\subfloat[2H3S]{\label{fig:rmse:2h3s}\makebox[.4\textwidth]{\includegraphics[scale=0.4]{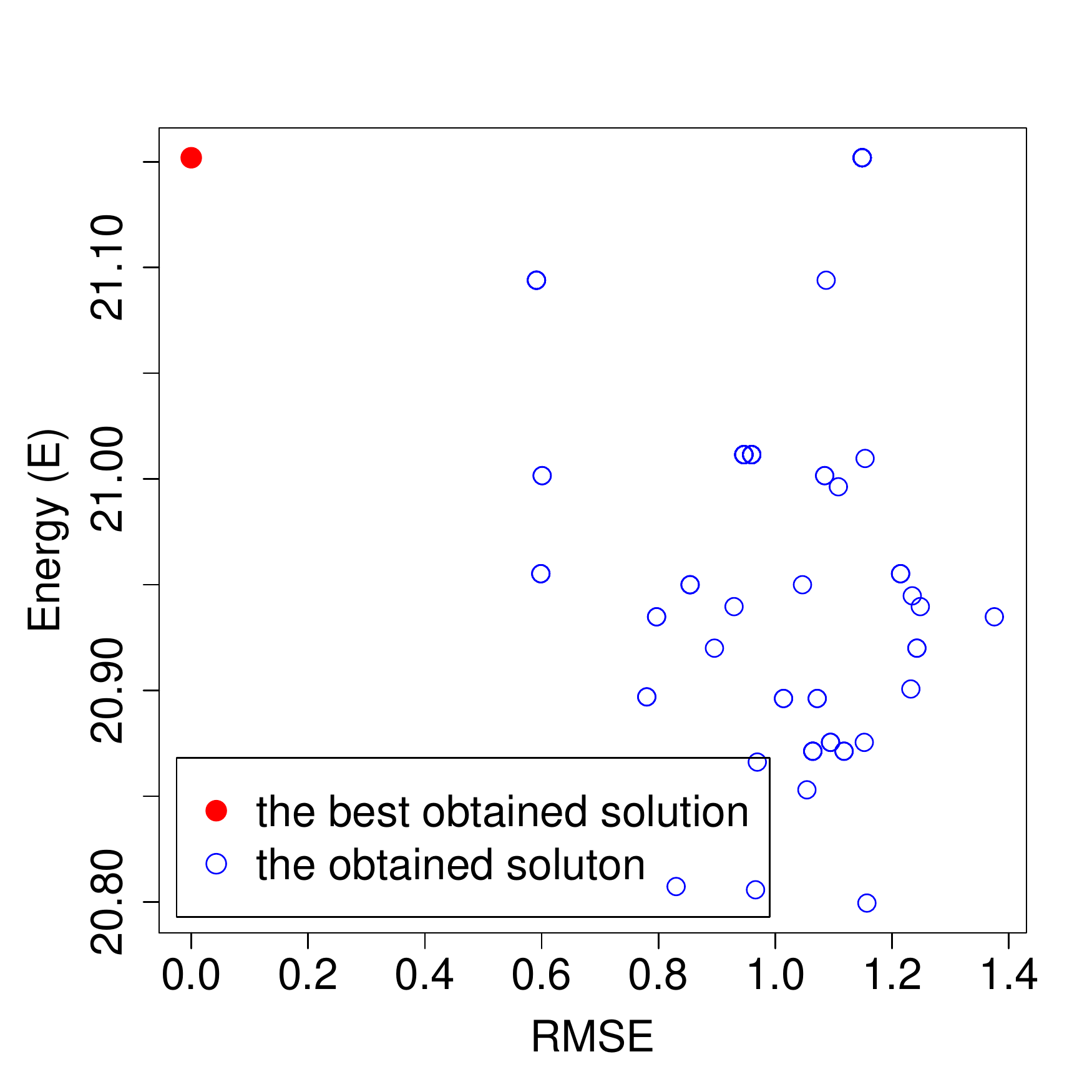}}}
\hfil
\subfloat[2EWH]{\label{fig:rmse:2ewh}\makebox[.4\textwidth]{\includegraphics[scale=0.4]{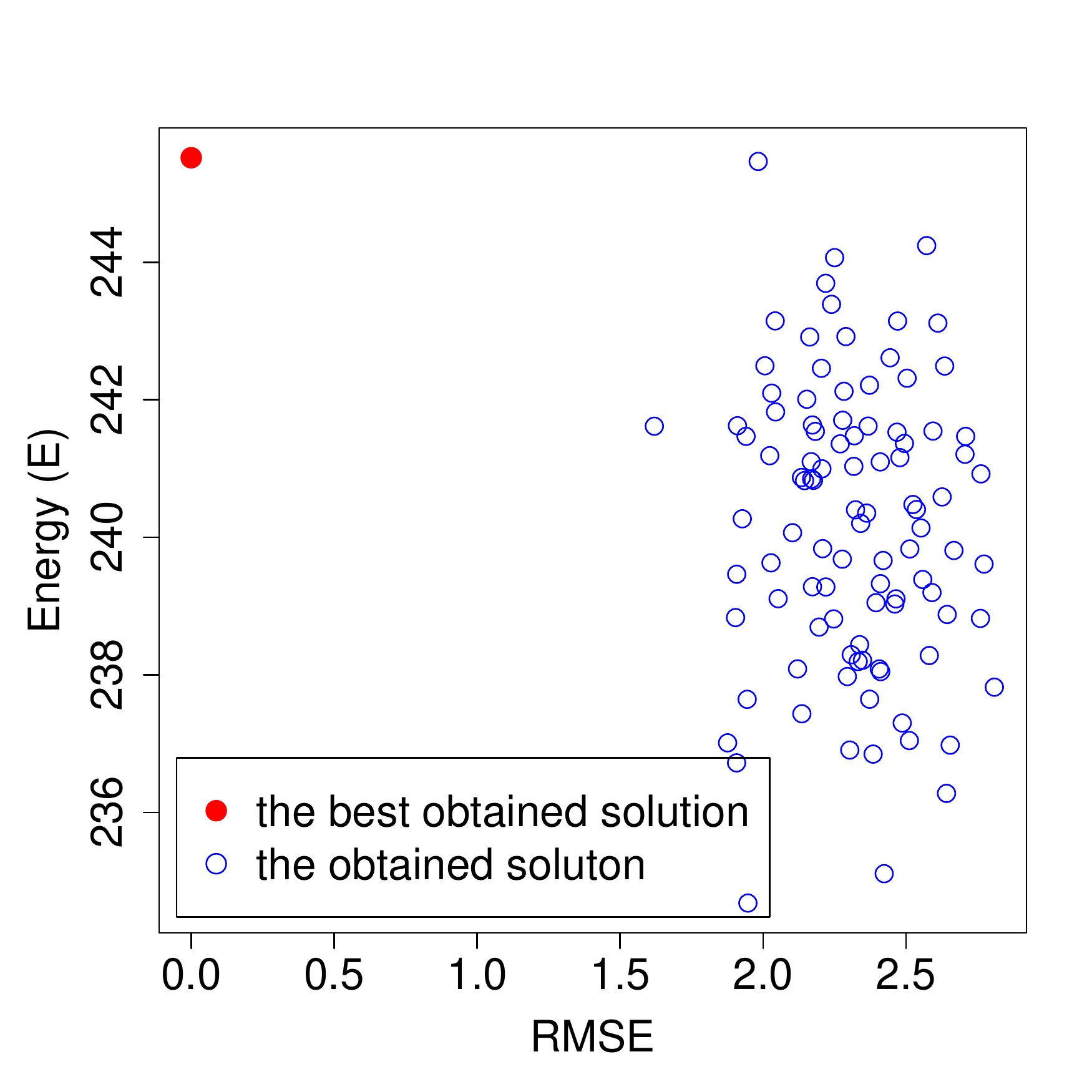}}}
\caption{Distribution of the Root-Mean-Square Error (RMSE) values as a function of energy for all 100 obtained conformations within a runtime limit of 4 days,
calculated from the best-known conformation.}
\label{fig:rmse}
\end{figure*}

\begin{figure*}[t!]
\centering
\subfloat[1BXP]{\makebox[.2\textwidth]{\includegraphics[scale=0.2]{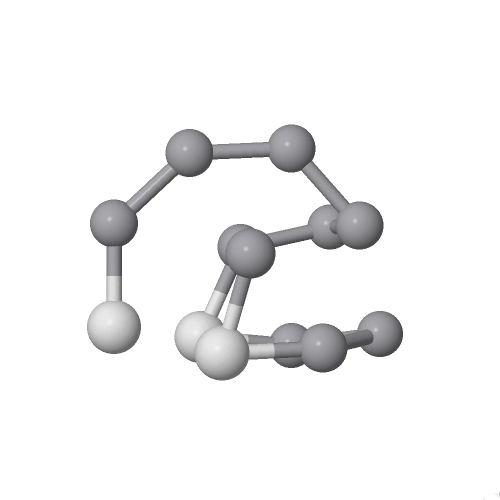}}}
\hfil
\subfloat[1BXP]{\makebox[.2\textwidth]{\includegraphics[scale=0.2]{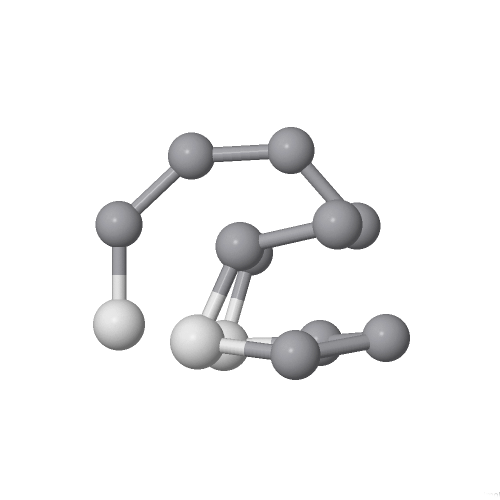}}}
\hfil
\subfloat[1CB3]{\makebox[.2\textwidth]{\includegraphics[scale=0.2]{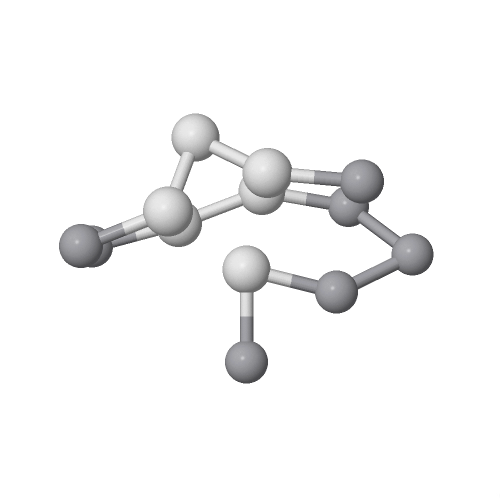}}}
\hfil
\subfloat[1CB3]{\makebox[.2\textwidth]{\includegraphics[scale=0.2]{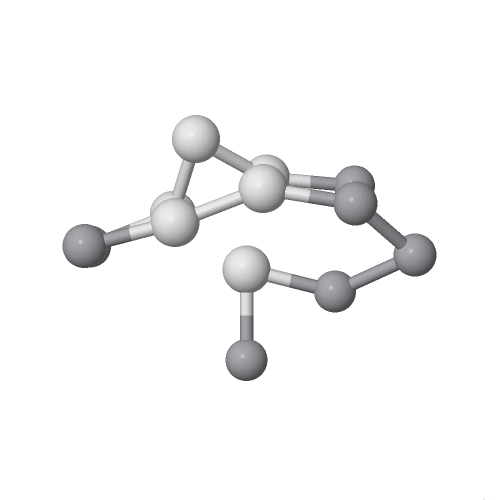}}}

\subfloat[1BXL]{\makebox[.2\textwidth]{\includegraphics[scale=0.2]{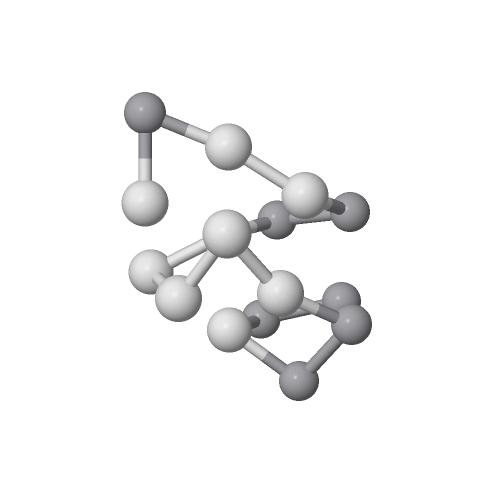}}}
\hfil
\subfloat[1BXL]{\makebox[.2\textwidth]{\includegraphics[scale=0.2]{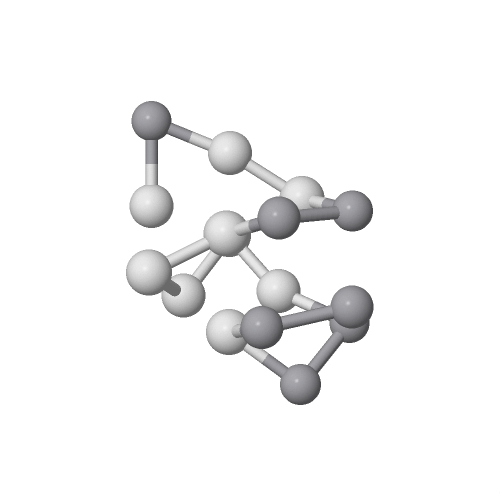}}}
\hfil
\subfloat[1EDP]{\makebox[.2\textwidth]{\includegraphics[scale=0.2]{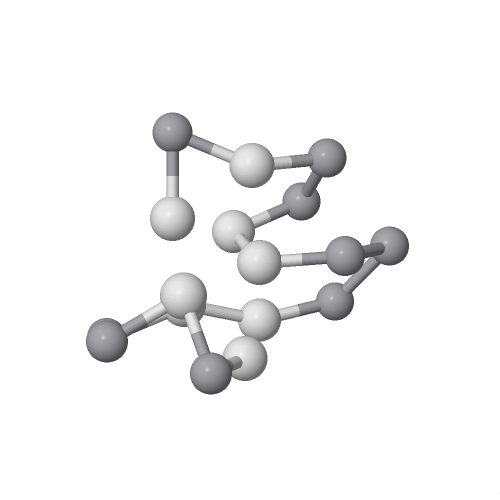}}}
\hfil
\subfloat[1EDP]{\makebox[.2\textwidth]{\includegraphics[scale=0.2]{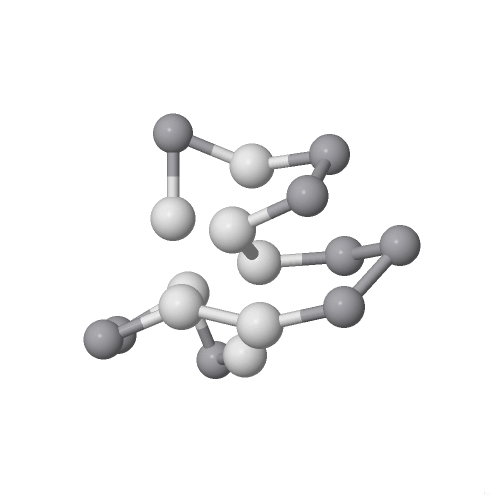}}}

\subfloat[2ZNF]{\makebox[.2\textwidth]{\includegraphics[scale=0.2]{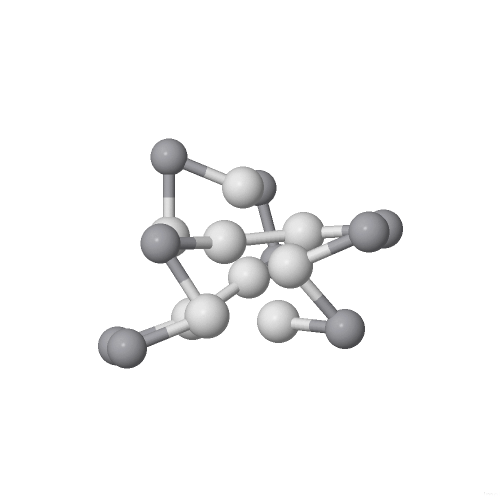}}}
\hfil
\subfloat[2ZNF]{\makebox[.2\textwidth]{\includegraphics[scale=0.2]{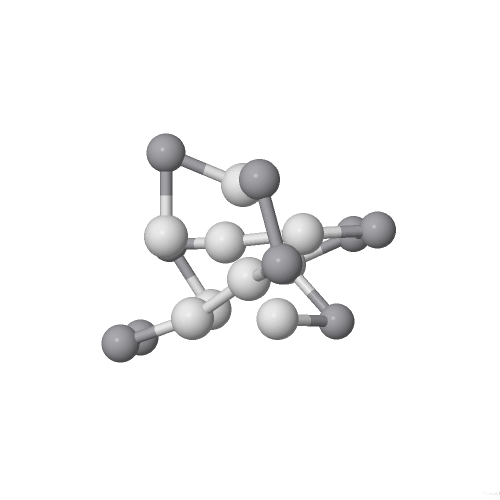}}}
\hfil
\subfloat[1EDN]{\makebox[.2\textwidth]{\includegraphics[scale=0.2]{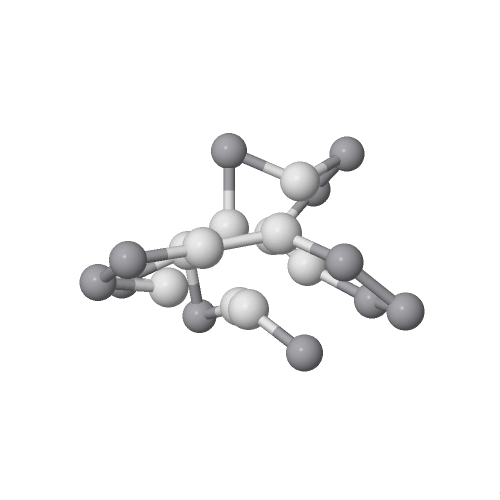}}}
\hfil
\subfloat[1EDN]{\makebox[.2\textwidth]{\includegraphics[scale=0.2]{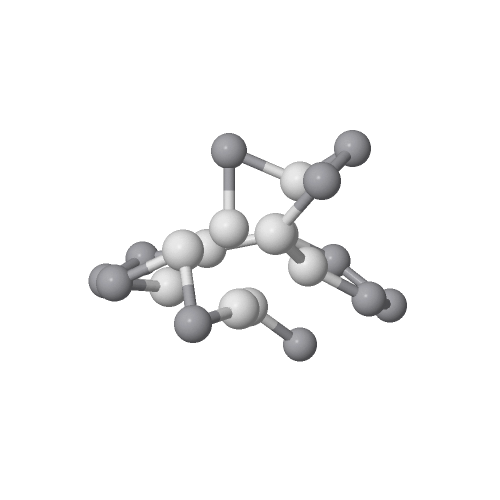}}}

\subfloat[2H3S]{\makebox[.2\textwidth]{\includegraphics[scale=0.2]{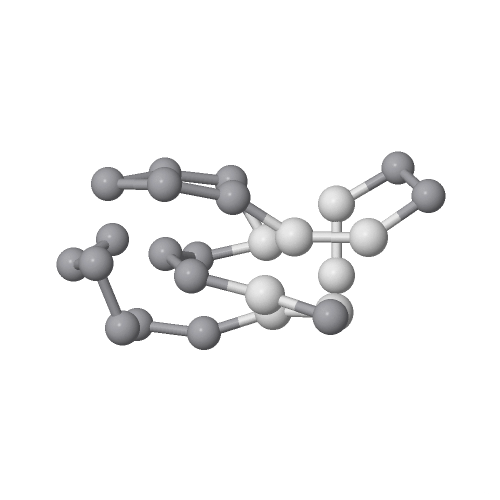}}}
\hfil
\subfloat[2H3S]{\makebox[.2\textwidth]{\includegraphics[scale=0.2]{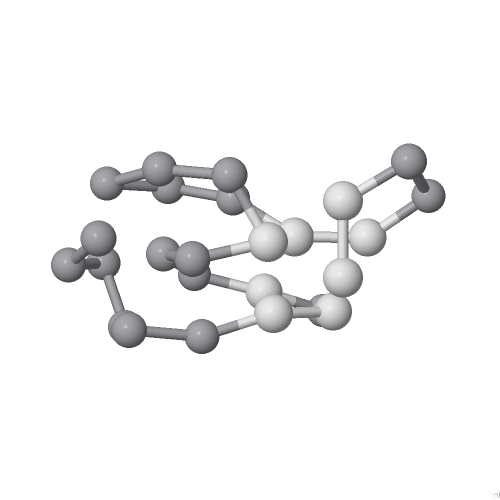}}}
\hfil
\subfloat[F13]{\makebox[.2\textwidth]{\includegraphics[scale=0.2]{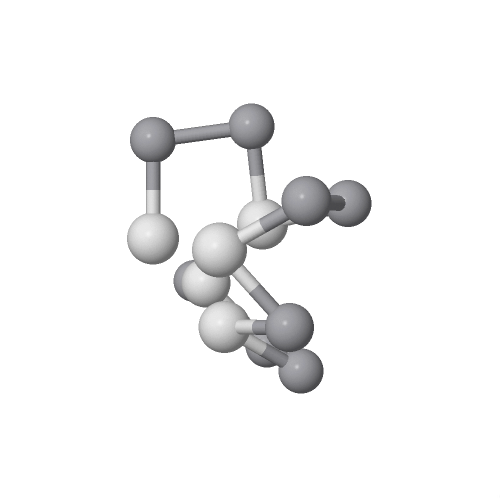}}}
\hfil
\subfloat[F13]{\makebox[.2\textwidth]{\includegraphics[scale=0.2]{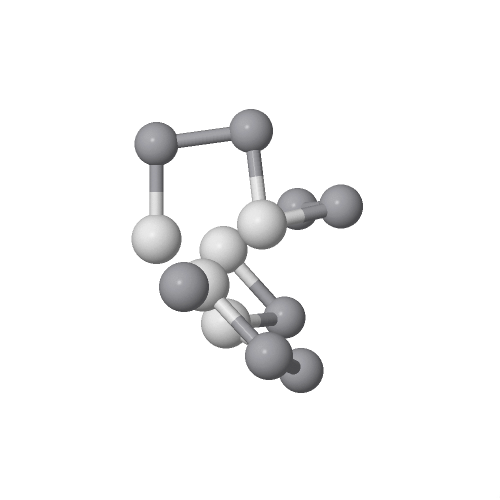}}}

\subfloat[F21]{\makebox[.2\textwidth]{\includegraphics[scale=0.2]{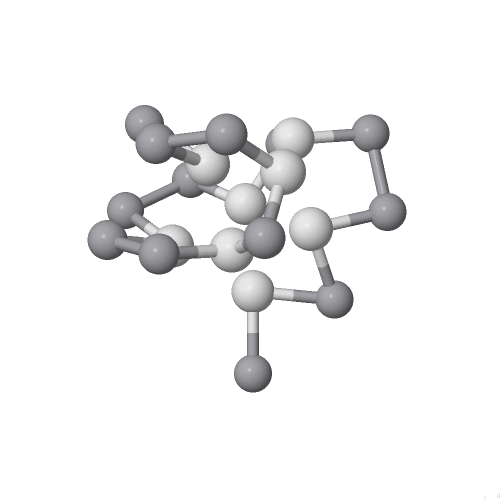}}}
\hfil
\subfloat[F21]{\makebox[.2\textwidth]{\includegraphics[scale=0.2]{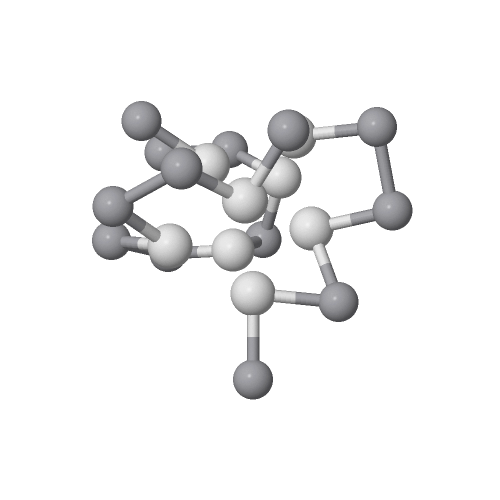}}}

\caption{The best obtained conformations that could be optimal.}
\label{fig:3dsolution}
\end{figure*}

For most of the sequences, the best conformations were obtained by using $\mathit{t_{lmt}} = 4$ days. Within this experiment, 100 solutions were
generated with 100 independent runs. Distribution of the Root-Mean-Square Error (RMSE) values as a function of energy for all these solutions
according to the best-known conformation for selected sequences is shown in Fig.~\ref{fig:rmse}. Note that the RMSE is calculated by using the
superposition between matched pairs. From Fig.~\ref{fig:rmse:2znf}, we can see that only two different solutions were reached for sequence
2ZNF. Similar graphs with only two different solutions were obtained for 6 sequences where $\mathit{hit_r} = 100$ (see Table~\ref{tab:4days}).
Three-dimensional representations of these solutions are displayed in Fig.~\ref{fig:3dsolution}. As is shown, for each of these sequences,
two solutions are symmetrical according to the XY-plane. This can also be seen from Tables~\ref{tab:solutions1} and~\ref{tab:solutions2}. 
Two reported solutions for one sequence are very similar. They are different in some components that belong to $\beta$ torsional angles 
(marked in bold typeface), and their values represent angles with opposite directions. For a little bit longer sequences more different solutions
were reached with different energy values, as shown in Figs.~\ref{fig:rmse:1edn} and~\ref{fig:rmse:2h3s} while, for the longest sequences, all
100 solutions are different with different energy values. For example, this is illustrated in Fig.~\ref{fig:rmse:2ewh} for sequence 2EWH. From
these results, we can conclude that all reported symmetrical solutions could be optimal, especially those obtained with $\mathit{hit_r} = 100$,
and all other solutions with $\mathit{hit_r} = 1$ are almost surely not optimal.

\captionsetup[subfloat]{font=scriptsize}

\section{Conclusions}
\label{sec:conclusions}
In this paper, we presented a novel Differential Evolution algorithm for protein folding optimization. To improve its efficiency, the algorithm 
is extended with a component reinitialization and local search that includes a local movement. The component reinitialization is designed to
redirect the search process to similar solutions that are different from the already found good solution in only a few components. Thus, the search
space around good solutions is explored thoroughly and, consequently, the algorithm can find better solutions. We also designed the local movement 
for a three-dimen-sional AB off-lattice model in such a way that only a two consecutive monomers are moved locally, while all the remaining monomers remain
in their positions. With additional data structure this type of movement allows us to reduce the runtime complexity of the energy calculation within the
local search from $\frac{L^2}{2}$ to $2 L$. 

The 23 sequences are used in the experiments to analyze the proposed algorithm and its mechanisms.
From the results of the algorithms with and without local search, it is evident that the local movement with additional data structure reduces the
runtime complexity of the energy calculation, or increases the number of function evaluations per second by factor 1.46 for the shortest sequences,
and by factor 3.91 for the longest sequence. This speed up is dependent on sequence length and the relationship between the number of solution evaluations
inside and outside the local search. The local search also improves the algorithm's convergence speed for most of the sequences. Because of both advantages,
the local search improves the efficiency of the algorithm, and this improvement is greater for longer sequences.

Using the best-known energy values as a stopping condition, we demonstrated the usefulness of component reinitialization. It reduces the required mean number 
of solution evaluations to reach the best-known energy value from 0.38 to less than 0.05. This indicates that the component reinitialization redirects the 
search process successfully to similar solutions, and allows the algorithm to locate the best-known solutions efficiently.

Our algorithm is the first algorithm that is capable of obtaining a hit ratio of 100\% for 6 shorter sequences within the budget of $10^{11}$ function evaluations.
Therefore, we introduce an approach for determining asymptotic average-case performances. Our algorithm obtained the best runtime asymptotic average-case
performance for sequence 2ZNF ($16.6030 \cdot 1.3505^L$) and the worst for 1BXP ($0.0015 \cdot 2.8911^L$). This approach shows additionally that 
the difficulty of the problem is not only dependent on sequence length, but also on the sequence itself.

The proposed algorithm was also compared with recently published state-of-the-art algorithms for PFO. It outperforms all competitors, and the obtained energy
values improve the best-known energy values from the literature for all sequences with $L\ge18$. For example, the best energy value of sequence 1PCH
was improved from 131.7787 to 156.5250 or by 24.7463.

The structure of the best obtained solutions was also analyzed. We figured out that two symmetric best-known solutions exist for sequences with $L \le 25$.
For these sequences, our algorithm obtained a hit ratio equal to or greater than 7\%. The solutions of these sequences could be optimal, especially those with 
a hit ratio of 100\%, and solutions for all other sequences are almost surely not optimal.

In the future work we will try to improve the algorithm further by using knowledge about symmetric solutions. This knowledge can be integrated within
the evaluation function, or used to reduce the size of the search space. Additionally, we will try to design an algorithm that will reduce the 
likelihood of the exploration of already explored search space.

\begin{table*}[t!]
  \caption{The best solutions obtained by the \DElscr\ algorithm.}
 \label{tab:solutions1}
 \scalebox{0.8}{
 \begin{tabular}{lp{19.8cm}}
  Label & Solution vector in degrees \\ 
  \hline 
  1BXP & 43.2915, 2.88166, -48.728, 0.0655009, 12.6242, 66.0927, -6.40805, 8.96332, 8.80015, 2.23544, 74.0763, {\bf -6.62061, 1.31798, 
         -104.099, 160.341, -177.384, -20.6892, 26.8003, 127.789, 166.27, 10.2979} \\ \smallskip
  1BXP & 43.2915, 2.88166, -48.728, 0.0654961, 12.6242, 66.0927, -6.40805, 8.96332, 8.80016, 2.23544, 74.0763, {\bf 6.62061, -1.31798,
         104.099, -160.341, 177.384, 20.6892, ${\bf -26.8003}$, -127.789, -166.27, -10.2979} \\ \smallskip
  1CB3 & -14.0758, 25.2546, -38.7359, -9.58086, 21.0366, 14.7617, -0.998265, 21.5393, 71.2738, -27.6012, -5.16526, {\bf -19.1483, -149.775,  
         172.54, 178.086, 178.164, 91.6772, 4.85452, -31.1093, 28.9806, 3.41538} \\ \smallskip
  1CB3 & -14.0758, 25.2546, -38.7359, -9.58086, 21.0366, 14.7617, -0.998265, 21.5393, 71.2738, -27.6012, -5.16526, {\bf 19.1483, 149.775, 
         -172.54, -178.086, -178.164, -91.6772, -4.85451, 31.1093, -28.9806, -3.41538} \\ \smallskip
  1BXL & -22.4292, -32.2737, -16.9254, 5.81295, 15.6175, 26.9979, -38.2372, 52.8361, -48.2442, -24.0736, 49.3335, -36.1178, 13.9215,  
         12.5486, {\bf 1.91872, 55.1452, 147.302, -127.63, 168.592, -62.9624, -27.0891, 28.7221, 27.4283, 152.122, -177.152, 67.7357, -5.21217} \\ \smallskip
  1BXL & -22.4292, -32.2737, -16.9254, 5.81296, 15.6175, 26.9979, -38.2372, 52.8361, -48.2442, -24.0736, 49.3335, -36.1178, 13.9215,
         12.5486, {\bf -1.91872, -55.1452, -147.302, 127.63, -168.592, 62.9624, 27.0891, -28.7221, -27.4283, -152.122, 177.152, -67.7357, 5.21217} \\ \smallskip
  1EDP & -22.6336, 7.26974, 60.7674, 23.936, -50.4261, 4.41672, 11.4886, 46.499, 13.2306, $-12.2668$, 22.7087, 4.07035, 30.6245, -69.1251,  
         16.9542, {\bf 26.0209, 124.911, ${\bf -155.575}$, -61.088, 1.55078, 53.7379, 159.421, -162.592, -156.44, -170.499, -85.1224, 2.36332, -25.7677,  
         67.3571 } \\ \smallskip
  1EDP & -22.6336, 7.26974, 60.7674, 23.9359, -50.4261, 4.41671, 11.4886, 46.499, 13.2306, $-12.2668$, 22.7087, 4.07036, 30.6245, -69.1251, 
         16.9542, {\bf -26.0209, -124.911, 155.575, 61.088, -1.55077, -53.7379, -159.421, 162.592, 156.44, 170.499, 85.1224, -2.36333, 25.7677,
         -67.3571} \\ \smallskip
  2ZNF & -22.512, 7.71692, -75.1038, 26.0695, 35.539, 19.645, 6.73951, 21.8104, -57.4641, 1.6924, 6.15567, 3.08902, 9.89786, 23.8155, -48.9192, 
         -4.31387, {\bf 78.7078, 2.66583, -114.943, -148.187, -162.564, -79.1176, 8.87759, -178.428, 42.9368, 15.8392, -18.6691,  -104.193, 166.46,
         12.876, 140.107} \\ \smallskip
  2ZNF & -22.512 7.71691, -75.1038, 26.0694, 35.539, 19.645, 6.73952, 21.8104, -57.4641, 1.6924, 6.15567, 3.08903, 9.89787, 23.8155, -48.9192, 
         -4.31386,  {\bf -78.7078, -2.66582, 114.943, 148.187, 162.564, 79.1176, -8.87759, 178.428, -42.9368, -15.8392, 18.6691, 104.193, -166.46,
         -12.876, -140.107} \\ \smallskip
  1EDN & -23.2048, 31.2208, 46.764, 48.9339, -43.6867, -28.0164, -17.6723, -38.3711, -25.1772, 10.6263, 9.07757, 33.5364, -4.83762, -6.09916,
         25.0581, -81.151, 15.5944, -3.62479, $-36.6783$, {\bf 41.0025, 127.461, -147.732, -53.6249, -22.4102, -68.6344, -166.972, 147.028, -171.451,
         -155.381, 121.71, 29.6786, 131.144, 15.2983, 24.5428, ${\bf -54.7787}$, -83.2637, -29.6805} \\ \smallskip
  1EDN & -23.2048, 31.2207, 46.764, 48.9338, -43.6867, -28.0164, -17.6723, -38.3711, -25.1772, 10.6262, 9.07754, 33.5364, -4.83764, -6.09917,
         25.058, -81.151, 15.5944, -3.62479, -36.6783, {\bf -41.0025, -127.461, 147.732, 53.6249, 22.4102, 68.6344, 166.972, -147.028, 171.451,
         155.381, -121.71, -29.6786, -131.144, -15.2983, -24.5428, 54.7786, 83.2637, 29.6805} \\ \smallskip
  2H3S & 30.6395, -51.1361, 34.4028, -0.410148, -32.439, -10.4102, -2.09416, 12.4798, -5.74202, -60.0842, 12.6704, -8.68551, -36.5963,  
         -14.4828,  -17.9172, 13.0794, 0.148026, 17.7335, -6.06512, 1.46386, -69.7022, 3.0363, 36.2347, {\bf -57.1061, -174.679, 173.256, -170.68,  
          -156.724, 142.58, 40.6316, 22.5668, -1.4454, 175.849, -114.818, -61.1893, -4.11275, -27.6809, 84.4735, 144.867, 176.731, 161.605,  
         -97.3254, -158.173, 113.225, 54.3451} \\ \smallskip
  2H3S & 30.6395, -51.1361, 34.4028, -0.410146, -32.439, -10.4102, -2.09415, 12.4798, -5.74202, -60.0842, 12.6704, -8.68551, -36.5963,
         -14.4828, -17.9172, 13.0794, 0.148029, 17.7335, -6.06512, 1.46387, -69.7022, 3.0363, 36.2347, {\bf 57.1061, 174.679, -173.256, 170.68, 
         156.724, -142.58, -40.6317, -22.5668, 1.44539, -175.849, 114.818, 61.1893, 4.11273, 27.6808, -84.4735, -144.867, -176.731,  -161.605,
         97.3254, 158.173, -113.225, -54.3451} \\ \smallskip
  1ARE & -11.5547, -1.31888, -15.2569, -42.6589, 17.4763, 5.06218, 14.6442, 24.251, 2.4936, -13.6942, 28.8183, -37.0023, -1.8785, -0.867767,
         -5.02831, 35.1061, -45.2208, -7.89329, 3.88011, -1.06756, -41.5237, 42.6134, 14.078, 1.71866, -70.4096, 19.7351, 23.088, -23.8173,
         -48.341, -27.8178, -175.825, 102.429, 138.697, -140.149, -46.4258, -148.917, -22.129, -165.228, 134.775, 48.8091, -12.7742, -50.1159,
         -163.389, 154.413, 126.397, -131.045, -60.8471, 167.884, 102.299, 52.433, -15.9838, -113.979, -58.9094 \\ \smallskip
  2KGU & -156.228, 84.3317, -1.89424, -22.9614, 4.96104, -10.8986, 42.0037, -54.9878, -4.36371, -80.394, 6.84565, -4.01855, -29.0786, 38.404,  
         -24.9304, 51.317, -53.2373, 15.7134, -51.9703, 1.34405, 37.6371, 36.5939, 35.6007, -52.9444, 32.6405, -108.259, -56.7621, 71.7249,  
         5.9403, 4.99762, 0.0626093, 8.48403, -161.728, -140.31, 137.06, 46.113, 21.1367, 45.0214, -27.4148, 37.097, -8.18763, -148.71,  
         107.671, -141.471, -176.445, 152.171, -23.7168, -63.0744, -154.472, 9.04166, -89.3673, 21.6149, -71.4051, 41.2427, -22.0274, 113.616,  
         22.7052, 159.166, -13.0884, -8.78814, 19.7018, 51.7085, 100.664 \\ \smallskip
  1TZ4 & -13.2782, 2.2117, -21.4873, 13.5614, -50.9456, -18.6314, 58.273, 35.906, -51.557, 43.4606, 14.4093, 26.9361, -9.90087, 51.937,  
         12.5408, -12.0182, -39.4559, -3.12819, -37.7837, 39.5619, 14.5525, -105.659, -2.39298, 23.2026, 13.2624, 7.00485, -63.913, 21.5608,  
         -2.32347, -4.49988, 14.2846, -2.28795, 25.5405, -52.7743, -3.52791, 88.4618, 172.62, 63.4655, 167.01, -112.19, -129.059, 165.903,  
         161.114,  -13.1829, -29.6599, -142.278, -118.354, -17.9561, 62.8846, 132.227, 150.392, 59.2519, -21.0203, -51.4332, 27.3873, 6.73164,  
         10.4224, -36.4599, -134.654, -177.842, -46.6888, 152.495, 60.1706, -2.02398, -21.7416, 80.9012, 145.281, -5.47398, -93.0592 \\ \smallskip
  1TZ5 & 19.2916, -25.4743, 37.9748, -2.24909, -71.5607, -62.5534, -1.20914, 60.5119, -37.8385, 56.051, -23.4795, 88.5824, -23.4208, 9.88257,  
         -27.3279, 2.64366, 4.72144, -32.8121, -37.9781, 28.6777, -1.79099, -1.45295, -1.55722, -28.7841, 53.5811, -7.1834, 28.1114,   
         -34.8815, -63.6796, -10.3914, 19.0723, -11.8679, -27.892, -37.0612, 62.2441, 52.8494, -172.157, 163.821, 57.7857, -56.1207, -158.293,  
         168.377, -23.9122, -20.1987, 4.84453, -2.90382, -36.4027, -130.461, 158.501, 160.429, -146.987, -129.731, 128.236, 34.5188, 28.0746,  
         -55.2047, -137.136, -167.33, 32.699, 35.1309, -64.3505, 35.5609, 22.6136, -27.8717, -112.09, 160.134, -131.465, -173.819, -165.595 \\ \smallskip
   1AGT & 26.1664, -11.7847, -37.6783, -3.62257, 75.6542, -40.6386, 24.7245, -92.2268, -13.2317, 23.7859, -94.496, -2.2422, 17.0125, 1.77619,  
          -39.6216, 113.613, 89.2441, -4.72246, -98.3805, -65.8427, 44.1904, -17.7537, -84.2295, -2.33273, -55.9952, -46.6065, -1.4073,  
          40.7682, -24.0458, 37.5641, 44.8005, -1.59772, 9.28805, 12.0621, -52.5228, 21.2213, 120.34, -179.746, -63.7778, 18.8875, 10.4973,  
          -6.76493, -21.1687, 57.2819, 176.355, 27.0186, -122.011, -33.6896, -59.127, -178.6, 14.4172, -4.50249, -165.323, -6.2693, 12.6379,  
          -54.0356, -62.0563, 7.63468, 127.685, 19.1897, 133.256, -157.687, -140.601, -147.971, -19.8046, 48.3979, 166.316, 54.8275, 156.099,  
          3.00416, -140.869 \\ \smallskip
  1CRN & 36.2044, -2.8726, -58.3456, -109.168, 67.0176, 63.8303, -34.4202, -9.13991, 27.0952, 6.87578, -175.904, -10.0323, -14.0642, 169.202,  
         139.54, 37.4324, -30.2161, 3.47033, 120.567, 8.05841, 74.7893, -51.1755, -78.2244, -6.56336, -37.0077, -0.404044, 22.8391, -11.2627,  
         -2.90337, -113.857, -122.645, -5.62342, 80.5936, -19.4761, 87.9704, 12.3212, -4.15216, 3.25955, 39.0676, 30.3454, 61.9086, 12.9802,  
         -97.5976, 8.44839, -76.9107, 30.3312, 41.3065, 24.8161, -34.1801, 20.2024, 33.4227, -14.1966, 80.1988, 28.2782, -166.287, -129.879,  
         -11.8488, 19.3045, -66.9439, -48.3743, -164.938, -12.7803, 7.1584, 29.2233, 12.4003, 39.0045, -58.1443, 52.0717, 43.5443, -0.577219,  
         -103.851, -146.583, -9.20535, -12.6377, -128.769, 27.4049, -32.4672, 16.7912, -135.646, -149.836, 98.1608, 22.2995, 23.6538, 11.9513,  
         104.699, -3.17593, 35.9215 \\ \smallskip
    2KAP & 46.469, 9.17062, -12.987, -39.265, -23.0536, 170.718, 7.46538, -139.561, 9.6654, -109.874, 39.7501, -77.1224, -8.16555, 82.4941,  -21.6873,
           93.4429, -10.6347, 10.7423, 20.8323, -4.45369, -11.4409, -5.20606, 147.482, 172.455, -52.8927, 8.19366, 92.0005, -44.9143, -45.2074, -1.91882,
           -16.8158, 4.77317, 17.6662, 124.018, 11.9037, -1.64667, 74.9571, 15.2233, -5.48327, -140.99, 19.2716, 15.4203, -48.7429, -34.6525, 5.87344,
           -6.16017, 41.6324, -16.8426, 49.2516, -28.6507, 29.562, -13.1308, 17.3443, -61.6342, 8.11077, -104.361, 26.4602, 4.19116, 18.9505, 67.8863,
           154.188, -116.346, 19.3865, -84.7317, -27.0054, -31.7152, -24.4805, -34.1621, -13.632, 63.0618, 8.2678, -2.60225, 37.1772, -9.33353, -46.953,
           -117.05, -167.457, 155.733, -103.59, 35.8403, 44.471, -168.41, -18.09, 28.324, 22.7177, -44.4081, -37.3318, -3.25116, 42.16, 57.0729, -4.27289,
           -157.026, 158.794, -19.8247, 37.3666, 142.492, -5.84281, 168.076, -163.22,  -4.06088, -38.3009, 6.22285, -20.4278, 52.41, 156.856, 11.6928,
           124.907, -164.531, 100.594, -172.86, -66.3312, -55.802, 13.087, 38.8668, 61.5511
  \end{tabular}
  }
\end{table*}
\begin{table*}[t!]
  \caption{The best solutions obtained by the \DElscr\ algorithm.}
 \label{tab:solutions2}
 \scalebox{0.8}{
  \begin{tabular}{lp{19.8cm}}
    Label & Solution vector in degrees \\ 
  \hline
  1HVV & 123.296, 175.071, -104.639, -8.91165, 17.898, 3.71452, 16.7371, -6.98533, -76.7403, -9.21776, 75.2042, 124.487, 46.9457, 18.594, 
	-46.3806, 145.54, 124.742, -1.01034, 2.75056, 10.5349, -7.0204, 118.317, 75.4052, -30.5719, 69.7023, -72.9042, 176.87, -87.9407,  
	-48.7546, 88.4046, -11.7349, 44.1027, -5.14578, -65.6298, 20.3053, -20.9559, -72.1726, 94.427, 45.6623, -45.1943, 55.0226, 121.96,  
	34.0003, 6.44231, 140.467, 17.4955, 85.266, -21.9109, -95.2265, -24.295, 2.64738, -17.4667, 155.115, -5.99421, -158.933, -6.23019,  
	-43.4673, -28.7354, -11.6413, -19.2361, -10.0374, 34.4593, -178.973, -160.129, -42.443, 5.40211, -87.2981, -36.4155, 10.9084,  
	64.5935, -5.14139, 28.5554, -18.3091, -3.52734, 45.1155, 9.74914, 114.875, 75.0862, 26.4021, 22.7921, -63.2735, -138.233, -64.6958,  
	-1.96622, 99.8458, -16.8097, 71.6329, 28.1128, -43.4581, -82.0519, 173.534, -96.2117, -16.7565, -32.4898, 48.5607, 11.8749,  
	-64.2088, 1.38784, -57.9374, 25.3469, 54.5768, 34.603, -28.9913, -102.93, -64.0207, -23.3654, 157.688, 118.979, 5.08255, 61.9001,  
	-18.6156, -45.7361, -10.7387, 47.8304, -26.517, -104.738, 22.5679, 115.444, -6.44774, 116.951, 14.5305, -0.364305, 25.7291, 114.23,  
	-22.2869, 97.1472, 0.163606, -93.1095, -14.0922, 67.6929, 103.708, 12.5582, -62.8467, -103.687, -55.8599, 11.7904, 63.4791,  
	74.9204, 26.812, -72.0632, -116.517, -46.752, 39.4218, 63.2466, 101.367 \\ \smallskip
  1GK4 & -156.276, -11.1754, -77.6765, -33.983, -9.25845, 27.2496, -22.3371, 55.198, -12.6002, 30.4017, 35.9669, -18.5758,  -87.0091, 63.2376,
         163.731, -40.6359, 102.397, 146.119, -16.2788, 0.676412, 169.967, -18.769, -43.9974, -62.1645,  -51.3378, -148.518, 144.287, 
         -170.774, -25.6785, -22.1326, 27.6078, -8.93322, -75.4448, -68.3403, -16.5373, -128.752, 53.7389, -140.548, -115.598, -1.7515,  
         -106.07, -14.6299, 82.7281, -28.3651, 36.138, 37.748, -23.152, -11.6747, 20.0652, -55.0055, -150.457, -49.5769, 23.078, 65.4316,  
         111.217, 34.2085, 8.50149, 21.775, -107.598, 139.532, -69.7933, 68.8643, 88.8784, -166.542, -174.708, 31.9934, 137.315, -3.2063,  
         -88.0773, -6.44718, 23.1364, 95.4164, -34.9923, 2.63385, 13.9679, -23.3478, -54.8455, 43.5591, -4.20436, -43.6293, -144.761, 18.6453,  
         179.845, -29.3419, 34.6059, 116.989, -2.74396, 108.058, 157.394, 13.5671, -34.673, -71.5358, 29.1635, -12.3351, -58.4977, -28.7525,  
         -42.495, -47.6578, 29.0888, -101.645, -92.0769, 10.8024, 101.05, 70.2237, 17.3966, -48.5046, 27.4408, 61.4589, 71.0815, 77.6142,  
         36.2921, -18.009, -121.503,  -122.087, -32.2391, 9.22399, 11.065, 70.686, 48.1796, -38.5792, -91.5736, 15.5793, -173.479, -25.563,  
         108.153, 54.9784, -50.888,  -86.9754, -52.2761, -23.2842, 20.5012, 16.6481, -23.1703, -40.3118, -64.6739, 26.9541, 177.66, 154.029,  
         125.888, 2.61904, -11.91,  -67.2757, 7.45577, 48.5355, -2.84037, -62.5508, -1.94643, 7.24978, 48.2981, 62.3802, 127.482, 80.9534,  
         -6.93496, -105.181, -59.8353, -45.6067, -113.32, 1.58149, -33.4263, -8.31888, 28.5097,  -8.75938, 151.901 \\ \smallskip
  1PCH & 41.1106, -75.4478, -108.35, -88.2808, 52.6905, 63.9346, 98.575, -146.011, -103.251, -81.3915, -7.97873, -5.67391,  -138.034, 77.0572,
         -3.43666, -168.458, -44.1355, -51.9614, -0.548426, -11.6916, -13.85, -90.7331, 162.157, -85.119, -80.3963, 2.41355, -4.42424, 54.2753,
         89.5966, -147.035, 78.814, 99.0202, -89.7149, -126.891, -85.1937, 16.9404,  -73.4292, 66.5445, 160.832, 40.1368, 145.935, 17.6023, 
         -17.2377, 45.6224, 136.733, -49.815, -43.4151, -102.65, 8.11703, -109.171, -24.407, -47.2699, -132.484, -42.5279, -119.761, 147.911, 
         73.3968, 3.65598, 74.2541, 19.778, -2.2315,  -111.056, -25.0304, -86.2442, 0.146651, 83.5428, -29.3382, 64.0093, 147.03, 73.9582, 
         29.3652, 154.906, 6.25221, 112.96, 165.03, 11.5624, -48.3887, -130.92, 158.099, -6.47314, -26.6561, -18.5458, -79.5949, 16.5753, 
         68.4287, 54.4969, 8.59587, -21.8077, -38.0325, -14.1827, -48.149, -7.71342, 5.32628, 34.3591, 4.97453, -19.9588, -29.6581, -39.8474, 
         152.45, -7.02004, -71.166, 40.6025, 90.9133, 41.1172, 10.2277, -41.2978, -21.4705, -9.53513, 33.3237, -170.287,  -147.144, 
         -21.8209, 27.7537, 19.2982, 29.7286, -13.7392, -55.8712, -28.3766, -8.00291, 125.209, 161.205, 19.9805, -46.2199, -2.73427, -12.1052, 
         106.535, 20.4942, -28.1138, 4.59628, 62.9273, -0.534962, 6.33028, -7.44623, 29.7984, -42.8088, 5.32914, 17.9307, 174.593, -36.0277, 
         3.2934, 24.3011, 174.231, -1.94325, 107.108, -4.82491, 44.1005,  -38.1958, 47.1931, 0.689977, -48.5599, -2.29967, -16.9223, -9.36993, 
         -13.2723, 23.7414, 17.0389, 157.955, 52.6113, 14.0859, -100.053, -42.2958, -9.25703, 37.9416, 58.9846, -28.9415, -56.1025, 37.1207, 
         50.4847, -55.062 \\ \smallskip
  2EWH & 151.436, -92.5903, -3.75076, -9.668, -0.975246, 10.9844, 84.1865, -57.1686, 38.7048, 72.7755, 64.7086, 50.2283, 29.5956, -111.649, 163.46, 
         -92.508, -148.929, 61.6607, 124.085, 171.298, -75.5506, -52.2331, -1.37142, -19.9511, 67.108, 14.27, -1.03143, 0.0222875,  -116.144,
         -32.3217, -125.421, -102.86, 145.85, 108.298, -60.4006, -54.4374,  -88.1894, -14.6692, 121.958, 125.926, 167.02, -74.5737, -130.938, 62.1354,
         106.58, 35.1401, 93.7873, 162.441, 14.5873, -4.93593, 6.27325, -7.31527, 10.8045, -53.8271, -132.021, -37.8517, 30.6805, -89.4307, -61.2859, 
         -31.1104, 90.5217, 118.145, -4.46292, -52.5412, 113.612, 159.141, -3.43442, 62.945, 12.7417, -19.7111,  -15.1737, 30.368, -27.5934, 
         -138.544, 7.81197, -59.2248, 9.7981, 122.127, -164.755, 36.2949, 27.611, -37.779, 39.5707, -22.2883, 38.9922, 4.30684, 71.8969, -21.7556, 
         -128.587, -76.5211, -0.480596, 65.3271, 8.48192, 158.34, 107.055, -66.142, 18.5806, -129.465,  -18.4294, -11.7806, 50.9098, 132.39, 101.434, 
         0.100595, -28.2695, 0.622333, 52.5959, 150.715, -11.0556, -45.8273, -39.1155, 8.30602, 56.8165, -46.0845, 17.9186, 11.1667, 19.5044, 89.7449, 
         -23.746, -0.808862, 84.8186, 0.457484,  -89.925, -21.9539, 26.9255, -59.9087, 45.3057, -9.74849, -70.462, -14.0544, 18.7807, 59.6121, -176.746, 
         -30.6629,  -54.0245, -8.68261, -14.3952, 43.302, 23.9289, 59.5115, -35.76, -58.2711, 2.27009, -152.857, -35.6967, -137.17, -49.407, 36.9083, 
         58.3246, -26.7135, -71.9911, 26.5305, 53.0114, 52.0147,  -21.6699, -14.1031, 12.009, -63.6288, -54.0997,  -10.6967, 5.20745, -124.069, 
         -19.2082, 55.6145, -3.07942, 12.0508, 131.359, -177.867, 32.7929, 24.0475, 25.0758,  -0.0780254, 5.44473, 36.0356, 28.5948, 22.5261, -174.132, 
         -3.91424, -11.5903, -132.095, -172.313, -25.055, -156.073, 11.1264, 29.3397, -64.3192, -38.8408, -177.009, 22.2676, 11.9142, 56.3727 \\ \smallskip
    F13 & 7.66522, -83.448, 13.0886, 0.55134, 29.1616, -47.908, 2.75327, -31.0327, -31.3119, -46.3918, 0.276218, {\bf 9.04884, -29.5745, -116.199, 160.508, 
          0.890189, 129.381, 24.5074, 113.38, -161.672, 98.7127} \\ \smallskip
    F13 & 7.66522, -83.448, 13.0886, 0.551338, 29.1616, -47.908, 2.75327, -31.0327, -31.3119, -46.3918, 0.276222, {\bf -9.04884, 29.5745, 116.199, -160.508,
         -0.890189, -129.381, -24.5074, -113.38, 161.672, -98.7127} \\ \smallskip
    F21 & -5.70817, -70.6345, 12.6013, -78.4561, 5.14012, 2.49148, 57.5974, -25.416, 27.2287, -35.8677, -5.33428, -13.9895, 3.02158, 19.9054, 74.4006, 
          -31.0707, 4.76465, -19.1022, -32.9492, {\bf 155.506, -16.0013, -169.101, 162.893, -94.9124, 155.503, -140.891, 153.332, 40.6752, 137.563, 48.1957, 
          -35.2245, 66.7533, -37.5734, 137.909, -144.521, -52.7295, -156.871} \\ \smallskip
    F21 & -5.70816, -70.6345, 12.6014, -78.4561, 5.14014, 2.49149, 57.5974, -25.416, 27.2287, -35.8677, -5.33427, -13.9895, 3.0216, 19.9055, 74.4006,
          -31.0707, 4.76466, -19.1022, -32.9492, {\bf -155.506, 16.0013, 169.101, -162.893, 94.9124, -155.503, 140.891, -153.332, -40.6752, -137.563, -48.1957, 
          35.2245, -66.7533, 37.5734, -137.909, 144.521, 52.7295, 156.871} \\ \smallskip
    F34 & 12.3298, -83.1718, 20.1532, 8.42606, 37.8998, -37.8448, 9.33408, -77.8143, 7.4245, -73.1774, 26.15, -80.0668, 46.3843, 6.49943,  -29.8816, 51.2622, 
          -33.6564, 38.6885, -67.9543, 46.7986, -10.4886, -27.9647, -10.0583, -39.8364, -49.6972, -25.641, 44.7456, -59.6061, 18.6305, -20.9127, 25.4877, 
          13.4228, 1.77009, 42.1284, 129.207, -149.941, 1.89517, -120.166, 18.4003, 159.01, -168.548, 143.358, 151.62, -49.9323, -164.471, -44.6816, 177.501, 
          -32.6178, 2.86468, -2.00479, -22.1516, -57.0231, -143.09, 131.37, -127.956, 147.157, 57.657, -21.2642, 27.2822, -52.9505, 17.7835, 119.254, 18.7327 \\ \smallskip
    F55 & -15.6437, 97.9193, 1.00666, 95.3815, 1.86855, -64.0331, -141.452, -2.83476, 104.146, 8.21281, -162.93, -74.3953, 1.96392, 7.65968, -29.2495, 52.5953,
          52.8264, -0.624594, 137.07, -4.89079, 0.957561, 150.771, 19.388, 7.34186, 59.4269, 8.22775, -64.6383, -54.8633, -8.8461, 59.752, 162.033, 13.6066, 
          -78.2664, 13.0242, 102.375, 3.23899, -2.60196, -16.3626, 36.9652, -37.8734, 30.0569, 3.86882, -34.6667, -22.5344, 25.3408, 89.0776, 16.5037, -17.6911, 
          -91.108, 4.84917, -9.27247, -4.88184, 6.63221, 4.31554, -28.558, -8.46761, 171.776, -66.1542, 29.7446, -114.307, -0.574113, 83.8969, -5.34284, 64.1091, 
          6.16523, 112.965, -5.20239, 68.6776, -17.6844, -113.527, 31.5684, -100.963, 152.521, -49.1181, -26.2181, -129.399, -24.1061, 42.4841, -45.0845, 67.3054, 
          -35.1296, 15.741, 34.8919, 33.4461, 7.88754, 179.833, 41.9367, -135.639, 118.444, -156.824, -55.2589, 169.967, 0.172651, 91.1955, -4.02574, -81.307, 
          23.9152, -29.0528, -154.072, -133.971, 16.3457, -93.6495, -5.71876, 80.2119, -21.5927 \\ \smallskip
    F89 & 179.886, -95.54, -17.2583, -27.4664, -23.3,38.1298, 19.5766, 34.4132, 8.84695, -117.344, -20.5596, -131.943, -11.8962, -48.6976, 129.133, -107.932,
          7.29644, 89.0642, 21.8161, 56.2631, -33.6006, -22.3247, -47.5945, -48.8527, -51.6802, 36.4829, 85.1461, 3.54137, 107.149, -0.955333, -51.2341, 65.8004,
          13.8676, -69.7918, -0.860014, -134.319, -37.8356, -2.74527, -12.5366, -93.6285, -28.5384, -105.157, 19.0699, 81.6706, 6.93831, 0.887398, 116.484,
          23.1153, 132.738, 10.4558, 73.2237, -15.4114, 6.88586, -109.859, -3.3155, -82.7065, 2.76043, -42.4804, 82.5479, -18.3209, -29.9615, -74.7318, -24.0277,
          60.1736, -26.3071, -15.531, -14.9412, -79.5093, -1.99245, -48.5295, -70.5006, 58.6443, -42.9465, 50.0326, -70.0616, -9.55698, -109.482, -2.75044, -87.6997,
          -8.8569, 86.5537, -22.4479, 72.1052, 14.0501, -27.4652, -8.8744, 56.3962, 0.863411, -142.989, -54.5377, 29.0611, -61.1795, 50.3774, -12.8387, 73.4752,
          12.0947, -39.6898, -28.42, 143.035, 28.1471, 39.6651, 10.0519, -140.34, -2.35037, 123.344, 3.62448, 125.741, 132.141, 71.1956, -36.3432, -36.7204,
          -39.5973, 57.8245, -31.8281, 13.6268, -143.946, -36.4178, -6.53297, 34.1645, 12.4669, -82.0619, 14.2377, -32.9623, 49.1945, 137.212, -16.0272, -178.526,
          12.3581,69.7334, -1.88293, 147.327, 145.168, 27.4845, -35.3688, 8.48146, -81.2594, -5.25881, 119.388, -139.654, 57.454, 159.88, 27.509, -0.616009, 114.258,
          -13.2053, -39.494, 65.2585, -42.1881, 0.311959, -22.2436, -162.899, -54.8573, -20.7022, -14.7214, 128.993, 5.42026, 114.069, -21.3562, -46.7891, 18.6434,
          15.3431, 121.287, -3.95967, -82.6683, -9.4711, -120.912, -1.33882, -28.956, 43.9338, -42.9642, -139.445, 137.938, 4.62324
  \end{tabular}
  }
\end{table*}

\section*{Acknowledgements}
The authors acknowledge the financial support from the Slovenian Research Agency (research core funding No. P2-0041).

\bibliographystyle{spmpsci}      

\end{document}